\documentclass{article}

\usepackage{enumitem}
\usepackage[margin=2cm]{geometry}
\usepackage[utf8]{inputenc}
\usepackage{indentfirst}
\usepackage{newtxtext}
\usepackage{pifont}
\usepackage{colortbl}
\usepackage{bookman}
\usepackage{graphicx}
\usepackage{adjustbox}
\usepackage{subcaption}
\usepackage{booktabs}
\usepackage[parfill]{parskip} % Cleaner, NeurIPS-style paragraphs (no indentation; slight spacing between paragraphs added); uncomment to use.
\usepackage[font={small}]{caption}
\usepackage[numbers,sort&compress]{natbib}
\usepackage{tikz}
\usetikzlibrary{shapes,arrows.meta,positioning,trees,mindmap,backgrounds}
\usepackage[hang,flushmargin]{footmisc} 
\usepackage{multirow}
\usepackage{ragged2e}
\usepackage{floatflt}
\usepackage{wrapfig}
\usepackage{caption}
\usepackage{float}
\usepackage{amsmath}
\usepackage{placeins}

\definecolor{citepurple}{rgb}{0.288,0.1196,0.7}
\usepackage[backref=page]{hyperref}
\hypersetup{
     colorlinks = true,
     linkcolor = citepurple,
     anchorcolor = blue,
     citecolor = citepurple,
     filecolor = blue,
}

\usepackage[capitalize]{cleveref}
\crefname{section}{Section}{Sections}
\crefname{table}{Table}{Tables}
\usepackage[textsize=tiny]{todonotes}

\definecolor{green1}{rgb}{0.0, 1.0, 0.0} % Pure green
\definecolor{green2}{rgb}{0.3, 0.9, 0.0} % Green with a bit of red
\definecolor{green3}{rgb}{0.5, 0.8, 0.0} % More red, still mostly green
\definecolor{yellow1}{rgb}{0.7, 0.7, 0.0} % Transitioning to yellow
\definecolor{yellow2}{rgb}{0.9, 0.6, 0.0} % Darker yellow
\definecolor{yellow3}{rgb}{1.0, 0.5, 0.0} % Dark yellow, approaching orange

\definecolor{color100}{rgb}{0.0, 0.8, 0.0} % Green
\definecolor{color90}{rgb}{0.1, 0.7, 0.0}
\definecolor{color80}{rgb}{0.2, 0.6, 0.0}
\definecolor{color70}{rgb}{0.3, 0.5, 0.0}
\definecolor{color60}{rgb}{0.4, 0.4, 0.0} % Light Green
\definecolor{color50}{rgb}{0.5, 0.5, 0.0} % Yellow
\definecolor{color40}{rgb}{0.6, 0.4, 0.0} % Light Orange
\definecolor{color30}{rgb}{0.7, 0.3, 0.0}
\definecolor{color20}{rgb}{0.8, 0.2, 0.0}
\definecolor{color10}{rgb}{0.9, 0.1, 0.0}
\definecolor{color0}{rgb}{1.0, 0.0, 0.0} % Red

\definecolor{darkorange}{rgb}{1.0, 0.54, 0}
\newcommand{\webpage}{https://robotics-fm-survey.github.io/}

\author{
\normalsize {Yafei Hu}$^1$\footnote{Equal contribution. \footnotesize{ \tt \{yafeih, quantinx, vidhij\}@andrew.cmu.edu}} \hspace{3em}
\normalsize {Quanting Xie}$^1$\protect\footnotemark[1]\hspace{3em}
\normalsize {Vidhi Jain}$^1$\protect\footnotemark[1] \\
\begin{normalsize}
\begin{tabular}{lllll}
{Jonathan Francis}$^{1,2}$  &
{Jay Patrikar}$^1$ &  
{Nikhil Keetha}$^1$   &  
{Seungchan Kim}$^1$ &
{Yaqi Xie}$^1$   \\   
{Tianyi Zhang}$^1$ &  
{Hao-Shu Fang}$^3$ & 
{Shibo Zhao}$^1$ &
{Shayegan Omidshafiei}$^4$ & 
{Dong-Ki Kim}$^4$ ;\\
{Ali-akbar Agha-mohammadi}$^4$ &
{Katia Sycara}$^1$ &
{Matthew Johnson-Roberson}$^1$ &
{Dhruv Batra}$^{5,6}$ &
{Xiaolong Wang}$^7$ \\
{Sebastian Scherer}$^1$ &
{Chen Wang}$^8$ & 
{Zsolt Kira}$^5$ & 
{Fei Xia}$^9$\footnote{Equal advising. \footnotesize{\tt xiafei@google.com, ybisk@cs.cmu.edu} }& 
{Yonatan Bisk}$^{1,6}$\protect\footnotemark[2]  \\
\end{tabular}
\end{normalsize}\\[1.5em]
\begin{small}
$^1${CMU}\hspace{3pt}
\hspace{3pt}$^2${Bosch Center for AI}
\hspace{3pt}$^3${MIT}
\hspace{3pt}$^4${Field AI}
\hspace{3pt}$^5${Georgia Tech}
\hspace{3pt}$^6${FAIR at Meta}
\hspace{3pt}$^7${UC San Diego}
\hspace{3pt}$^8${SAIR Lab}
\hspace{3pt}$^9${Google DeepMind}
\end{small}
}

\title{ \LARGE \bf
Toward General-Purpose Robots via Foundation Models:\\ A Survey and Meta-Analysis
\\
\vspace{-10pt}}

\date{}

\begin{document}
\maketitle

\begin{figure} [ht!]
    \centering
    \vspace{-10mm}
    \includegraphics[width=\linewidth]{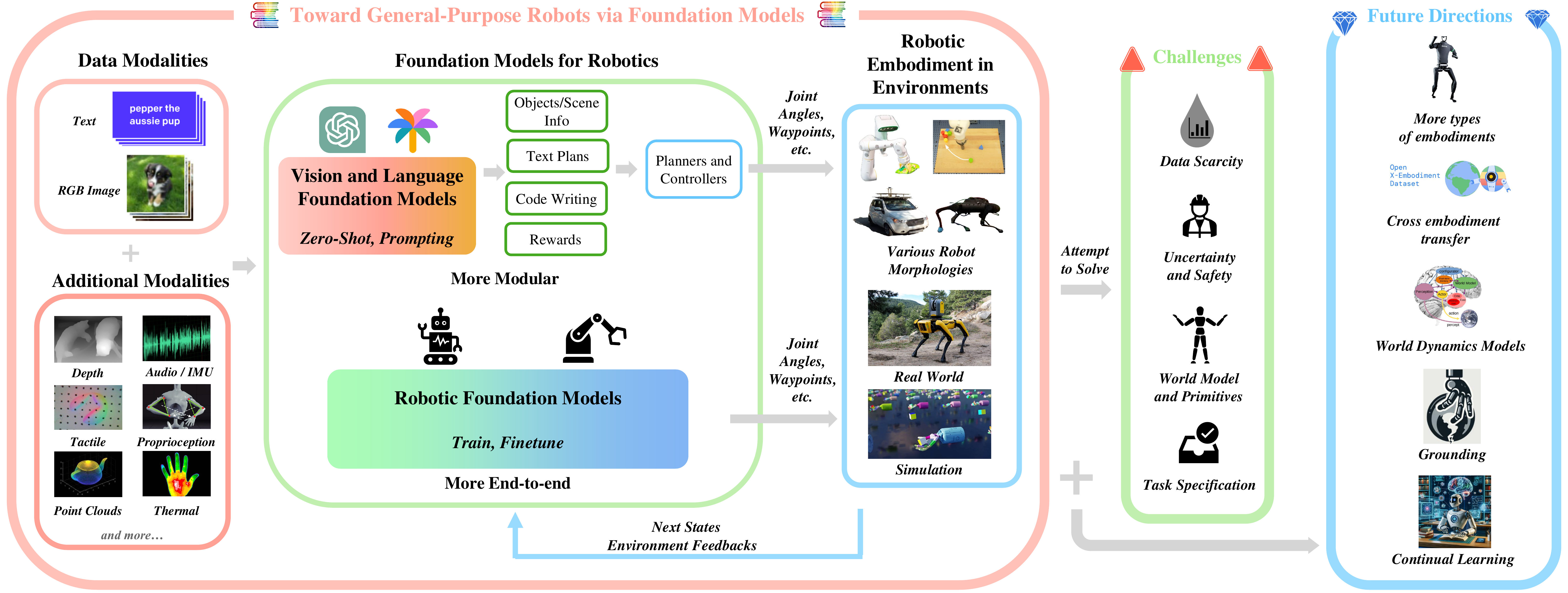}
    \vspace{-5mm}
    \caption{In this paper, we present a survey toward \emph{building general-purpose robots via foundation models}. We mainly categorize the foundation models into vision and language models used in robotics, and robotic foundation models. We also introduce how these models could mitigate the challenges of classical robotic challenges, and projections of the potential future research directions. \protect \footnotemark}
    \label{fig:teaser}
\end{figure}

\footnotetext{Some images in this paper are screenshots from the papers we surveyed, icon images from Microsoft PowerPoint and MacOS Keynote, Google Images results, or are images we generated with OpenAI GPT-4.}

\begin{abstract}
Building general-purpose robots that operate seamlessly in any environment, with any object, and utilizing various skills to complete diverse tasks has been a long-standing goal in Artificial Intelligence.
However, as a community, we have been constraining most robotic systems by designing them for specific tasks, training them on specific datasets, and deploying them within specific environments. These  systems require extensively-labeled data and task-specific models. When deployed in real-world scenarios, such systems face several generalization issues and struggle to remain robust to distribution shifts.
Motivated by the impressive open-set performance and content generation capabilities of web-scale, large-capacity pre-trained models (i.e., \textbf{foundation models}) in research fields such as Natural Language Processing (NLP) and Computer Vision (CV), we devote this survey to exploring (i) how these existing foundation models from NLP and CV can be applied to the field of general-purpose robotics, and also exploring (ii) what a robotics-specific foundation model would look like. We begin by providing a generalized formulation of how foundation models are used in robotics, and the fundamental barriers to making generalist robots universally applicable.
Next, we establish a taxonomy to discuss current work exploring ways to leverage existing foundation models for robotics and develop ones catered to robotics.
Finally, we discuss key challenges and promising future directions in using foundation models for enabling \textbf{general-purpose robotic systems}.
We encourage readers to view our living GitHub repository
% YF: hide now for TRO
\footnote{The current version of this paper is v2.1-2024.09 (In the format of `[major].[minor]-YYYY.MM').}
of resources, including papers reviewed in this survey, as well as related projects and repositories for developing foundation models for robotics:
\href{\webpage}{\textbf{\color{citepurple}{\webpage}}}.
\end{abstract}

% \keywords{foundation models, general-purpose robotics}

% \input{abstract}
% hide for the TRO submission version
\tableofcontents
\clearpage
\section{Overview} \label{sec:intro}
\subsection{Introduction}
We face many challenges in developing general-purpose robotic systems that can operate in and adapt to different environments and perform multiple tasks. Previous robotic perception systems that leverage conventional deep learning methods usually require a large amount of labeled data to train the supervised learning models \cite{Geiger2013KITTI, maturana2018offroad, Calli2017YCB}; meanwhile, the crowdsourced labeling processes for building these large datasets remain rather expensive. Moreover, due to the limited generalization ability of classical supervised learning approaches, the trained models usually require carefully designed domain adaptation techniques to deploy these models to specific scenes or tasks \cite{Donahue2014DeCAF, Tzeng2017ADDA}, which often require further steps of data-collection and labeling
% These methods often suffers from distributional shift and thus struggle to handle open-world perception tasks. 
% With the emergence of self-supervised vision learning \cite{he2019moco} \cite{He_2021_MAE} \cite{2021_Caron_DINO} and image-language pre-training \cite{Radford_2021_CLIP} \cite{Gu_2021_ViLD} \cite{Li_2022_LSeg}, we see the possibility of open-world visual recognition without supervised fine-tuning on task or scene-specific dataset. 
% this paragraph talks about the chanllenges on conventional plannings and control 
% Planning: Yafei and Seungchan
Similarly, classical robot planning and control methods often require carefully modeling the world, the ego-agent's dynamics, and other agents' behavior \cite{shen2020learning, kim2020learning, Williams2016MPPI}. These models are built for each individual environment or task and often need to be rebuilt as changes occur, exposing their limited transferability \cite{Williams2016MPPI};
% For instance, mapping of the environments, obstacle modelling for motion planning;  domain-dependent heuristics \cite{shen2020learning}, action-orderings \cite{garrett2016learning}, and state encoding \cite{kim2020learning} for task and motion planning (TAMP); dynamics model of the robot and cost function for action control.
in fact, building an effective model can be either too expensive or intractable.
Although deep (reinforcement) learning-based motion planning \cite{qureshi2019mpnet, fishman2022mpinets} and control methods \cite{2020_Peng_LearningAgileImitating, 2016_Levine_end2endDeepVisuomotor, 2019_Hwangbo_LearningAgile, 2019_Nagabandi_DeepDynamicsModels}
% which learn motion planning/control policies or world dynamics models. 
% these methods have been successfully deployed in real-world robotic applications, 
could help mitigate these problems, they also still suffer from distribution shifts and reductions in generalizability \cite{Kalashnkov2021MTOPT, jang2021bc}. 

% \clearpage
% put sec 1 in fig. 1
\begin{floatingfigure}[r]{0.45\textwidth}
    \centering
    \includegraphics[width=0.45\linewidth]{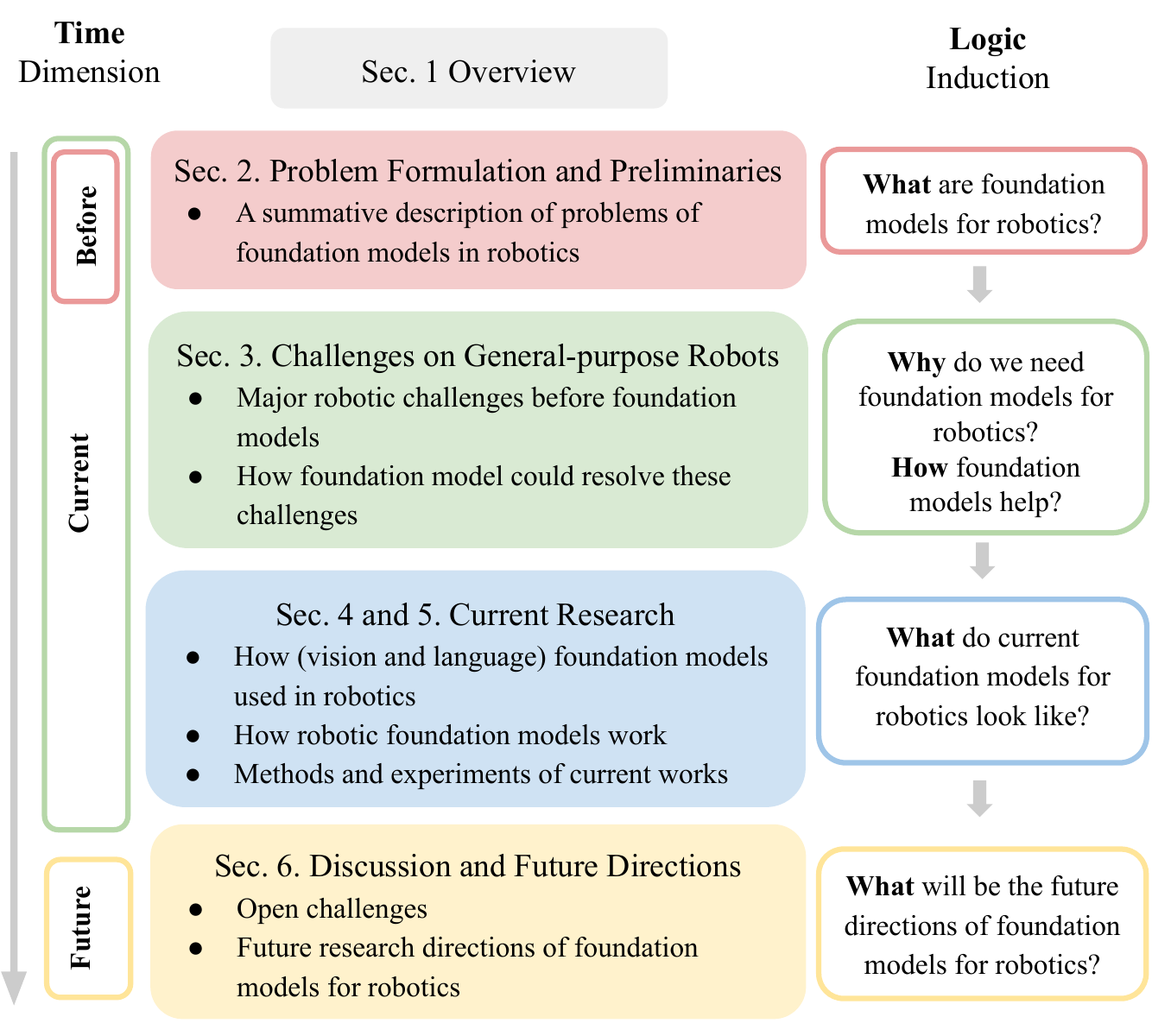}
    % \vspace{-1mm}
    \caption{Overall structure of this survey paper. The left side shows the time dimension governing the emergence of foundation models, and the right side shows the logical induction, for applying and developing foundation models for robotics. Sections \ref{sec:preliminary} and \ref{sec:challenges_robotics} answer the ``what" questions: what are foundation models for robotics, what are the challenges in building general-purpose robots. Section \ref{sec:current} and \ref{sec:experiments} deal with the ``why" and ``how" questions: why do we need foundation models in general-purpose robotics, how can foundation models be applied in robotics, and how do foundation models specially designed for robotics work. Section \ref{sec:discussion} closes with existing work, and posits what future models might look like.
    % \zk{There are a few issues in this figure: The left and right side don't align, e.g. Sec. 2 is preliminaries but next to it, it says ``What are the challenges''. Actually that's in the next part (Sec. 3). Same with other ones as well. ``What are future robotic foundation models look like'' is also repeated, and Sec 4 is about current research not future. Also I would change to ``What will future robotic foundation models look like?'' Yafei : Done}
    }
    \label{fig:overall_structure}
    % \vspace{3mm}
\end{floatingfigure}

% this paragraph talks about foundation models
Concurrent to the challenges faced in building generalizable robotic systems, we notice significant advances in the fields of Natural Language Processing (NLP) and Computer Vision (CV)---with the introduction of Large Language Models (LLMs) \cite{brown2020language} for NLP, diffusion models to generate high-fidelity images and videos  \cite{Ramesh2022DALLE2, Saharia2022ImaGen}, and zero-shot/few-shot generalization of CV tasks with large-capacity Vision Foundation Models (VFM) \cite{2021_Caron_DINO, Oquab_2023_DINOv2, kirillov2023segany}, Vision Language Models (VLMs) and Large Multi-Modal Models (LMM) \cite{Radford_2021_CLIP, Alayrac2022FlamingoAV}
% by vision-language pre-training \cite{Radford_2021_CLIP} 
% with self-supervised visual learning 
% In robotics, we have not seen the rapid growth as in natural language and computer vision, benefited by the large language model as well as vision foundation models respectively. 
, coined \textbf{foundation models} \cite{bommasani2021opportunities}.
% , or simply Large Pre-Trained Models (LPTMS), 

These large-capacity vision and language models have also been applied in the field of robotics \cite{Ahn_2022_saycan, Chen_2022_nlmapsaycan, tang2023saytap}, with the potential for endowing robotic systems with open-world perception, task planning, and even motion control capabilities. Beyond just applying existing vision and/or language foundation models in robotics, we also see considerable potential for the development of more robotics-specific models, e.g., the action model for manipulation \cite{brohan2022rt, 2023rt2}, motion planning model for navigation \cite{Shah_2023_ViNT}, and more generalist models which can conduct action generation though vision and language inputs hence Vision Language Action models (VLA) \cite{Reed_2022_Gato, driess2023PaLMe, 2023rt2, kim2024openVLA}. These robotics foundation models show potential generalization ability across different tasks and even embodiments. 
% Vision/language foundation models have also been applied directly to robotic tasks \cite{Reed_2022_Gato, driess2023PaLMe}, showing the possibility of fusing different robotic modules into a single unified model.

% the overall structure of the paper
The overall structure of this paper is formulated as in Figure \ref{fig:overall_structure}. In Section \ref{sec:preliminary}, we provide a brief introduction to robotics research before the foundation model era and discuss the basics of foundation models. In Section \ref{sec:challenges_robotics}, we enumerate challenges in robotic research and discuss how foundation models might mitigate these challenges. In Section \ref{sec:current}, we summarize the current research status quo of foundation models in robotics. Finally, in Section \ref{sec:discussion} we offer potential research directions which are likely to have a high impact on this research intersection.

Although we see promising applications of vision and language foundation models to robotic tasks, and the development of novel robotics foundation models, many challenges in robotics out of reach. From a practical deployment perspective, models are often not reproducible, lack multi-embodiment generalization, or fail to accurately capture what is feasible (or admissible) in the environment. Furthermore, 
% from a research direction perspective,
most publications leverage transformer-based architectures and focus on semantic perception of objects and scenes, task-level planning, or control \cite{2023rt2}; other components of a robotic system, which could benefit from cross-domain generalization capabilities, are under-explored---e.g., foundation models for world dynamics or foundation models that can perform symbolic reasoning.
% In addition, one lesson we learned from these foundation models is that, we still need tremendous amount of dataset. Thus, 
Finally, we would like to highlight the need for more large-scale real-world data as well as high-fidelity simulators that feature diverse robotics tasks.

In this paper, we investigate how foundation models serve as a potential solution for general-purpose robotics, aiming to understand how foundation models could help mitigate the core challenges general-purpose robotics face. We use the term ``\textit{\textbf{foundation models for robotics}}" to include two distinct aspects: (1) the application of existing (mainly) vision and language models \textbf{to} robotics, largely through zero-shot and in-context learning; and (2) developing and leveraging \textbf{robotics foundation models} especially for robotic tasks by using robot-generated data. We summarize the methodologies of foundation models for robotics papers and conduct a meta-analysis of the experimental results of the papers we surveyed. A summary of the major components of this paper in Figure \ref{fig:teaser}.
% \zkn{After having read the whole thing, it would be nice to have more structured organization in the intro aligned with the main things in the document (e.g. FMs for robotics and RFMs, etc.) with bolded text. Also would be good to emphasize 1) Major contributions (e.g. this decomposition, the meta-analysis, etc.) and 2) Major take-aways. } Yafei: Done

% To have a more systematic understanding of the structure and components of this paper, we include the taxonomy of this paper in Figure \ref{fig:taxonomy}.
%\clearpage
% also talk about this in Sec. 1
\subsection{Related Survey Papers}
\label{related_work}
% \subsubsection{robotics survey papers} Previous survey papers in the field of robotics focus on various methodologies, e.g. RL for robotics \cite{2013RLRoboticsSurvey, Ibar2021How}; specific robotic tasks, e.g. manipulation \cite{2019ManipulationSurvey}, social navigation \cite{francis2022core, Francis2023Principles}, and safety in robotics \cite{survey_safeHRI_MIT}.

% \subsubsection{Foundation model survey papers} 
Recently, with the popularity of foundation models, there are various survey papers on vision and language foundation models that are worth mentioning \cite{Kaddour2023Challenges, Zhang2023Text, 2023Harnessing, Yang2023SurveyFMDecision}. These survey papers cover foundation models, including Vision Foundation Models (VFMs) \cite{zhang2023survey, awais2023foundational}, Large Language Models (LLMs) \cite{2023Harnessing}, Vision-Language Models (VLMs) \cite{Du2022ASurvey, Gu2023ASystematic}, and Visual Content Generation Models (VGMs) \cite{Kaddour2023Challenges}. There are a few existing survey papers combining foundation models and robotics, perhaps the most relevant survey papers are \cite{firoozi2023foundation, Yang2023SurveyFMDecision, Wang2023SurveyLLMAutonomous, Lin2023LLMsEmbodiedNavigation, majumdar2023robotics, xiao2023robot}, however, there are still significant differences between those papers and ours, for instance: Yang~\textit{et al.}  \cite{Yang2023SurveyFMDecision} and Wang~\textit{et al.} \cite{Wang2023SurveyLLMAutonomous} focus on broadly-defined autonomous agents, instead of physical robots; Lin~\textit{et al.}~\cite{Lin2023LLMsEmbodiedNavigation} focus on LLMs for navigation; the connection between foundation models and robotics is limited in~\cite{majumdar2023robotics}. Compared with ~\cite{xiao2023robot}, we propose a breakdown of current research methodologies, providing a detailed analysis of the experiments, and also focus on how foundation models could resolve the typical challenges for general-purpose robotics. Concurrently, Firoozi \textit{et al.}~\cite{firoozi2023foundation} conducted a survey regarding foundation models in robotics. Both their and our works shed lights on the opportunities and challenges of using foundation models in robotics, and identifies key pitfalls to scale them further. Their work focuses on how foundation models contribute to improving robot capabilities, and challenges going forward. Comparatively, our survey attempts to taxonomize robotic capabilities together with foundation models to advance those capabilities. We also propose a dichotomy between robotic(-first) foundation models and other foundation models used in robotics, and provide a thorough meta-analysis of the papers we survey.

\looseness = -1
In this paper, we provide a survey that includes existing unimodal and multimodal foundation models applied in robotics, as well as all forms of robotics foundation models in various robotics tasks as we know of. We also narrowed the scope of papers being surveyed to only those with experiments on real physical robotics, in high-fidelity simulation environments, or using real robotics datasets. We believe that, by doing so, it could help us understand the power of foundation models in the real world robotic applications.

% \clearpage
\section{Problem Formulation and Preliminaries}  \label{sec:preliminary}

In this section, we summarize foundation models for robotics as a unified problem formulation. Let us define the foundation model relevant for robotics as a function $f$, which takes sensory inputs $\mathbf{x_t}$, context $\mathbf{c}$, and outputs $\mathbf{y_t}$ to fulfill the downstream robotic tasks. 
\begin{equation}
    f(\mathbf{x_{tk}, c_k}) \rightarrow \mathbf{y_{tk}} \quad\quad \quad  \forall k \in N \quad \forall t \in T
\end{equation}
Here $\mathbf{x}$ denotes sensory inputs like observations through visual cameras, textual scene descriptions, scene graphs, object poses and detections, audio signals through a microphone,  haptic sensors, and so on. For multi-task policies, we need contextual information $\mathbf{c}$ that denotes task specification~\cite{2023rt2} or details on the embodiment~\cite{arenas2023how}. 

The state $\mathbf{x}$ and context $\mathbf{c}$ are often defined in terms of visual observations and language instructions respectively. However, both $\mathbf{x}$ and $\mathbf{c}$ could be images, videos, language prompts, or a combination of multiple modalities. This also depends on how we define the scope of the environment.

The action $\mathbf{y}$ can be defined in terms of target objects' pose in the map, task plans, next state, reward functions, and control outputs like target end-effector pose. If $\mathbf{y}$ is a task plan, we often assume that the low-level components of the plan can be executed perfectly. However, there are open questions on how to enable suitable and timely recovery in the robot policies when the low-level plan is not executed completely. 

The data to train or prompt the model $f$ have several open questions. For training, how should expert demonstrations, offline data, or online experiences with apt reward functions be collected? For prompting, what instructions and examples enable desirable in-context learning? How should the models be trained and fine-tuned? In the next following sections, we aim to answer these questions by summarizing and analyzing existing research works.

\section{Challenges on General-purpose Robots}
\label{sec:challenges_robotics}

\begin{figure*}[ht]
\resizebox{\linewidth}{!}{%
\centering
    \begin{tikzpicture}[
    edge from parent fork down,
    level distance=2cm,
    every node/.style={fill=green!20, rounded corners, align=center, minimum height=0.8cm},
    edge from parent/.style={-Stealth, thick, draw=black!100},
    level 1/.style={sibling distance=63mm, level distance=25mm},
    level 2/.style={sibling distance=22mm}
    % level 3/.style={sibling distance=1mm}
    ]
    
    % Root node
    \node [minimum width=3cm, minimum height=1cm] {Challenges in Robotics (\ref{sec:challenges_robotics})}
    % Children and grandchildren
    child { node [minimum width=3cm, minimum height=1cm] {Generalization (\ref{challenge_generalization})}  
            child { node [minimum width=1.5cm,minimum height=1cm] {perception}}  
            child { node  [minimum width=1.5cm,minimum height=1cm] {morphology}}  
            }
    child { node [minimum width=3cm, minimum height=1cm] {Data Scarcity (\ref{challenge_data})} 
        child { node [minimum width=1.5cm,minimum height=1cm] {simulation}}  
        child { node [minimum width=1.5cm,minimum height=1cm] {real-world}}  
            }
    child { node [minimum width=3cm, minimum height=1cm] {Requirements of Models (\ref{challenge_model})} 
        child { node [minimum width=1.5cm,minimum height=1cm] {model-based}}  
        child { node [minimum width=1.5cm,minimum height=1cm]  {model-free}}  
        }
    child { node [minimum width=3cm, minimum height=1cm] {Task Specification (\ref{challenge_task_specification}) }  
        child  { node [minimum width=1.5cm, minimum height=1cm] {goal \\ image}}  
        child { node [minimum width=1.5cm, minimum height=1cm] {reward \\ generation}}  
        child { node [minimum width=1.5cm, minimum height=1cm] {language \\ prompt}} 
        }
    child { node [minimum width=3cm, minimum height=1cm] {Uncertainty and Safety (\ref{challenge_safety}) } 
        child { node [minimum width=1.5cm, minimum height=1cm] {epistemic}}  
        child{ node [minimum width=1.5cm, minimum height=1cm] {aleatoric}}  
        child{ node [minimum width=1.5cm, minimum height=1cm] {provable \\ safety}}  
        };
    \end{tikzpicture}
    }
    \caption{Taxonomy of the challenges in robotics that could be resolved by foundation models. We list five major challenges in the second level and some, but not all, of the keywords for each of these challenges.
    % \zk{I would order task specification by most specific on left (goal image) ttoleast specific (language prompt). Of course, align narrative later with whatever is in the figure too. Yafei: Done}
    }
    \label{taxonomy_challenges}
\end{figure*}
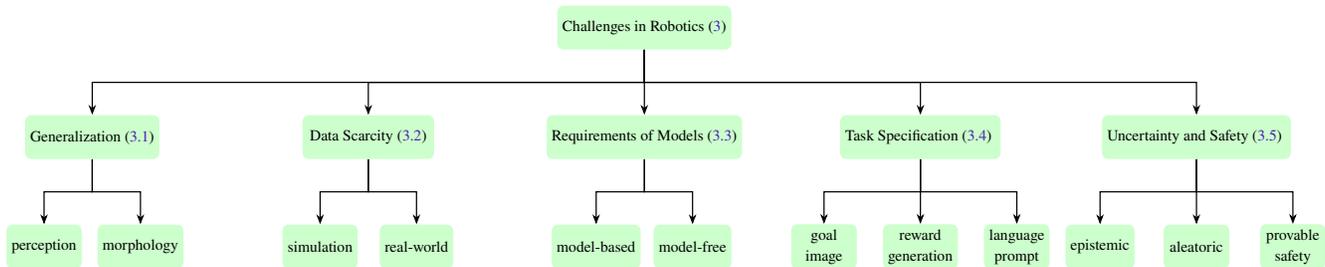

In this section, we summarize 5 core challenges that general-purpose robotic system face, each detailed in the following subsections. Whereas similar challenges have been discussed in prior literature (Section \ref{related_work}), this section mainly focuses on the challenges that may potentially be solved by appropriately leveraging foundation models, given the evidence from current research results. We also depict the taxonomy in the section for easier review in Figure~\ref{taxonomy_challenges}.

% \subsubsection{Robustness and Generalization}
\subsection{Generalization}
\label{challenge_generalization}
%Classical MPC controller and geometric motion planning often 
Robotics systems often struggle with accurate perception and understanding of their environment. Limitations in computer vision, object recognition, and semantic understanding made it difficult for robots to effectively interact with their surroundings. Traditional robotics systems often relied on analytic hand-crafted algorithms, making it challenging to adapt to new or unseen situations. They also lacked the ability to generalize their training from one task to another, further limiting their usefulness in real-world applications. This generalization ability is also reflected in terms of the generalization of planning and control in different tasks, environments, and robot morphologies. For example, specific hyperparameters for, e.g., classical motion planners and controllers need to be tuned for specific environments \citep{huang2023went,herman2021learn,francis2022learn}; RL-based controllers are difficult to  transfer across different tasks and environments \citep{francis2022core,francis2022knowledge,yenamandra2024towards}. In addition, due to differences in robotic hardware, it is also challenging to transfer models across different robot morphologies \citep{tatiya2023transferring,tatiya2023crosstool}. By applying foundation models in robotics, the generalization problem is partially resolved, which will be discussed in the next Section~\ref{sec:current}. Further challenges, such as generalization across different robotic morphologies, remain demanding.
% Foundation models have enabled open-set robotic perception, allowing robots to better understand and navigate their environment.
% By leveraging transfer learning and multi-modal learning, foundation models have allowed robots to generalize their learning across different tasks and domains. This has led to increased adaptability and robustness in the face of new challenges.

% In summary, some of the main challenges for the robotics community are:
% Research Reproducibility
% Hardware-Reality Integration
% Multi-Embodiment Generalization
% Collaboration with Other Fields

\subsection{Data Scarcity}
\label{challenge_data}
\begin{figure*}[t]
    \centering
    \includegraphics[width=\linewidth]{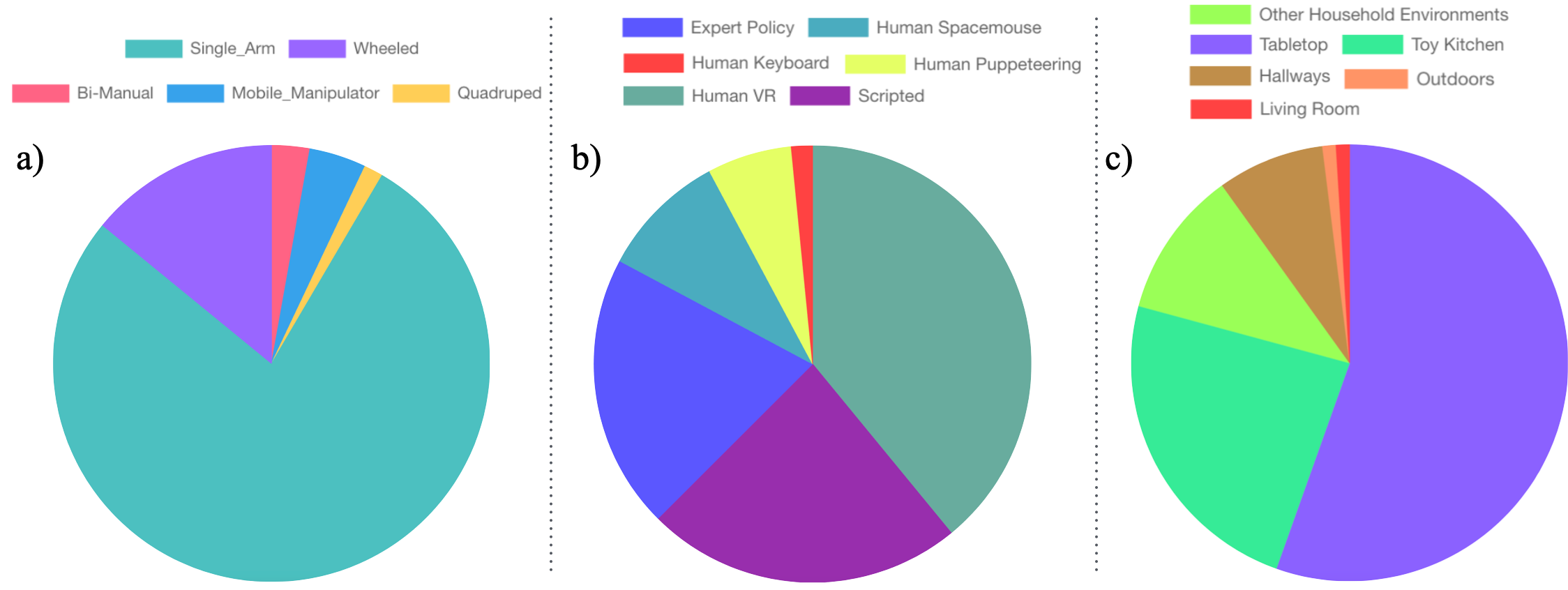}
    \caption{\justifying Comprehensive visualizations of the Open-X Embodiment and Droid Dataset encompassing robot morphologies, environment types, and data collection methods.}
    \label{fig:dataset_fig}
\end{figure*}

Data has always been the cornerstone of learning-based robotics methods. The need for large-scale, high-quality data is essential to develop reliable robotic models. Several endeavors have been attempted to collect large-scale datasets in the real world, including autonomous driving~\cite{Geiger2013KITTI,maturana2018offroad,Sun_2020_CVPR}, robot manipulation trajectories~\cite{2018_Kalashnikov_QT-Opt, 2016_Levine_LearningHandEye,rlscale2023rss}, etc. Collecting robot data from human demonstration is expensive~\cite{brohan2022rt}. The diverse range of tasks and environments where robots are used even complicates the process of collecting adequate and extensive data in the real world. Moreover, gathering data in real-world settings can be problematic due to safety concerns \cite{huang2023went}. To overcome these challenges, many works~\cite{todorov2012mujoco, makoviychuk2021isaac, mittal2023orbit, savva2019habitat, szot2022habitat, puig2023habitat} attempt generating synthetic data in simulated environments. These simulations offer realistic virtual worlds where robots can learn and apply their skills to real-life scenarios. Simulations also allow for domain randomization, as well as the potential to update the parameters of the simulator to better match the real world physics \cite{huang2023went}, helping robots to develop versatile policies. However, these simulated environments still have their limits, particularly in the diversity of objects, making it difficult to apply the learned skills directly to real-world situations. % \textbf{Real-World Data Collection} 
Collecting real-world robotic data with a scale comparable to the internet-scale image/text data used to train foundation models is especially challenging. One promising approach is collaborative data collection across different laboratories and robot types~\cite{embodimentcollaboration2023open, fang2024rh20t, shafiullah2023bringing}, as shown in Figure~\ref{fig:dataset_fig}.
However, an in-depth analysis of the Open-X Embodiment Dataset reveals certain limitations regarding data type availability. 
First, in Figure 4a, the robot morphology utilized for data collection is restrictive; out of the 73 datasets, 55 are dedicated to single-arm manipulation tasks. Only one dataset pertains to quadruped locomotion, and a single dataset addresses bi-manual tasks. Second, in Figure 4b, the predominant scene type for these manipulation tasks is tabletop setups, often employing toy kitchen objects. Such objects come with inherent assumptions including rigidity and negligible weight, which may not accurately represent a wider range of real-world scenarios. Third, in Figure 4c, our examination of data collection methods indicates a predominance of human expert involvement, predominantly through virtual reality (VR) or haptic devices. This reliance on human expertise highlights the challenges in acquiring high-quality data and suggests that significant human supervision is required. For instance, the RT-1 Robot Action dataset necessitated a collection period of 17 months, underscoring the extensive effort and time commitment needed for data accumulation with human involvement. 
% Despite this, the available data, mainly from fixed-base manipulators, are limited in diversity and primarily focused on indoor environments, neglecting outdoor challenges.

% \begin{figure}[ht!]
%     \centering 
    % \begin{subfigure}[t]{1.0\linewidth}
    %     % \vspace{-60pt} % Adjust the negative value as needed to move the subfigure up
    %     \includegraphics[width=\linewidth]{figs/rt-x_figure.png}
    %     \caption{Joint Effort in Real-World Data Collection}
    %     \label{fig:rtx_structure}
    % \end{subfigure}
    % \hfill % add some horizontal spacing
    % \begin{subfigure}[b]{0.7\linewidth}
    %     \includegraphics[width=\linewidth]{figs/dataset_morphology.png}
    %     \caption{Dataset Morphology Distribution}
    %     \label{fig:collect_method}
    % \end{subfigure}
    % \newline % to break the line and put the next two figures on the next line
    % \begin{subfigure}[b]{0.7\linewidth}
    %     \includegraphics[width=\linewidth]{figs/dataset_env_types.png}
    %     \caption{Environment Type Classification}
    %     \label{fig:morphology}
    % \end{subfigure}
    % \hfill % add some horizontal spacing
    % \begin{subfigure}[b]{0.7\linewidth}
    %     \includegraphics[width=\linewidth]{figs/dataset_collect_method2.png}
    %     \caption{Data Collection Methods}
    %     \label{fig:scene_type}
    % \end{subfigure}
    % \caption{\justifying Comprehensive visualizations of the Open-X Embodiment Dataset encompassing robot morphologies, environment types, and data collection methods,}
    % % \vspace{-1em}
    % \label{fig:combined_figures}
% \end{figure}
%% moved from sec 3

\subsection{Requirements of Models and Primitives}
\label{challenge_model}

% Traditional robotics involves designing control algorithms based on physical models and optimization techniques to achieve desired tasks. These methods are often deterministic and rely on well-defined rules and heuristics.
% On the other hand, ML techniques, particularly in the context of robotics, can be used for various aspects of Computer Vision, Reinforcement Learning, and Language modeling for planning, control and communicating with people.

\looseness = -1
Classical planning and control methods usually require carefully engineered models of the environment and the robot. Optimal control methods require good dynamics models (i.e., world transition models) \cite{Williams2016MPPI, arcari2023bayesian}; motion planning requires a map of the environment~\cite{Cao2021TARE}, the states of the objects robots interact with~\cite{islam2020provably}, or a set of pre-defined motion primitives~\cite{Saxena_2021}; task planning requires pre-computed object classes and pre-defined rules~\cite{Garrett2010InteTAMP,curtis2022long}, etc. Previous learning-based methods (e.g., imitation and reinforcement learning) train policies in an end-to-end manner that directly gets control outputs from sensory inputs~\cite{2018_Kalashnikov_QT-Opt}, avoiding building and using models. These methods partially solve this problem of relying on explicit models, but they often struggle to generalize across different environments and tasks. This raises two problems: (1) How can we learn model-agnostic policies that can generalize well? Or, alternatively, (2) How can we learn good world models so that we can apply classical model-based approaches? We see some recent works that aim to resolve these problems using foundation models (especially in a model-free manner), which will be systematically discussed in Section~\ref{sec:current}. 
However, the call for world models for robotics remains an intriguing frontier, which will be discussed in Section~\ref{sec:discussion}.

\subsection{Task Specifications}
\label{challenge_task_specification}
% vidhi 
Understanding the task specification and grounding it in the robot's current understanding of the world is a critical challenge to obtaining generalist agents. Often, these task specifications are provided by users with limited understanding of the limitations on the robot's cognitive and physical capabilities. This not only raises questions about what the best practices are for providing these task specifications, but also about the naturalness and ease of crafting these specifications. Understanding and resolving ambiguity in task specifications, conditioned on the robot's understanding of its own capabilities, is also challenging. Foundation models are a promising solution for this challenge: task specification can be formulated as language prompts \cite{brohan2022rt, 2023rt2, Ahn_2022_saycan, liu2024okrobot}, goal images~\cite{cui2022zest}, rewards for policy learning~\cite{tang2023saytap, ma2023eureka}, etc. 
% For details, please refer to Section~\ref{sec:current}.

% As the most natural
% way to describe a task can vary depending on the user, environment, or task, robotic foundation
% models for task specification should accept a variety of description modalities, such as goal states
% [Fu et al. 2018; Singh et al. 2019], natural language [MacGlashan et al. 2015; Karamcheti et al. 2017;
% Misra et al. 2017b; Co-Reyes et al. 2019; Shao et al. 2020], videos of humans [Shao et al. 2020; Chen
% et al. 2021c; Liu et al. 2018], pairwise or ranking comparisons [Biyik and Sadigh 2018], interactive
% corrections [Co-Reyes et al. 2019; Karamcheti et al. 2020] and physical feedback [Ross et al. 2011;
% Bajcsy et al. 2017].

\subsection{Uncertainty and Safety}
\label{challenge_safety}
\looseness = -1
One of the critical challenges in deploying robots in the real world comes from dealing with the uncertainty inherent in the environments and task specifications. Uncertainty, based on its source, can be characterized either as epistemic (uncertainty caused by a lack of knowledge) or aleatoric (noise inherent in the environment). Epistemic uncertainty often manifests as out-of-distribution errors when the robot encounters unfamiliar situations in the test distribution. While the adoption of learning-based techniques for decision-making in high-risk safety-critical fields has prompted efforts in uncertainty quantification (UQ) and mitigation \cite{gawlikowski2023survey}, out-of-distribution detection, explainability, interpretability, and vulnerability to adversarial attacks remain open challenges.          
Uncertainty quantification can be prohibitively expensive and may lead to sub-optimal downstream task performance \cite{li2022interpretable}. Given the large-scale over-parameterized nature of foundation models, providing UQ methods that preserve the training recipes with minimal changes to the underlying architecture are critical in achieving the scalability without sacrificing the generalizability of these models. Designing robots that can provide reliable confidence estimates on their actions and in turn intelligently ask for clarification feedback remains an unsolved challenge \cite{chi2020just}. Conformal predictions \cite{angelopoulos2021gentle} provide a distribution-free way of generating statistically rigorous uncertainty sets for any black-box model and have been demonstrated in VLN tasks for robotics \cite{ren2023robots}. 

In its traditional setting, provable safety in robotics \cite{wabersich2023data,hsu2023safety} refers to a set of control techniques that provide theoretical guarantees on safety bounds for robots. Control Barrier Functions \cite{ames2019control}, reachability analysis \cite{bansal2017hamilton,chen2021safe} and runtime monitoring via logic specifications \cite{leung2023backpropagation} are well-known techniques in ensuring robot safety with bounded disturbances. Recent works have explored the use of these techniques to ensure safety of the robot \cite{gu2022review}. While these contributions have led to improved safety, solutions often result in sub-optimal behavior and impede robot learning in the wild \cite{dawson2023safe}. Thus, despite recent advances, endowing robots with the ability to learn from experience to fine-tune their policies while remaining safe in novel environments still remains an open problem.

% \textcolor{red}{An example of the exact challenge: like a robot picking up dishes of cat and put it into the dishwasher}
% Suggestions from Zsolt
% How foundation models could help with these chanllenge? perception OOD & mis-claibration, control constraints, etc. 

% \subsubsection{Real-time capabilities and Certifiability}

% [Do we want to look at this?][SWAP-C, Edge AI, training cost? Stakeholders and policy implications of using foundation models?]

% Maybe move this paer to section 5?
% \subsubsection{Risks, Ethics and Security}
% A potential pitfall when applying foundation models to robotics is the disparity that can arise when a model is trained in one geographical region and then implemented in another. For instance, laws, including those documented in legal texts, vary between countries. Consequently, robots employed in various countries may require specific training aligned with the local laws. This becomes particularly crucial in activities like deploying self-driving cars, where adherence to local traffic laws and regulations is not just a matter of legal compliance, but also of safety. 
%A taxonomy of the important concepts in this section is presented in \ref{fig:foundation_model_taxonomy}. We also show the timeline of represented papers based on the release dates (mostly on arXiv) in Figure \ref{fig:timeline}.

\section{Review of Current Research Methodologies} \label{sec:current}

In this section, we summarize the current research methodologies of foundation models for robotics. 

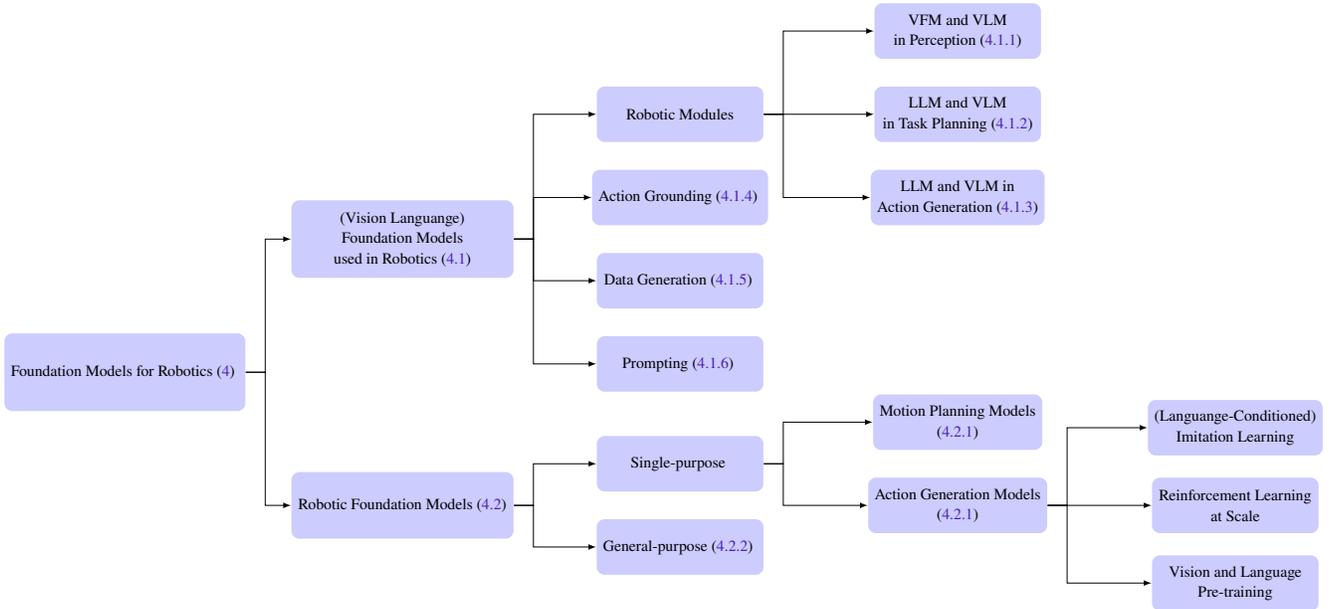
\begin{figure*}
    % \centering
    % \includegraphics[width=0.9\linewidth]{figs/method_overview.pdf}
% \begin{figure}
%         \centering
%         \includegraphics[width=0.9\linewidth]{figs/taxonomy_concept.png}
%         \caption{Enter Caption}
%         \label{fig:enter-label}
%     \end{figure}
\resizebox{\linewidth}{!}{%
\centering
\begin{tikzpicture}[
    nodes={anchor=base},
    grow=right, % Set the direction of tree growth
    % edge from parent fork down,
    level 1/.style={sibling distance=48mm, level distance=50mm},
    level 2/.style={sibling distance=15mm, level distance=50mm},
    level 3/.style={sibling distance=15mm, level distance=50mm},
    level 4/.style={sibling distance=14mm, level distance=50mm},
    every node/.style={rectangle, rounded corners, align=center, fill=blue!20, font=\footnotesize, minimum width=2cm, minimum height=1cm,  font=\footnotesize},
    edge from parent/.style={draw, -latex},
    edge from parent path={(\tikzparentnode.east) -- ++(10pt,0) |- (\tikzchildnode.west)} % This specifies the path to have right angles
]
% Root node
\node[minimum width=4cm, minimum height=1.4cm]{Foundation Models for Robotics (\ref{sec:current}) }
  child { node [minimum width=4cm,minimum height=1.2cm]   {Robotic Foundation Models (\ref{subsec:rfm})}
    child { node[minimum width=3cm, minimum height=1cm] {General-purpose (\ref{subsec:general_purpose}) } } 
    child { node[minimum width=3cm, minimum height=1cm] {Single-purpose } 
        child { node[minimum width=3cm, minimum height=1cm] {Action Generation Models \\ (\ref{rfm_action})} 
            child { node[minimum width=3cm, minimum height=1cm] {Vision and Language \\ Pre-training  } } 
            child { node[minimum width=3cm, minimum height=1cm] {Reinforcement Learning \\ at Scale  } } 
            child { node[minimum width=3cm, minimum height=1cm] {(Languange-Conditioned) \\ Imitation Learning} }
            }
        child { node[minimum width=3cm, minimum height=1cm] {Motion Planning Models \\ (\ref{rfm_motionplanning})} }
    }
  }
  child { node[minimum width=4cm, minimum height=1.4cm] {(Vision Languange) \\ Foundation Models \\ used in Robotics (\ref{subsec:vlfm_robo})}
    child { node[minimum width=3cm, minimum height=1cm] {Prompting (\ref{subsec:special_prompting}) } }
    % child { node[minimum width=3cm, minimum height=1cm] {Solve Task Specification \\ (\ref{subsec:special_task_specific}) } }
    child { node[minimum width=3cm, minimum height=1cm] {Data Generation (\ref{subsec:special_data_generation}) } }
    child { node[minimum width=3cm, minimum height=1cm] {Action Grounding (\ref{subsec:grounding}) } }
    child {node[minimum width=3cm, minimum height=1cm] {Robotic Modules}
        child { node[minimum width=3cm, minimum height=1cm] {LLM and VLM in \\  Action Generation (\ref{subsec:llm_action}) } }
        child { node[minimum width=3cm, minimum height=1cm] {LLM and VLM \\ in Task Planning (\ref{subsec:llm_vlm_planning}) } }
        child { node[minimum width=3cm, minimum height=1cm] {VFM and VLM \\ in Perception (\ref{vfm_vlm_perception})} }
    }
  };
\end{tikzpicture}
}
\caption{\justifying Conceptual Framework of Foundation Models in Robotics: The figure illustrates a structured taxonomy of foundational models, categorized into two primary segments: the application of existing foundation models (vision and language models) to robotics, and the development of robotic-specific foundation models. This includes distinctions between vision and language models used as perception tools, in planning, and in action, as well as the differentiation between single-purpose and general-purpose robot foundation models.
% \zk{Great breakdown! The only comment is that for action generation should we generalize beyond LLMs? E.g. I don't see anything in the whole survey about diffusion policies. Should we add that? Yafei: I think diffusion policy is not yet considered as foundation models?}
}
\vspace{-1em}
\label{fig:foundation_model_taxonomy}
\end{figure*}

In Section \ref{subsec:vlfm_robo}, we mainly discuss foundation models for robotics in two categories: \textbf{{Foundation Models used in Robotics}} and \textbf{Robotics Foundation Models} (RFMs). For Foundation Models used in Robotics, we mainly highlight applications of \textbf{vision and language} foundation models used in a \textit{zero-shot} manner, meaning no additional fine-tuning or training is conducted. In Section \ref{subsec:rfm}, however, we mainly focus on Robotics Foundation Models, wherein these approaches may warm-start models with vision-language pre-trained initialization and/or directly train the models on \textbf{robotics datasets}. Figure \ref{fig:foundation_model_taxonomy} shows the detailed taxonomy of this section. 

% As introduced in the Section \ref{sec:preliminary} (Preliminaries), . 

We review the methods presented in these papers following the convention of a typical robotic system which consists of perception, planning, and control modules. Here, we combine motion planning and control into one piece---action generation and treat motion planning modules as higher-level and control as lower-level action generation.
% Foundation models currently could empower many of these modules, we will discuss these empowerment in the following parts based on the major contributions. 
It is important to notice that although most of the works use foundation models in different functionality modules of the robotic systems, we will classify these papers into categories based on the module to which the paper contributes the most. There are, however, certain applications of the vision and language foundation models that go across these robotics modules, e.g., grounding of these models in robotics, and generating data from LLMs and VLMs. Given the autoregressive nature of current LLMs, they often grapple with extended horizon tasks. Thus, we also delve into advanced prompting methods proposed in the literature to ameliorate this limitation and enhance planning power. We list these applications in sections \ref{subsec:grounding}, \ref{subsec:special_data_generation} and \ref{subsec:special_prompting}, as a different perspective to analyze these works. 

We find that works in Section \ref{subsec:vlfm_robo} typically follow a modular strategy, in applying vision and language foundation models to serve a single robot functionality, e.g., applying VLMs as open-set robot perception modules which are then ``plugged in" to work alongside motion planners and controllers~\cite{Chen_2022_nlmapsaycan}, downstream. Since such foundation models are applied in a zero-shot manner, there are no gradients flowing between the module in which the foundation models are applied and the other modules in the robotic system. Conversely, works in Section \ref{subsec:rfm} mostly follow an end-to-end differentiability paradigm, which blurs the boundary of the typical robotics modules in methods (described in Section \ref{subsec:vlfm_robo}; e.g., perception and control~\cite{brohan2022rt, bousmalis2023robocat}), with some robotics foundation models even providing a unified model to perform different robot functions~\cite{Reed_2022_Gato, driess2023PaLMe}.

\begin{figure*}
    \centering
    \includegraphics[width=\linewidth]{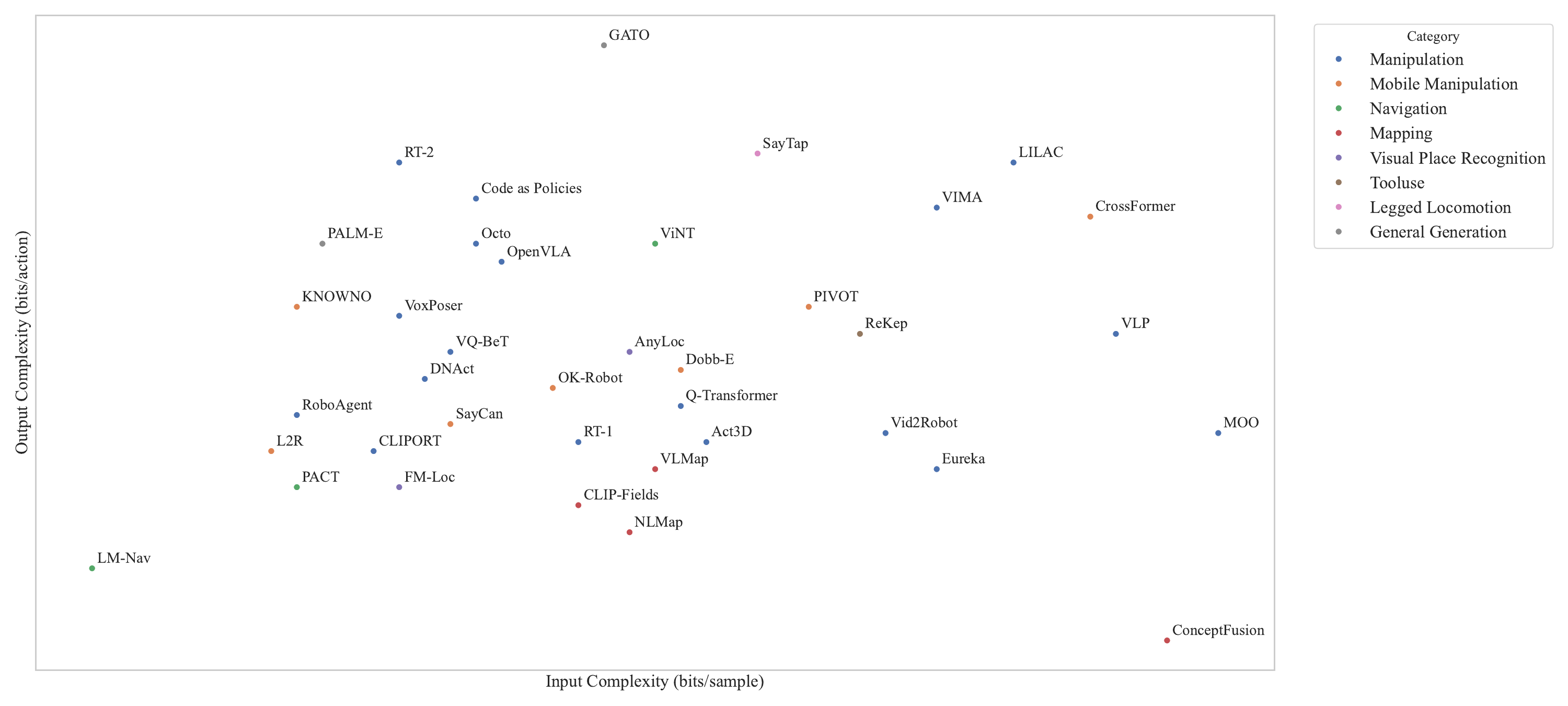}
    % {figs/category_analysis_impact.pdf}
    \caption{\justifying  Here we plot some representative works of foundation models used in robotics, and robotic foundation models. The horizontal axis represents the complexity of input data; the vertical axis represents the complexity of output action space. The complexity of the input data is brought by the modality of the data, e.g. image data is more complex than the text data. The complexity of output action space is mainly determined by the output dimension of the foundation model in the robotic tasks. }
    \label{fig:all_curr_works}
\end{figure*}

% \subsection{Research }
% \label{subsec:ResearchMethods}

% \textcolor{red}{make this figure into two. Distinguish based on imapct}
% \begin{figure} []
%     \centering
%     \includegraphics[width=0.95\textwidth]{figs/timeline.pdf}
%     \caption{Timeline of representative works of foundation models for robotics based on their release dates.}
%     \label{fig:timeline}
% \end{figure}

\subsection{Foundation Models used in Robotics}. \label{subsec:vlfm_robo}
In this section, we focus on \textit{zero-shot} applications of vision and language foundation models in robotic tasks, which include mainly zero-shot deployment of VLMs used in robotic perception, in context learning of LLMs for task-level and motion-level planning, as well as action-generation. 

\subsubsection{VFMs and VLMs in Robot Perception}. \label{vfm_vlm_perception}
Recently, the grounding of vision and language foundation models with geometric and object-centric representations of the world has enabled tremendous progress in context understanding, which is a vital requirement for robots to interact with the real world. We will thoroughly examine the application of VFMs and VLMs in robotic perception from various perspectives.

\paragraph{VFMs, VLMs for Object and Scene Representations}
The most straightforward application of VLMs in robotics is to leverage their ability to perform open-set object recognition and scene understanding in robotics-oriented downstream tasks, including semantic mapping and navigation \cite{Chen_2022_nlmapsaycan, huang23vlmaps, Shafiullah_2022_CLIPFields, Jatavallabhula_2023_ConceptFusion, Shah_2022_LMNav},  manipulation \cite{Shridhar_2021_CLIPORT, Shen2023f3rm, Ze2023GNFactor, homerobotovmmchallenge2023}, etc. The methods proposed by these works share a common attribute: they attempt to extract \textit{semantic} information (from the VLM) and \textit{spatial} information (from other modules or sensor modalities) from objects and scenes that the robots interact with. This information is then used as representations in semantic maps of scenes or representations of objects. 

% The semantic and spatial perception part, as we discover, is rather loosely coupled with other robotic modules. 

%We here list some examples of recent work in this category. 
For semantic mapping and/or navigation, \texttt{NLMap} \cite{Chen_2022_nlmapsaycan} is an open-set and queryable scene representation to ground task plans from LLMs in surrounding scenes. 
The robot first explores the environment using frontier-based exploration to simultaneously build a map and extract class-agnostic regions of interest, which are then encoded by VLMs and aggregated to the map. 
Natural language instructions are then parsed by an LLM to search for the availability and locations of these objects in the scene representation map.
\texttt{ConceptFusion} \cite{Jatavallabhula_2023_ConceptFusion} builds open-set multimodal 3D maps from RGB-D inputs and features from foundation models, allowing queries from different modalities such as image, audio, text, and clicking interactions. 
It is shown that \texttt{ConceptFusion} can be applied for real-world robotics tasks, such as tabletop manipulation of novel objects and semantic navigation of autonomous vehicles. 
Similarly, \texttt{CLIP-Fields} \cite{Shafiullah_2022_CLIPFields} encodes RGB-D images of the scene to language-queryable latent representations as elements in a memory structure, that the robot policy can flexibly retrieve. 
\texttt{VLMap} \cite{huang23vlmaps} uses \texttt{LSeg} \cite{Li_2022_LSeg} to extract per-pixel representations to then fuse with depth information, in order to create a 3D map. This semantic 3D map is then down-projected to get a 2D map with the per-pixel embedding; these embeddings can then be matched with the language embedding from \texttt{LSeg} to obtain the per-pixel semantic mask for the 2D map. As for applying VLMs in topological graphs for visual navigation, \texttt{LM-Nav} \cite{Shah_2022_LMNav} is a good example: 
it uses an LLM to extract landmarks used in navigation from natural language instructions. 
These landmark descriptions, as well as image observations, are then grounded in a pre-built graph via a VLM. Then, a planning module is used to navigate the robot to the specified landmarks.

Most of the previous works discussed above utilize only 2D representation of the objects and environment. To enrich the representation of foundation models in 3D space, \texttt{F3RM}~\cite{Shen2023f3rm}, \texttt{GNFactor}~\cite{Ze2023GNFactor} and \texttt{GeFF}~\cite{qiuhu2024learning} distill 2D foundation model features into 3D space, by combining with (generalizable) NeRF. In addition, \texttt{GNFactor}~\cite{Ze2023GNFactor} also apply these distilled features in policy learning. \texttt{Act3D}~\cite{gervet2023act3d} takes a similar approach but build 3D feature field via sensed depth.
%learn about non-visual object properties, e.g., via a process called interactive perception, in order to improve agents' performance in downstream interactive robot learning tasks \cite{sinapov_grounding_2014,sinapov_learning_2014,gemici_learning_2014,tatiya2019sensorimotor,tatiya2020framework,tatiya2020haptic,tatiya2023crosstool,tatiya2023mosaic}, in a similar fashion to how human infants first learn about the physical world \cite{thesen_Neuroimaging_2004,wilcox2007multisensory}. 

% Beyond scene understanding, visual foundation models have also been used to learn affordance functions for policy learning.
% One such example is CLIPort \cite{Shridhar_2021_CLIPORT}.
% It leverages a semantic stream (based on RGB images and language instructions) and a spatial stream (based on a RGB-D transporter network~\cite{zeng2020transporter}) to infer the affordance function for a manipulation policy.

% It takes RGB images and language instructions as the semantic stream, along with a spatial stream that uses a transporter network \cite{zeng2020transporter} to infer spatial displacements from RGBD inputs, to estimate the affordance function for manipulation policy.

\paragraph{VLMs for State Estimation and Localization} Beyond context understanding, a few approaches explore the use of open-vocabulary properties of VLMs for state estimation~\cite{kassab2023language,mirjalili2023fm,mirjalili2023fm,keetha2023anyloc,he2023foundloc}.
Two such approaches, \texttt{LEXIS}~\cite{kassab2023language} and \texttt{FM-Loc}~\cite{mirjalili2023fm}, explore the use of \texttt{CLIP}~\cite{Radford_2021_CLIP} features to perform indoor localization and mapping. 
In particular, \texttt{FM-Loc}~\cite{mirjalili2023fm} leverages the vision-language grounding offered by \texttt{CLIP} and \texttt{GPT-3} to detect objects and room labels of a query image, then uses that semantic information to match it with reference images. 
Similarly, \texttt{LEXIS}~\cite{kassab2023language} builds a real-time topological SLAM graph where \texttt{CLIP} features are associated with graph nodes, enabling room-level scene recognition.
Although these approaches display the potential of vision-language features for indoor place recognition, they do not explore the broad applicability of foundation model features.
In this context, \texttt{AnyLoc}~\cite{keetha2023anyloc} explored the properties of dense foundation model features and combined them with unsupervised feature-aggregation techniques to achieve state-of-the-art place recognition, by large margin---anywhere, anytime, and under any view---showcasing broad applicability of self-supervised foundation model features for SLAM.
Extending this further, \texttt{FoundLoc}~\cite{he2023foundloc} coupled \texttt{AnyLoc} with a VIO pipeline to perform visual GNSS-denied localization.
For the first time, this work showcased that VLMs can be deployed on resource-constrained unmanned aerial vehicles (UAVs) \& embedded Jetson hardware for state estimation in the wild.

\paragraph{VLMs for Interactive Perception}
% More recently, work such as ConceptGraphs~\cite{gu2023conceptgraphs} and OVSG~\cite{chang2023contextaware} has explored ways to build open-vocabulary scene graphs to enable multi-modal queries for localization, object goal navigation, manipulation and traversability affordance. 
\looseness = -1
Several works consider the notion of enabling robots to leverage the process of interactive perception~\cite{bohg2017interactive}, for extrapolating implicit knowledge about object properties in order to obtain performance improvements on downstream interactive robot learning tasks \cite{sinapov_grounding_2014,sinapov_learning_2014,gemici_learning_2014,tatiya2019sensorimotor,tatiya2020framework,tatiya2020haptic,tatiya2023crosstool,tatiya2023transferring,tatiya2024mosaic,Fang2024Uncos}. This process of interactive perception is often modeled after the way in which human infants first learn about the physical world---i.e., through interaction, and by learning representations of object concepts, such as weight and hardness, from the sensory information (haptic, auditory, visual) that is generated from those physical exploratory actions (e.g., grasping, lifting, dropping, pushing) on objects with diverse properties. In particular, \texttt{MOSAIC} \cite{tatiya2024mosaic} leverages LMMs to expedite the acquisition of unified multi-sensory object property representations; the authors show competitive performances of their framework in category recognition and ambiguous target-object fetch tasks, despite the presence of distractor objects, under zero-shot transfer conditions. \texttt{UncOS}~\cite{Fang2024Uncos} queries LVMs to generate a distribution of object segmentation hypotheses. They convert the hypotheses into a distribution over world models and use that to select robot perturbation actions for embodied disambiguation of object segmentation.

% Additionally, to account for physical factors such as temperature or the movability of objects, the approach mentioned in \cite{piglet} incorporates physical commonsense knowledge through interactions within the map. Symbolic representations are used to describe this physical knowledge, allowing for a more comprehensive understanding of the environment.

\subsubsection{LLMs and VLMs in Task Planning} \label{subsec:llm_vlm_planning}

The planning community in Robotics has always harbored aspirations of a model capable of generalizing across diverse tasks and environments, with minimal demonstrations for robotic tasks. Given the demonstrated prowess of vision and language foundation models in intricate reasoning and contextual generalization, it is a natural progression for the robotics community to consider the application of foundation models to robotic planning problems. This section organizes works based on the granularity of planning, delineating between task-level and motion-level planning. We will mainly introduce task-level planning in this part and leave motion-planning to the next part, together with action generation (Section \ref{subsec:llm_action}). 

% \paragraph{Task-level Planning} 
Task-level planning is to divide a complicated task into small actionable steps. In this case, we mainly talk about the agent planning on its own, in contrast to, e.g., using LLMs as a parser like Vision Language Navigation~\cite{anderson2018vision}. The agent needs to take intelligent sub-steps to reach the goal by interacting with the environment. \texttt{SayCan}~\cite{Ahn_2022_saycan} is a representative example of task-level planning: it uses LLMs to plan for a high-level task, e.g., ``I spilled my drink, can you help?''. Then it gives concrete task plans like going to the counter, finding a sponge, and so on. Similarly, \texttt{VLP}~\cite{du2023video} aims to improve the long-horizon planning approach with an additional text-to-video dynamics model. These task-level planning methods do not have to worry about the precise execution of those sub-tasks in the environment, since they have the luxury of utilizing a set of pre-defined / pre-trained skills, then using LLMs to simply find proper ways to compose skills to achieve desired goals. There are more papers in this category, for example: \texttt{LM-ZSP}~\cite{ZSP} introduce this task-level granularity as actionable steps; \texttt{Text2Motion}~\cite{lin2023text2motion} uses similar ideas and augments the success rate of language based manipulation task. Previous methods typically generate task plans in the form of text, Some works like \texttt{ProgPrompt} \cite{singh2022progprompt}, \texttt{Code as Policy}~\cite{codeaspolicy}, \texttt{GenSim}~\cite{wang2023gensim}, etc. obtain task plans in the form of code generation using LLMs. \cite{huang2024rekep} take a different approach by obtaining relational constraints from code writing LLMs. Using code as a high-level plan has the benefit of expressing functions or feedback loops that process perception outputs and parameterize control primitive APIs. In addition, it can describe the spatial position of an object accurately. This improved compositionality saves time in collecting more primitive skills. It also prescribes precise values (e.g., velocities) to ambiguous descriptions like 'faster' and 'to the left', depending on the context. Therefore, due to these benefits, code appears to be a more efficient and effective task-level planning language than natural language. 
Other forms of planning techniques such as expressing the high-level plans in Planning Domain Definition Language (PDDL)\cite{seipp2022pddl} also showed significant improvement in LLMs planning power over long horizon tasks, more on this will be discussed in Section \ref{subsec:special_prompting}.

\looseness = -1
In addition to using LLMs to directly generate plans, they are also used in searching and evaluating with external memory structures such as scene graph. \texttt{SayPlan}~\cite{rana2023sayplan} employs 3D scene graph (3DSG) representations to manage the complexity of expansive environments. By exploiting hierarchical 3DSGs, LLMs can semantically search for relevant sub-graphs in multi-floor household environments, reducing the planning horizon and integrating classical path-planning to refine initial plans, iteratively. \texttt{Reasoned Explorer}~\cite{xie2023reasoning} employs LLMs as evaluators to score each node in a 2D undirected graph. It uses this graph as a map to store both visited points and the frontiers' LLM evaluations. These external memories and incremental map-building approaches break the context length limit of using LLMs to generate long plans, which scales LLM-based navigation agents to large environments. One thing to note is that, although task-level planning is agnostic to physical embodiment, it does require grounding to a specific robot's physical form (or ``morphology") and environment when deployed; grounding techniques will be covered in Section \ref{subsec:grounding}.

\label{action}
\subsubsection{LLMs and VLMs in Action Generation} \label{subsec:llm_action}
Directly controlling a robot just by prompting off-the-shelf LLMs/VLMs can be challenging, perhaps even unachievable, without first fine-tuning these models with action data. Unlike high-level robot task planning, where LLMs are used for their ability to compose and combine different skills for task completion, individual actions, both high-level actions like waypoints, and low-level actions like joint angles are usually not semantically meaningful or compositional. The community is attempting to find a suitable interface to circumvent this problem. For motion planning in navigation tasks, \texttt{ReasonedExplorer} \cite{xie2023reasoning} and \texttt{Not\_Train\_Dragon} \cite{chen2023not} propose such an interface: using LLMs as evaluators for the expanded frontiers, which are defined as potential waypoints for exploration (typically in a two-dimensional space); here, LLMs are tasked with scoring frontiers based on the similarity between the given observations and the intended goal. Similarly, VoxPoser~\cite{huang2023voxposer} apply VLMs to obtain affordance function (called 3D value map in the original paper) used in motion planning. 
% \zkn{Incomplete/unfinished sentence! Yafei: Done}

Some papers investigate the use of LLMs to directly output lower-level actions. \texttt{Prompt2Walk}~\cite{wang2023prompt} uses LLMs to directly output joint angles through few-shot prompts, collected from the physical environment. It investigates whether LLMs can function as low-level controllers by learning in-context with environment feedback data (observation-action pairs). \texttt{Saytap}~\cite{tang2023saytap}, introduces a novel concept of utilizing foot contact patterns as an action representation. In this model, the language model outputs a `0' for no contact and a `1' for contact with the floor, thereby enabling Large Language Models (LLMs) to generate zero-shot actionable commands for quadrupedal locomotion tasks such as jumping and jogging. However, the generalizability of these approaches to different robot morphologies remains in question, since they have only been tested on the quadruped platform. Instead, language to reward~\cite{du2023guiding, kwon2023reward, xie2023text2reward} in robotics~\cite{yu2023language, ma2023eureka} is a more general approach than direct action generation through LLMs; these approaches involve using LLMs as generators to synthesize reward functions for reinforcement learning-based policies and thus are usually not confined by robotic embodiments~\cite{ma2023eureka}. The reward synthesizing approach with LLM can generate rewards which are hard for human to design, e.g., \texttt{Eureka}~\cite{ma2023eureka} demonstrates that it enables robots to learn dexterous pen-spinning task that were considered very hard using human reward design.

\subsubsection{Action Grounding}
\label{subsec:grounding}
\looseness = -1
Traditionally, grounding refers to associating abstract and contextual meaning with sensory signals, such as identifying objects in images \cite{Cangelosi2010IntegrationOA} or recognizing sounds in audio \cite{deshmukh2023pengi}. This form of "direct grounding" in perceptual modalities has been extensively studied. However, in robotics, grounding extends to a crucial additional dimension: associating abstract concepts with executable actions. In other words, it involves translating the output of foundational models into meaningful changes in the environment. Thus, in this survey, we focus on action grounding.

Action grounding is just as vital as sensory grounding for robotics. For robots to interact with the physical world, foundational models require a bridge that maps their outputs to actionable behaviors, either explicitly or implicitly.

The initial approaches~\citep{Ahn_2022_saycan, arenas2023how, wang2023prompt,liu2024okrobot} that tried to explicitly map foundation models' outputs or representations to robot actions are fairly straightforward. The foundational models used are mainly large language models, and, like traditional robotics approaches, they adopt a modular concept that breaks down the robot's preliminary skills. These skills are obtained either through pre-trained policies or pre-programmed scripts, and the entire skill library is described to the foundation models in natural language, allowing LLMs to output action plans using these skills. Recent progress in robotic grasping foundation models like \texttt{AnyGrasp}~\cite{fang2023anygrasp} and other open knowledge models like \texttt{CLIP}~\cite{Radford_2021_CLIP} and \texttt{SAM}~\cite{kirillov2023segany} makes this approach more attractable. However, there are still two bottlenecks that it needs to solve: first, the limitations of natural language as a general interface, since the granularity that natural language can describe is limited. For example, if you were to use language to describe how to play Cat's Cradle with a dexterous robot hand, language would likely struggle to accurately describe it. The second limitation is that, if we want the model to generalize to different tasks, more pre-trained or pre-programmed skills are needed, making it difficult to scale.

Therefore, researchers are exploring different ways to address these issues. Code as an interface alleviates some limitations of natural language granularity \cite{codeaspolicy,singh2022progprompt}, as it can directly describe an object's x, y, z position for grasping. Other novel interfaces try to involve directly anchoring foundational models to output joint torques, circumventing intermediary interfaces such as text tokens. A notable example is \texttt{Gato} \cite{Reed_2022_Gato}, which directly maps the model's output to the Atari game action space and robotic arm joint space. \texttt{Gato} dynamically decides its output format—be it text, joint torques, button presses, or other tokens—based on the context of the task at hand. Another related development is \texttt{RT-2} \cite{2023rt2}, which, despite specifying the end-effector space in textual form, is capable of directly generating executable commands for robotic manipulator operation. \texttt{SayTap} \cite{tang2023saytap} has experimented with directly generating binary foot patterns to control quadruped locomotion tasks.

Building on this idea of grounding models to real-world action spaces, approaches such as \texttt{CLIP-Fields} \cite{Shafiullah_2022_CLIPFields}, \texttt{VLMap} \cite{huang23vlmaps}, and \texttt{NLMaps} \cite{Chen_2022_nlmapsaycan} project \texttt{CLIP} visual and semantic label representations directly onto 3D point clouds. By aligning semantic understanding with spatial information, these methods create more interpretable 3D maps for robotic applications. Going beyond explicit 3D mapping, \texttt{GLAM} \cite{carta2023grounding} uses reinforcement learning to ground language models through interactions with the environment, demonstrating that LLMs can function effectively as RL agents in textual environments. Moreover, \texttt{AutoRT} \cite{gdm2024autort} acts as an orchestrator for task performance by a fleet of robots, leveraging vision-language models \cite{chen2023palix,dwibedi2024flexcap} for scene description to prompt LLMs for task generation.

\looseness = -1
Additionally, researchers are trying different interfaces to map the output of the foundation model to actions. \texttt{Voxposer} \cite{huang2023voxposer} maps the output to a 3D value map and then uses classical optimization techniques to calculate robot actions. This approach adds more robustness to disturbances in the pipeline and allows the model to manipulate trajectories to add safety constraints and avoid obstacles in real time. In addition to value maps, using constraints is also a popular approach to achieve grounding in more diverse tasks. Geometric constraints are used to compute feasible trajectories that avoid obstacles and achieve goals, while contact constraints are used to plan forceful or contact-rich behaviors. \texttt{ReKep} uses keypoint relationship constraints to represent both geometric and contact constraints as Python functions that map a set of keypoints to a numerical cost. Each keypoint is task-specific and represents a semantically meaningful 3D point in the scene. 

While several grounding-to-action techniques are discussed above, there is no definitive answer as to which approach offers the most optimal solution. Pretrained skill libraries provide high levels of dexterity and precision in task execution, but at the expense of task diversity. Conversely, map- or constraint-based grounding techniques offer greater task flexibility but have primarily been demonstrated on simpler 2D gripper pick-and-place tasks, with no tests on more dexterous, complex actions. If we view grounding to action as a spectrum, the ideal interface should balance both task diversity and task complexity, excelling not only in the breadth of tasks it can perform but also in the intricacy of those tasks.

\subsubsection{Data Generation with LLMs and VGMs}
\label{subsec:special_data_generation}

Recently, we have witnessed the power of content generation ability of LLMs and VGMs. Utilizing this ability, researchers have begun attempts to address the data scarcity problem by using foundation models to generate data. Ha \textit{et al.}~\cite{ha2023scalingup} propose a framework, `\textit{scaling up and distilling down}', which, given natural language instructions, can automatically generate diverse robot trajectories labeled with success conditions and language. \texttt{RoboGen} by Wang \textit{et al.}~\cite{wang2023robogen} further enhances this approach by incorporating automatic action trajectory proposals within a physics-realistic simulation environment, Genesis, enabling the generation of potentially-infinite data. Nevertheless, these approaches still face limitations: the data generated suffers from low diversity in assets and robot morphologies, issues that could be ameliorated with advanced simulations or hardware. \texttt{GenSim}~\cite{wang2023gensim} by Wang \textit{et al.} proposes to generate novel long-horizon tasks with LLMs given language instructions, leading to over 100 simulation tasks for training language-conditioned multitask robotic policy. This framework demonstrates task-level generalization in both simulation and the real world, and sheds some light on how to distill foundation models to robotic policies through simulation programs. \texttt{ROSIE} by Yu \textit{et al.} \cite{Yu_2023_ROSIE} uses a text-guided image generator to modify the robot's visual observation to perform data augmentation when training the control policy. The modification commands are from the user's language instruction, then the augmentation regions are localized by the open vocabulary segmentation model. \texttt{RT-Trajectory} \cite{gu2023rttrajectory} generates trajectories for the policy network to condition on. The trajectory generation also helps the task specification in the robot learning tasks. Black \textit{et al.}~\cite{black2023zeroshot} use a diffusion-based model to generate subgoals for a goal-conditioned RL policy for manipulation \cite{du2023learning}.

\subsubsection{Enhancing Planning and Control Power through Prompting} 
\label{subsec:special_prompting}
% \todo{Quanting; text needs to be polished. \textcolor{blue}{Done}} 
The technique of Chain-of-Thought, as introduced by Wei \textit{et al.} \cite{wei2023chainofthought}, compels the LLM to produce intermediate steps alongside the final output. This approach leverages a broader context window to list the planning steps explicitly, which enhances the LLM's ability to plan. The underlying reason for its effectiveness is the GPT series' nature as an autoregressive decoder. Semantic similarities are more pronounced between instructions to steps and steps to goal, than between instructions to the direct output. Nonetheless, the sequential nature of the Chain-of-Thought implies that a single incorrect step can lead to exponential divergence from the correct final answer~\cite{dziri2023faith}.

Alternative methodologies attempt to remedy this by explicitly listing steps within graph~\cite{besta2023graph} or tree structures~\cite{yao2023tree}, which have demonstrated improved performance. Additionally, search-based methods such as Monte Carlo Tree Search (MCTS)~\cite{zhang2023planning} and Rapidly-exploring Random Tree (RRT)~\cite{xie2023reasoning} have been explored to augment planning capabilities.

Furthermore, translating goal specifications from natural language into external planning languages, such as the Planning Domain Definition Language (PDDL), has also been shown to increase planning accuracy~\cite{liu2023llmp}. Finally, as opposed to an open-loop prompting style, iterative prompting approaches that incorporate feedback from the environment provide a more grounded and precise enhancement for long-term planning capability~\cite{innermonologe, driess2023PaLMe}.

\subsection{Robotics Foundation Models (RFMs)} \label{subsec:rfm}

With the increasing amount of robotics datasets, containing state-action pairs from real robots, the class of Robotics Foundation Models (RFMs) have likewise become increasingly viable \citep{2023rt2, embodimentcollaboration2023open, Shah_2023_ViNT, shafiullah2023bringing, shafiullah2022behaviorTransformers, jain2024vid2robot}. These models are characterized by the use of robotics data to train them, to solve robotics tasks. In this subsection, we summarize and discuss different types of RFMs. We will first introduce RFMs that can perform tasks according to a single robotic capability (e.g., perception, planning, and control), which is defined as \textit{single-purpose} Robotic Foundation Models. For example, an RFM that can generate low-level actions to control the robot, or a model that can generate higher-level motion plans. We later introduce RFMs that can carry out tasks in multiple robotic modules, hence \textit{general-purpose} models that can perform perception, control, and even non-robotic tasks \citep{Reed_2022_Gato, driess2023PaLMe}. 

\subsubsection{Robotics Action Generation Foundation Models}
\label{rfm_action}
% While large foundational models have been effective in performing high-level task planning, recent works have also demonstrated their use in generating direct control outputs for robotic applications. Often referred to as ``end-to-end control'', these methods directly learn to predict robot actions like accelerations, torques, joint velocities, etc. from high-dimensional inputs like language or vision.
% \subsubsection{Special Topic: Pre-training in Robotics} \label{subsec:pretraining}
% \cite{radosavovic2023real} \cite{Majumdar2023VC1}.
Robotic action foundation models could take raw sensory observations, e.g., images or videos, and learn control output that is directly applied to robotic end-effectors. Models in this category include \texttt{RT} series~\cite{brohan2022rt, 2023rt2, embodimentcollaboration2023open}, \texttt{RoboCat}~\cite{bousmalis2023robocat}, \texttt{MOO}~\cite{stone2023open}, etc. According to the papers' results, these models show generalization in robot control tasks such as manipulation. 

\paragraph{Imitation Learning}
Imitation learning has been applied to robot control for a considerable period. Initially, methods primarily aimed at imitating a single skill. The concept of utilizing imitation learning to master multiple tasks, which is the focus of this review, emerged with the works of~\cite{duan2017one, finn2017one}. This approach is known as one-shot imitation learning. Various conditions are employed to define the task, including robot goal images~\cite{duan2017one, finn2017one, dasari2021transformers, mandi2022towards, lynch2020learning, james2018task, zhou2019watch, wang2023mimicplay}, human goal images~\cite{jang2021bc, yu2018one, bonardi2020learning}, language prompts~\cite{lynch2020language, stepputtis2020language, jang2021bc}, and task vectors~\cite{rahmatizadeh2018vision}, among others. A recent study also investigates the use of multi-modal input as a task descriptor~\cite{Jiang2022vima}. These work represent early exploration in this direction. Lately, efforts have been made to scale this approach, emulating the success of large language models. Lynch \textit{et al.}\cite{lynch2022interactive} employed behavioral cloning on a dataset of hundreds of thousands of language-annotated trajectories to learn tabletop block rearrangement in the real world. Brohan \textit{et al.}\cite{brohan2022rt} trained a large transformer model on over 130K tabletop rearrangement episodes with a more realistic setup and diverse objects, demonstrating the robustness of imitation learning algorithms on some unseen tasks. \texttt{RoboCat}\cite{bousmalis2023robocat} and \texttt{RT-X}\cite{embodimentcollaboration2023open} trained a single model with data from multiple embodiments, showing some generalization capability to unseen embodiments during testing. Overall, these efforts have primarily shown success in pick-and-place tasks or variants thereof, indicating significant potential for further exploration.

\paragraph{Reinforcement Learning at Scale}
% As discussed in Section~\ref{subsec:control} (Preliminaries), offline RL has the potential to learn good policies without even interacting with the environment. 
With the availability of large-scale robotic datasets, offline RL starts to play an important role in developing effective RFMs. Early large-scale offline RL models such as \texttt{QT-OPT}~\cite{2018_Kalashnikov_QT-Opt} use a Q-learning-based approach in an offline manner to learn policy from robotics data which are collected by a robot farm. The successors of \texttt{QT-OPT} extend it to multitask learning by incorporating multi-task curriculum or predictive information~\cite{Kalashnkov2021MTOPT, lee2022pi, herzog2023deep}. Recently, with the success of Transformer models, Q-learning based on transformer (\texttt{Q-Transformer}) also shows its potential \cite{chebotar2023qtransformer}. \texttt{PTR}~\cite{kumar2023pretraining} is another promising work that adopts Conservative Q-Learning (\texttt{CQL})~\cite{kumar2020conservative} in a multi-task learning setting. When it comes to online RL, due to the large exploration space, most methods trained a visual policy in the simulation~\cite{huang2021generalization, chen2022system} and then transferred the model to real world~\cite{chen2023visual}. Still, this field is challenging, and we look forward to seeing more RL-based robotic foundation models.

\paragraph{Vision and Language Pre-training}
Another direction of action foundation models involves vision or language pre-training ~\cite{parisi2022unsurprising, li2022pre, nair2022r3m, radosavovic2023real, radosavovic2023robot, hansen2023pretraining, Majumdar2023VC1, 2023rt2}. For example, inspired by the great generalization ability of self-supervised vision-based pre-training, \texttt{MVP} by Radosavovic \textit{et al.}~\cite{radosavovic2023real} trains visual representations on real-world images and videos from the internet and egocentric video datasets, via a masked autoencoder,
and demonstrated the effectiveness of scaling visual pre-training for robot learning. Following this work, \texttt{RPT}~\cite{radosavovic2023robot} proposes mask-pretraining with real robot trajectory data. \texttt{VC-1}~\cite{Majumdar2023VC1} did a comprehensive study on the effectiveness of vision-based pre-training on policy learning. 
% We also recommend readers to learn more information about this problem from that paper. 
Despite the effectiveness of these visual pretraining methods, \cite{hansen2023pretraining} reexamined some of these methods and discovered significant domain gaps, thus proposed learning-from-scratch approach. This provides us new perspective to think about visual pretraining in robotics.

Beyond just using visual information, \texttt{RT-2}~\cite{2023rt2} and \texttt{Moo}~\cite{stone2023open} use vision and language pre-trained model as a control policy backbone. \texttt{PaLM-E}~\cite{driess2023PaLMe} and \texttt{PALI-X}~\cite{chen2023palix} were used to transfer knowledge from the web into robot actions.
% Combined vision and language input have also been explored to improve end-to-end robot control. In \cite{stone2023open}, the authors leverage a pre-trained vision-language model to
% extract information from vision and language, and use it to condition the control policy on the current image, the instruction, and the extracted information to achieve robot control. 
Slightly different from previous methods, \texttt{VRB}~\cite{bahl2023affordances} learns affordance functions (instead of the policy itself) from large-scale video pertaining, providing another thought process for us to study how RFMs may generalize in real-world tasks.

Using a similar approach as in pretraining with vision or language modality, we also see self-supervised pretraining with less-explored modalities such as audio ~\cite{thankaraj2022sounds} and tactile sensing~\cite{guzey2023dexterity}.

% \subsubsection{Single-purpose Robotic Foundation Models}
% \label{subsec:single_purpose}
\paragraph{Robotics Motion Planning Foundation Models}
\label{rfm_motionplanning}
Recently we have seen the rise of RFMs especially used for motion-planning purposes in visual navigation tasks \cite{Shah_2023_GNM,Shah_2023_ViNT,Truong_2023_IndoorSimtoOutdoorReal}. These foundation models take advantage of the large-scale heterogeneous data and show generalization capability in predicting high-level motion-planning actions. These methods rely on rough topological maps \cite{Shah_2023_GNM,Shah_2023_ViNT} consisting of only image observations 
% or just rough top-down 2D maps \cite{Truong_2023_IndoorSimtoOutdoorReal}, 
instead of accurate metric maps and accurate localization as in conventional motion-planning methods (as described in Section \ref{challenge_model}). Unlike vision and language foundation models applied to motion planning, the robotic motion planning foundation model is still quite in its early stages.%, so we have not yet seen many publications in this field.

%% (Yafei)
\subsubsection{General-purpose Robotics Foundation Models}
\label{subsec:general_purpose}
% One Model to Rule them all! :-)
Developing general-purpose robotic systems is always a holy grail in robotics and artificial intelligence. Some existing works \cite{Reed_2022_Gato, driess2023PaLMe} take one step towards this goal. \texttt{Gato} \cite{Reed_2022_Gato} proposes a multimodal, multi-task, and multi-embodiment generalist foundation model that can play Atari games, caption images, chat, stack blocks with a real robot arm, and more---all with the same model weights. Similar to \texttt{Gato}, \texttt{PaLM-E} \cite{driess2023PaLMe} is also a general-purpose multimodal foundation model for robotic reasoning and planning, vision-language tasks, and language-only tasks. 
Although not proven to solve all the robotics tasks that we introduced in Section \ref{sec:preliminary}, \texttt{Gato} and \texttt{PaLM-E} show a possibility of merging perception and planning into one single model. Moreover, \texttt{Gato} and \texttt{PaLM-E} show promising results of using the same model to solve various seemingly-unrelated tasks, highlighting the viability of general-purpose AI systems. Designed especially for robotic tasks, \texttt{PACT} \cite{Bonatti_2022_PACT} proposes one transformer-based foundation model with common pre-trained representations that can be used in various downstream robotic tasks, such as localization, mapping, and navigation. Although we have not seen many unified foundation models for robotics, we would expect more endeavors in this particular problem.

% (Yafei)
\begin{table*}[]
\color{black}
\centering
\resizebox{\linewidth}{!}{ 
{\renewcommand{\arraystretch}{1.5}%
\begin{tabular}{c|c|c|c|c|c|c}
\toprule
% Modules & Foundation Models  & Generalization \ref{challenge_generalization} & Data \ref{challenge_data} & Model  \ref{challenge_model} & Task Specification \ref{challenge_task_specification} & Uncertainty \ref{challenge_safety} \\ \hline
Modules & Foundation Models  & \multicolumn{5}{c}{Challenges} \\ \cline{3-7}
       &                    & Generalization \ref{challenge_generalization} & Data \ref{challenge_data} & Model  \ref{challenge_model} & Task Specification \ref{challenge_task_specification} & Uncertainty \ref{challenge_safety} \\ \hline

\multirow{3}{*}{Perception} &VFM &\texttt{Conceptgraphs} \cite{gu2023conceptgraphs}  &\texttt{Conceptgraphs} \cite{gu2023conceptgraphs}  & -   & -   & -  \\ \cline{2-7}

&VGM & -  &\begin{tabular}[c]{@{}l@{}}\texttt{ROSIE}~\cite{Yu_2023_ROSIE} \\ \texttt{RoboGen}~\cite{wang2023robogen} \end{tabular} & -  & - \\ \cline{2-7}

&VLM & \texttt{NLMap}~\cite{Chen_2022_nlmapsaycan} &\begin{tabular}[c]{@{}l@{}}\texttt{NLMap}~\cite{Chen_2022_nlmapsaycan} \\ \texttt{RT-Traj.}~\cite{gu2023rttrajectory} \end{tabular}& - & \texttt{RT-Traj.}~\cite{gu2023rttrajectory} & -  \\ \hline

\multirow{2}{*}{\shortstack{Task Planning and \\ Action Generation}}
&LLM  &\texttt{SayCan}~\cite{Ahn_2022_saycan} &\begin{tabular}[c]{@{}l@{}} \texttt{SayCan}~\cite{wang2023robogen} \\ \texttt{RT-Traj.}~\cite{gu2023rttrajectory} \end{tabular} &\texttt{RAP}~\cite{hao2023reasoning} \texttt{UniSIM}~\cite{yang2023learning} &\begin{tabular}[c]{@{}l@{}}\texttt{LILA}~\cite{karamcheti2022lila} \\ \texttt{L2R} ~\cite{yu2023language} \end{tabular} & \texttt{KNOWNO} \cite{ren2023robots}\\ \cline{2-7}

 &RFM \ref{subsec:rfm} &\begin{tabular}[c]{@{}l@{}}\texttt{RT-1}~\cite{brohan2022rt} \\ \texttt{RT-2}~\cite{2023rt2} \\ \texttt{RoboCat}~\cite{bousmalis2023robocat} \\ \texttt{VINT}~\cite{Shah_2023_ViNT} \\ \texttt{OpenVLA}~\cite{kim2024openVLA} \\ \texttt{CrossFormer}~\cite{Doshi24-crossformer} \end{tabular}  & \begin{tabular}[c]{@{}l@{}}\texttt{RT-X}~\cite{embodimentcollaboration2023open} \\ \texttt{RH20T}~\cite{fang2024rh20t} \end{tabular} &\begin{tabular}[c]{@{}l@{}}\texttt{RT-1}~\cite{brohan2022rt} \\ \texttt{RT-2}~\cite{2023rt2} \\ \texttt{RoboCat}~\cite{bousmalis2023robocat} \end{tabular}  &\begin{tabular}[c]{@{}l@{}} \texttt{Zest}~\cite{cui2022zest} \end{tabular} & - \\
\bottomrule
\end{tabular} 
}
}
\vspace{1mm}
\caption{\justifying A quick summary of foundation models solving robotic challenges. Here we only list part of the works due to space limit. We find that uncertainly and safety are still largely unexplored.}
\label{Tab:summary_challenges}
% \vspace{-1.5em}
\end{table*}

\subsection{How do Foundation Models Help Solve Robotics Challenges}
% As we discussed in section \ref{sec:challenges_robotics}, vision and language foundation models have helped with the perception module of robotics, task planning, and even control in many scales. However, the inherent inexplicability of these foundation models poses security issues, which has already been reflected in LLM \cite{}. In order to apply these vision and language models in robotics, we may also have to address this issue. 

In Section~\ref{sec:challenges_robotics}, we listed five major challenges in Robotics. In this section, we summarize how foundation models---either vision and language models or robotic foundation models---could help resolve these challenges, in a more organized manner. 

All the foundation models related to visual information, such as VFMs, VLMs, and VGMs, are used in the perception modules in Robotics. LLMs, on the other hand, are more versatile and can be applied in both planning and control. We also list RFMs here, and these robotic foundation models are typically used in planning and action generation modules.  We summarize how foundation models solve the aforementioned robotic challenges in Table \ref{Tab:summary_challenges}. We notice from this table that all foundation models are good at generalization in tasks of various robotic modules. Also, LLMs are especially good at task-specification. RFMs, on the other hand, are good at dealing with the challenge of dynamics model
% \zkn{I would remind reachers what ``challenge in model'' means as it was early in the document and readers will not remember. Note that there is an overleading of the term model here that makes it confusing, so should be paired with something (dynamics model, world model, etc.) Yafei: Done} 
since most RFMs are model-free approaches. For robot perception, the challenges in generalization ability and model are coupled, since, if the perception model already has very good generalization ability, there's no need to get more data for domain adaptation or additional fine-tuning. In addition, the call for solving the safety challenge is largely missing, and we will discuss the particular problem in Section \ref{sec:discussion}.

% Consider a table showing how many months of data collection, and how many trajectories
% compare the data size with other foundation models such as LLM and VLM, shown in the bar plot. 

% With the emergence of robotic foundation models \cite{Shah_2023_ViNT} \cite{2023rt2}, we see a viable approach towards more general robots. However, existing works are largely limited to manipulation and navigation tasks. In addition, the robustness of these methods still requires further investigation. We will discuss the remaining challenges of robotics which haven't yet been solved by foundation models.
\paragraph{Foundation Models for Generalization} 
Zero-shot generalization is one of the most significant characteristics of current foundation models. Robotics benefits from the generalization ability of foundation models in nearly all aspects and modules. For the first one, generalization in perception, VLM and VFM are great choices as the default robotics perception models. The second aspect is the generalization ability in task-level planning, with details of task plans generated by LLMs~\cite{Ahn_2022_saycan}. The third one is in generalization in motion-planning and control, by utilizing the power of RFMs.

% \vspace{-1em}

%% moved from sec 3
\paragraph{Foundation Models for Data Scarcity} 
Foundation models are crucial in tackling data scarcity in robotics. They offer a robust basis for learning and adapting to new tasks with minimal specific data. For example, recent methods utilize foundation models to generate data to help with training robots, such as robot trajectories~\cite{ha2023scalingup} and simulation~\cite{wang2023robogen}. These models excel in learning from a small set of examples, allowing robots to quickly adapt to new tasks using limited data. From this perspective, solving data scarcity is equivalent to solving the generalization ability problem in robotics. Apart from this aspect, foundation models---especially LLMs and VGMs---could generate datasets for robotics used in training perception modules~\cite{Yu_2023_ROSIE} (see Section \ref{subsec:special_data_generation}, above), and for task-specification~\cite{gu2023rttrajectory}. %We also summarize data generation in the previous part .

\paragraph{Foundation Models to Relieve the Requirement of Models} As discussed in Section \ref{challenge_model}, building or learning a model---either a map of the environment, a world model, or an environmental dynamics model---is vital for solving robotic problems, especially in motion-planning and control. However, the powerful few/zero-shot generalization ability that foundation models present may break that requirement. This includes using LLMs to generate task plans~\cite{Ahn_2022_saycan}, using RFMs to learn model-free end-to-end control policies~\cite{brohan2022rt, chebotar2023qtransformer}, etc.

\paragraph{Foundation Models for Task-Specification} 
Task-specifications as language prompts \cite{brohan2022rt, 2023rt2, Ahn_2022_saycan}, goal images \cite{cui2022zest, Jiang2022vima}, videos of humans demonstrating the task \cite{bahl2022human, jain2022transformers}, rewards \cite{tang2023saytap, ma2023eureka}, rough scratch of trajectory~\cite{gu2023rttrajectory}, policy sketches \cite{andreas2017modular}, and hand-drawn images \cite{skubic2007using} have allowed goal specifications in a more natural, human-like format. Multimodal foundation models allow users to not only specify the goal but also help resolve ambiguities via dialogue. Recent work in understanding trust and intent recognition within the human-robot interaction domain has opened up newer paradigms in our understanding of how humans use explicit and implicit cues to convey task-specifications. While significant progress has been made, recent work in prompt engineering for LLMs implies that even with a single modality, it is challenging to generate relevant outputs.  
% [in-context learning?]
Vision-Language Models are proven to be especially good at task-specification, showing potential for resolving this problem in robotics. Extending the idea of vision-language-based task-specifications, explore methods to achieve multi-modal task specification using more natural inputs like images obtained from the internet ~\cite{brohan2022rt} explores this idea of zero-shot transfer from task-agnostic data further, by providing a novel model class that exhibits promising scalable model properties. The model encodes high-dimensional inputs and outputs, including camera images, instructions, and motor commands into compact token representations to enable real-time control of mobile manipulators. 

\paragraph{Foundation Models for Uncertainty and Safety} Though being a critical problem in robotics, uncertainty and safety using foundation models for robotics is still underexplored. Existing works like \texttt{KNOWNO}~\cite{ren2023robots} proposes a framework for measuring and aligning the uncertainty of LLM-based task planners. Recent advances in Chain-of-Thought prompting \cite{wei2022chain}, open vocabulary learning \cite{wu2023towards}, and hallucination recognition in LLMs \cite{cui2023holistic} may open up newer avenues to address these challenges.

%\pagebreak
%  (Quanting, Seungchan, Tianyi and Yaqi) 
\section{Review of Current Experiments and Evaluations} \label{sec:experiments}

In this section, we summarize the datasets, benchmarks, and experiments of the current research works. 

\subsection{Datasets and Benchmarks}

Relying solely on knowledge learned from language and vision datasets is limiting. Some concepts, like friction or weight, are not easily learned through these modalities alone, as suggested by~\cite{gao2023physically} and~\cite{tatiya2024mosaic} in their works on physically grounded VLMs. Therefore, in order to make robotic agents that can better understand the world, researchers are not just adapting foundational models from the language and vision domains; they are also advancing the development of large, diverse, and multimodal \textit{robotic datasets} for training or fine-tuning these foundation models. This effort is now diverging into two directions: collecting data from the real world, versus collecting data from simulations and then transferring it to the real world. Each direction has its pros and cons. We will cover these datasets and simulations in the following paragraphs and discuss their respective advantages and disadvantages.

\subsubsection{Real World Robotics Datasets}
Real-world robotics datasets are highly appealing due to their diverse object classes and multimodal inputs, offering a rich resource for training robotic systems without the need for complex and often inaccurate physical simulations. However, creating these large-scale datasets presents a significant challenge, primarily due to the absence of a substantial ‘data flywheel’ effect. This effect, which greatly benefited fields like CV and NLP through contributions from millions of internet users, is less evident in robotics. The limited incentive for individuals to upload extensive sensory inputs and corresponding action sequences poses a major hurdle in data acquisition. Despite these challenges, current efforts are focused on addressing these gaps. RoboNet~\cite{dasari2020robonet} is a notable effort in this direction, offering a large-scale, diverse dataset across different robotic platforms for multi-robot learning. Bridge Dataset V1~\cite{ebert2021bridge} collects 7200 hours of demonstrations in real household kitchen manipulation tasks, and its following Bridge-V2 \cite{walke2023bridgedata} contains 60,096 trajectories collected across 24 environments on common low-cost robots. Language-Table~\cite{lynch2022interactive} collects 600,000 language-labeled trajectories---an order of magnitude larger than prior available datasets. RT-1~\cite{brohan2022rt} contains 130k episodes that cover 700+ tasks, collected using a fleet of 13 Google mobile manipulation robots, over 17 months. While the aforementioned datasets represent significant advancement over prior lab-scale datasets, offering a relatively large volume of data, they are limited to single modalities or specific robot tasks.
 
To overcome these limitations, some recent initiatives have made notable progress. For example, GNM~\cite{Shah_2023_GNM} successfully integrated six different large-scale navigation datasets, utilizing a unified navigation interface based on waypoints. A recent collaborative effort among various laboratories called RT-X~\cite{embodimentcollaboration2023open} has aimed to standardize data across different datasets, by using a 7-degree-of-freedom end-effector's pose as a universal reference across different embodiments. To offer a more comprehensive data source, RH20T~\cite{fang2024rh20t} collected over 110,000 manipulation episodes across 7 embodiments, covering more than 140 diverse, contact-rich skills. The modalities include RGB, depth, end-link force-torque, tactile, proprioception, audio and language instruction. All sensors are well-calibrated and synchronized.
 
Building on these advancements, the scale of real-world robotics datasets is beginning to grow, albeit still lagging behind the immense volume of internet-scale language and vision corpora. The accessibility of advanced hardware such as the Hello Stretch Robot, Unitree Quadrupeds, and open-source dexterous manipulators~\cite{shaw2023leap} is expected to catalyze this growth. As these technologies become more widely available, they are likely to initiate the desired `data flywheel' effect in Robotics. 

\subsubsection{Robotics Simulators}
While we await the widespread deployment of robotic hardware to gather massive amounts of robot data, another approach is to develop simulators that closely mimic real-world graphics and physics. The advantage of using simulation is the ability to deploy tens of thousands of robot instances in a simulated world, enabling simultaneous data collection.

Simulators focus on different aspects, such as photorealism, physical realism, and human-in-the-loop interactions. For navigation tasks, photorealistic simulators are crucial. AI Habitat addresses this by utilizing realistically-scanned 3D scenes from the Matterport3D~\cite{chang2017matterport3d} and Gibson~\cite{xia2018gibson} datasets. Furthermore, Habitat~\cite{19iccvhabitat} is a simulator that allows AI agents to navigate through various realistic 3D spaces and perform tasks, including object manipulation. It features multiple sensors and handles generic 3D datasets. Habitat 2.0~\cite{szot2022habitat} builds upon the original by introducing dynamic scene modeling, rigid-body physics, and increased speed. Habitat 3.0~\cite{puig2023habitat} further integrates programmable humanoids to enhance the simulation experience. Additionally, the AI2THOR simulator \cite{kolve2017ai2} is another promising framework for photorealistic visual foundation model research, as evidenced in \cite{huang23vlmaps,min2022film}. Other simulators, like Mujoco~\cite{todorov2012mujoco}, focus on creating physically realistic environments for advanced manipulation and locomotion tasks.

Moreover, simulators like AirSim~\cite{shah2017airsim} and the Arrival Autonomous Racing Simulator~\cite{herman2021learn}, both built on Unreal Engine, provide a balance of reasonable physics and photorealism. Ultimately, while the aforementioned simulators excel in various areas, they face common challenges such as parallelism. Simulators like Issac Gym~\cite{makoviychuk2021isaac} and Mujoco 3.0~\cite{mujoco3} have attempted to overcome these challenges by using GPU acceleration to expedite the data-collection process.

Despite the abundance of data available in simulators, there are inherent challenges in their use. Firstly, the domain gap between simulations and the real world makes transferring from sim to real problematic---issues that early works are already seeking to resolve \cite{huang2023went}. Secondly, the diversity of environments and base objects is still lacking. Therefore, to effectively utilize simulations in the future, continuous improvements in these two areas are essential.

\subsection{Analysis of Current Method Evaluation}
\label{analysis}
\begin{figure}
    \centering
    \includegraphics[width=\linewidth]{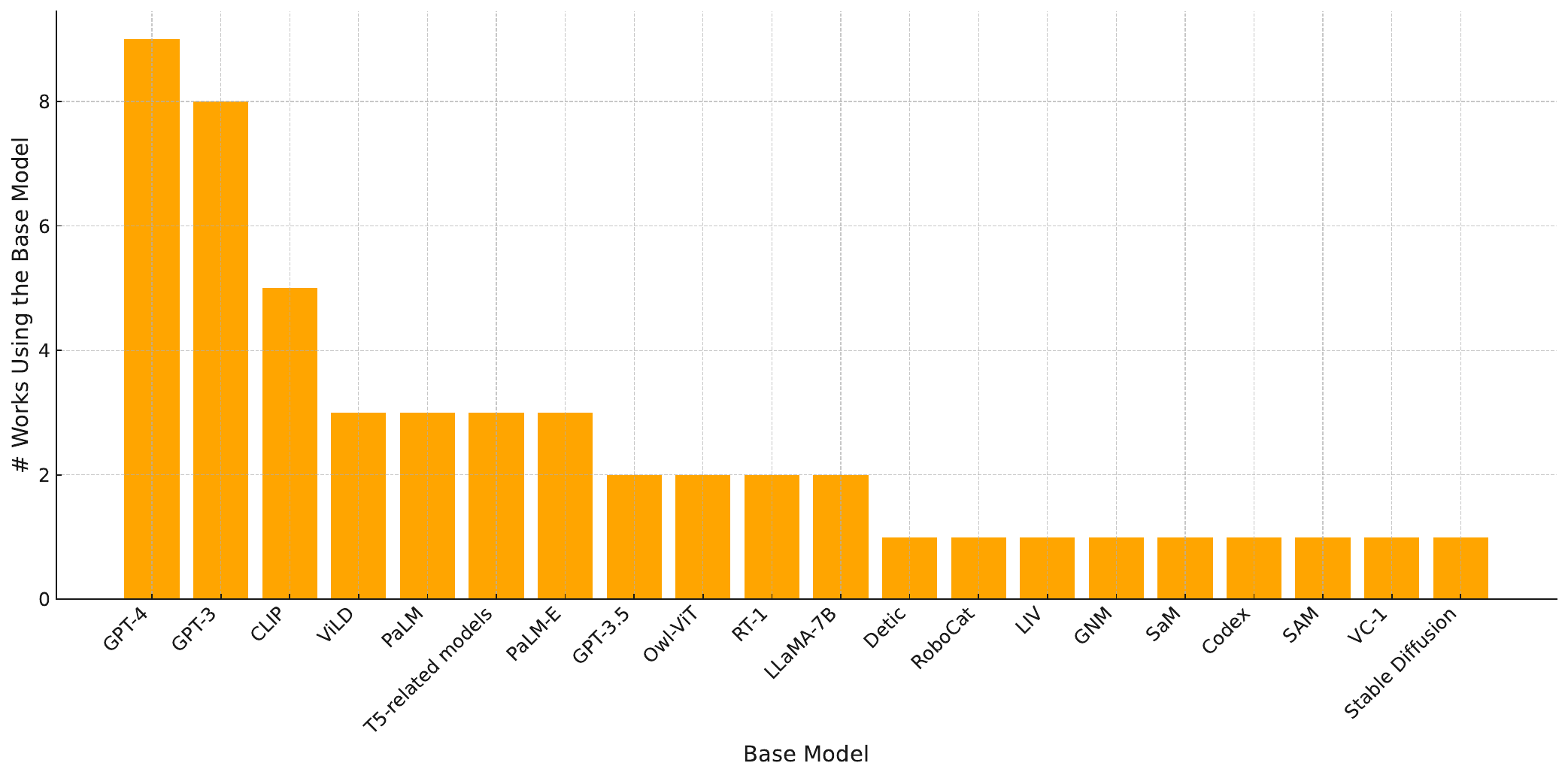}
    \caption{\justifying The histogram showing the number of times different base foundation models are used in developing robotics systems, among the papers we included in this survey. In the plot we can see GPT-4, GPT-3 are among top choices due to their few-shot promptable nature, as well as accessibility through APIs. CLIP and ViLD are frequently used to bridge image and text representations. Apart from CLIP, T5 family models are frequently used to encode text to get text features. PaLM and PaLM-E are used for robot planning. RT-1, which is originally developed for manipulation, emerges as a new base model which other manipulation models are built upon.}
    \label{fig:basemodelhistogram}
\end{figure}

We conduct a meta-analysis of the experiments of papers listed in Tables  \ref{table:manipulation} to \ref{table:multi-tasks} and Figure~\ref{fig:basemodelhistogram}, encouraging readers to consider the following questions
% \zkn{This is great!}:
\begin{enumerate}
    \item What tasks are being solved?
    \item On what datasets or simulators have they been trained? What robot platforms are used for testing?
    \item What foundation models are being utilized? How effectively are the tasks solved?
    \item What base foundation models are more frequently used in these methods?
\end{enumerate}

We summarize several key trends observed in the current literature concerning the experiments conducted:
\paragraph{Imbalanced Focus among Manipulation Tasks:} There is a significant emphasis on general pick-place tasks, particularly tabletop and mobile manipulation. This is likely due to the ease of training for tabletop gripper-based manipulation skills and their potential to form skill libraries that interact with foundation models. However, there is a lack of extensive exploration in low-level action outputs, such as dexterous manipulation and locomotion.

% \paragraph{Imbalanced Focus among Manipulation Tasks:} There is a significant emphasis on general pick-place tasks, particularly for tabletop \cite{zeng2020transporter, Shridhar_2021_CLIPORT} 
% and mobile manipulation \cite{homerobotovmmchallenge2023, parashar2023slap}. This is likely due to the ease of training for tabletop gripper-based manipulation skills and their potential to form skill libraries that interact with foundation models. 
% % There is often an implicit assumption of mostly collision-free space for inverse kinematics for the predicted gripper pose, which also enables the policy to be somewhat embodiment independent \cite{embodimentcollaboration2023open}. 
% However, there is a lack of extensive exploration in situations with clutter \cite{fishman2022motion}, or dexterous manipulation with prehensile \cite{Lee2019IKEAFA, Qin2019KETOLK} and non-prehensile \cite{zhou2023hacman} motion control. 

\paragraph{Need for Improved Generalization and Robustness} Generalization and robustness of end-to-end foundational robotics models have room for improvement. In tabletop manipulation, the use of foundation models leads to performance drops ranging from 21\%~\cite{huang2023voxposer, brohan2022rt} to 31\%~\cite{2023rt2} in unseen tasks. In addition, these models still need improved robust to disturbances, performance drops 14\%~\cite{brohan2022rt} to 18\%~\cite{huang2023voxposer} for similar tasks.

\paragraph{Limited Exploration in Low-Level Actions} There remains a gap in the exploration of direct low-level action outputs. The majority of research focuses on task-level planning and utilizes foundation models with pre-trained or pre-programmed skill libraries. However, existing papers~\cite{2023rt2, embodimentcollaboration2023open, Reed_2022_Gato} that explore low-level action outputs mainly focus on table-top manipulation, where the action space is limited to the end effector's 7 degrees of freedom (DoF). Models that directly output joint angles for tasks like dexterous manipulation and locomotion still require a more thorough research cycle. 

\paragraph{Control Frequencies Too Slow to be Deployed on Real Robots} Most current approaches to robotic control are open-loop, and even those that are closed-loop face limitations in inference speed. These speeds typically range from 1 to 10 Hz, which is considered low for the majority of robotics tasks. Particularly for tasks like humanoid locomotion, a high-frequency control of around 500 Hz is required for the stabilization of the robot's body~\cite{chignoli2021humanoid}.
    
\paragraph{Lack of Uniform Benchmarks for Testing} 
    % The field suffers from a lack of uniform benchmarks for testing methods. 
    The diverse nature of simulations, embodiments, and tasks in robotics leads to varied benchmarks, complicating the comparison of results. Additionally, while success rate is often used as the primary metric, it may not sufficiently evaluate the performance of real-world tasks involving large foundation models, as latency is not captured by the success rate alone. More nuanced evaluation metrics that consider inference time, such as the Compute Aware Success Rate (CASR)~\cite{xie2023reasoning}.

\section{Discussions and Future Directions} \label{sec:discussion}

% Suggestions from Zsolt
% 1. reproducible research, how to research
% 2. etc, scale robotics to CV, NLP, learning from youtube, IL, offline RL?

% \begin{figure} [ht!]
%     \centering
%     \includegraphics[width=\textwidth]{example-image-a}
%     \caption{Statistics of foundation models for robotics in terms of their contribution for major robotic components, we see that we have the most papers in terms of planning.}
%     \label{fig:taxonomy_sensor}
% \end{figure}

% (brief related work, different survey paper, scope of other paper, how ours will be different, additional info, etc)
% \zkn{May want to put this after the large tables since one has to skip across many pages to read on. You can put the small one here maybe to prevent whitespace. Yafei : Done}
\subsection {Remaining Challenges and Open Discussions}
\paragraph{Grounding for Robot Embodiment}

Although numerous strategies have been explored to address the problem of grounding, as discussed in Section \ref{subsec:grounding}, there are many open challenges in this area. 
First, grounding needs an effective medium or interface that bridges concepts and robot actions. Existing interfaces, such as those employing natural language \cite{Ahn_2022_saycan, driess2023PaLMe} and code \cite{codeaspolicy, singh2022progprompt}, are limited. While concepts can be articulated through language and code, they are not universally applicable to nuances such as dexterous body movements. Furthermore, these interfaces often depend on predefined skill libraries that are not only time-intensive to develop but also lack generalization to new environments. Using reward as an interface \cite{yu2023language, xie2023text2reward, ma2023eureka} may alleviate some of the generalization issues in simulations by acquiring skills dynamically. However, the time-consuming and potentially unsafe nature of training RL algorithms in the real world raises questions about the feasibility of this method, with real-world validations of its effectiveness yet to be demonstrated.

% From the data perspective, the type of data necessary for grounding robots is a critical consideration.
Second, we need to move from an unimodal notion of grounding, like mapping the word to meaning to a more holistic  grounding of multiple sensory modalities.  
Approaches that rely solely on visual data \cite{gao2023physically} may capture certain physical properties such as material, transparency, and deformability. Yet, they fall short in grasping concepts like friction, which requires interactive data with proprioceptive feedback, or the scent of an object, which cannot be acquired without additional modalities such as olfaction.

Lastly, we should consider grounding from an embodiment perspective. The same task may necessitate distinct actions based on the robot's embodiment; for example, opening a door would require drastically different maneuvers from a humanoid robot compared to a quadruped. Current research on grounding often emphasizes environmental adaptation while affording less consideration to how embodiment shapes interaction strategies.

\paragraph{Safety and Uncertainty}
As we pursue deployments of real robots to work alongside humans in factories, to provide elderly care, or to assist in homes and offices, these autonomous systems (and the foundation models that power them) will require more effective measures of safety. While formal hardware and software safety checks still apply, the use of foundation models to support provable safety analysis will become an increasingly necessary direction. With the goal of deploying robots to safety-critical scenarios, prior works have considered leveraging Lyapunov-style safety index functions \cite{zhao2021modelfree, herbert2021scalable, chen2021safe}, in attempts to provide hard safety guarantees for nonlinear systems with complex dynamics and external disturbances (see also Section \ref{challenge_safety}). Traditionally, the systems under consideration by the provable safety literature are of %just a few degrees-of-freedom (since the solution for the safety certificates scales exponentially in computational complexity with the dimension of the state space)
low dimension, often require careful specification of a world/dynamics model, require specifying an initial safe set and/or set-boundary distance functions, require some heuristics and training ``tricks" to obtain useful safety value functions that balance conservativeness versus performance, do not naturally support multi-agent settings, and present challenges in \textit{safely} updating the safety value function and growing the safe set online. Herbert \textit{et al.}~\cite{herbert2021scalable} synthesized several techniques into a framework---thereby easing computation, streamlining updates to the safe sets by one or more orders of magnitude compared to the prior art, and extending Hamilton-Jacobi Reachability analysis to 10-dimensional systems that govern quadcopter control. Chen \textit{et al.}~\cite{chen2021safe} combine RL with HJ Reachability analysis to learn safety value functions from high-dimensional inputs (RGB images, plus vehicle state), to trade off a performance-oriented policy and a safety-oriented policy, within a jointly-optimized dual actor-critic framework, for simulated autonomous racing. Tian \textit{et al.}~\cite{tian2022safety} integrate HJ Reachability analysis in the context of multi-agent interactions in urban autonomous driving, by formulating the problem as a general-sum Stackelberg game. 

However, in all of these works, open questions remain on integrating socially-acceptable safety constraints and formal guarantees for systems with robotic foundational models. % fei's thoughts on this?
One of the directions is to formulate safety as an affordance \cite{safetyasaffordance8525627}. The definition of safety changes based on the capability of the robot and social context. Another focus for safety is to ensure robust alignment of the robot's inferred task specification to a human user's communicative intent. Foundation models offer a way to encode the enormous world knowledge, which can serve as commonsense priors to decode the underlying intent. Recent works improve the use of LLMs for robotics with conformal prediction \cite{ren2023robots} and explicit constraint checking \cite{yang2023plug}. % moved from sec 3
Despite these advances, foundation models currently lack native capacity to reason about the uncertainty associated with their outputs. If properly calibrated, uncertainty quantification in foundation models can be used to trigger fall-back safety measures like early termination, pre-defined safe maneuvers, or human-in-the-loop interventions. 

% (FM as intermediate solution)

% (combine with previous section)

% \paragraph{How should we approach foundation models for robots? Should it be a unified solution or modular solution? }
% % \paragraph{Problems of model-free foundation models for planning and control}
% Current foundation models present a scalable solution for generalist robots but these are model-free, that is  
\paragraph{Is there a Dichotomy between End-to-End and Modular Approaches?}

% Could be modular, holistic. multi-platform, multi-task for one module (planning, control), multi-task for multi-module, multi-modal.
The human brain serves as an example of a functional approach to learning and generalization. While neuroscientists have identified specific regions of the brain, such as the visual cortex, somatosensory cortex, and motor cortex, the brain demonstrates remarkable plasticity and the ability to reorganize its functions to adapt to changes or brain lesions. This flexibility suggests that the brain may have evolved to be modular as a consequence of unified training, combining specific functionalities while maintaining the capacity for general learning \cite{sporns2016modular,meunier2010modular}. 
Similarly, in ``Bertology", NLP researchers show how local parts of trained networks can specialize in one area over others. This indicates that certain modules of large-scale models may become highly specialized for specific functions, which can be adapted for downstream tasks without re-training the entire network. This transfer learning approach can lead to more efficient use of computational resources and faster adaptation to new tasks.

In the context of robotics, taking a premature stand for either modular or end-to-end policy architectures may limit the potential of foundation models for robotics. Modular solutions can provide specific biases and enable effective task-specific performance, but they may not fully leverage the potential of general learning and transferability. On the other hand, end-to-end solutions have a history of working well on certain tasks on CV and NLP, but they might not offer the desired flexibility for adaptation to new situations. As \cite{vanhoucke2018endtoend} pointed out, there appears to be a misconception about the modular versus end-to-end dichotomy. This is because the former pertains to architecture while the latter relates to optimization – they are not mutually exclusive.

Regarding the architecture and optimization design for foundation models used in robotics, we can focus on a functional approach rather than categorizing it as either modular or end-to-end differentiable. One of the goals of a robotic foundational model is to allow flexible modular components, each responsible for specific functionalities, with unified learning that leverages shared representations and general learning capabilities.

\paragraph{Adaptability to Physical Changes in Embodiment}
From employing a pen to flip a light switch to maneuvering down a staircase with a broken leg encased in a cast, the human brain demonstrates versatile and adaptable reasoning. It is a single unit that controls perceptual understanding, motion control, and dialogue capabilities. For motion control, it adapts to the changes in the embodiment, due to tool use or injury. This adaptability extends to more profound transformations, such as individuals learning to paint with their feet or mastering musical instruments with specialized prosthetics. We want to build such interactive and adaptable intelligence in Robotics.

In the previous discussions, we saw existing works successfully deploying navigation foundation models for various robot platforms \cite{Shah_2023_ViNT}, such as different wheeled robots and quadrupedal robots. We also witnessed the manipulation foundation model used in different manipulators \cite{2023rt2, arenas2023how} which can be used across different robotic platforms, ranging from tabletop robot arms to mobile manipulators. 
% We think that navigation is higher-level task and learning control for manipulation is relatively lower-level.

One of the key open research question is how robotics foundational models should enable motion control across different physical embodiments \cite{Doshi24-crossformer}. Initial results \cite{Doshi24-crossformer} show the possibility of one model for different policies across different embodiment. 

Robot policies deployed in homes and offices must be robust to mechanical motion failures, such as sensor malfunctions or actuator breakdowns, ensuring continued functionality in challenging environments. % robot dog adaptability papers, genetic algorithms work.
Furthermore, robotic systems must be designed to adapt to a variety of tools and peripherals, mirroring the human capability to interact with different instruments for specific tasks and physical tool uses. 

% While some works \cite{qin2019keto, Turpin2021GIFTGI, Qi2023LearningGT} have explored  learning representations for diverse tool use, these approaches are yet to be scaled up with foundation models. 
% mention about RT-X, RT-2, roboCat, VINT

\paragraph{World Model, or Model-agnostic?}
In classical robotics, especially in planning and control problems, it was common to attempt to model as much as possible about the world that would be needed for robotics tasks. This was often carried out by leveraging structural priors about the tasks, or by relying on heuristics or simplifying assumptions. Certainly, if it was possible to perfectly model the world, solving robotics problems would become a lot simpler. Unfortunately, due to the complexity of the real world, world modeling in Robotics remains extremely difficult and sometimes even intractable. As a consequence, obtaining policies that generalize across tasks and environments remains a core problem.

The foundation models surveyed in this paper mostly take a model-agnostic (model-free) approach, leveraging the strength of expansive datasets and large-scale deep learning architectures. Some exceptions have attempted to emulate model-based approaches by directly employing LLMs as dynamic models. However, these attempts are still constrained by the inherent limitations of text-only descriptions and are prone to encountering issues with hallucinations, as discussed in \cite{hao2023reasoning, xie2023reasoning}. Many researchers would argue~\cite{lecun2022path} that the data-scaled learning paradigm of these foundation models is still quite different from how humanity and animals learn, which is in an extreme data- and energy-efficient manner. Achieving even close to the joint performance and efficiency of human learning ability remains intriguing. In~\cite{lecun2022path}, LeCun argues that one possible answer to resolving that puzzle may lie in the \textit{learning} of world models, a model that predicts how the state of the world going to change as consequences of the actions taken. 

If we were to develop world models that can emulate the precision of the world's representation through rigorous mathematical and physical modeling, it would bring us significantly closer to addressing and generalizing complex issues in robotics. These sophisticated and reliable world models would enable the application of established model-based methodologies, including search-based and sample-based planning, as well as trajectory optimization techniques. This approach would not only facilitate the resolution of planning and control challenges in robotics but also augment the explainability of these processes. It is posited that the pursuit of a 'foundation world model', characterized by remarkable generalization abilities and zero-shot learning capabilities, holds the potential to be a paradigm-shifting development in the field. And we have already seen the reality of 'foundation world model' getting closer with action/language conditioned video generation ~\cite{videoworldsimulators2024, bruce2024genie}

% \textcolor{red}{Yafei: DOTO: consider adding Sora here. DONE}
% similar to a By do the surveying, conclusion and future work?
\paragraph{Novel Robotics Platforms and Multi-sensory Information} 
% data collection and fusion

As demonstrated in Figure \ref{fig:dataset_fig} and the Meta-analysis in Tables \ref{table:manipulation}-\ref{table:multi-tasks}, existing real robot platforms utilized for deploying foundation models are predominantly limited to gripper-based, single-arm robot manipulators. The range of concepts learnable from tasks executed by these hardware systems is restricted, primarily because the simple opening and closing actions of a gripper are easily describable by language. To enable robots to achieve a level of dexterity and motor skills comparable to those of animals and humans, or to perform complex domestic tasks, it is essential for foundation models to acquire a deeper understanding of physical and household concepts. This learning necessitates a broader spectrum of information sources, such as diverse sensors (including smell, tactile, and thermal sensors), and more intricate data such as proprioception data from robot platforms with high degrees of freedom.

Current dexterous manipulators, e.g., Shadow Hand \cite{shadowrohand}, are prohibitively expensive and prone to frequent breakdowns, hence they are predominantly experimented with in simulation. Moreover, tactile sensors are still limited in their application, often confined to the fingertips, as in \cite{Dong_2017}, or offer only low resolution, as observed in the robot-sweater \cite{si2023robotsweater}. Recent progress has started to explore full-hand, high-resolution tactile sensing for humanoid hands \cite{romero2024eyesight}.

Furthermore, since the bulk of data-collection is still conducted through human demonstrations, platforms that facilitate more accurate and efficient data acquisition, such as ALOHA \cite{zhao2023learning}, AirExo \cite{fang2024airexo} and Leap Hands \cite{shaw2023leap}, are gaining popularity. Therefore, we posit that significant contributions are yet to be made---not only in terms of software innovations, but also in hardware. These advancements are crucial for providing richer data-collection and, thus, expanding the conceptual space of robotics foundation models.

\paragraph{Continual Learning} 
% by Vidhi
Continual learning broadly refers to the ability to learn and adapt to dynamic and changing environments. Specifically, it refers to learning algorithms that can learn and adapt to the underlying training data distribution and changing learning objective, as they evolve through time. 

% why continual learning 
Continual learning is challenging, as neural network models often suffer from catastrophic forgetting, leading to a significant decrease in overall model performance on prior tasks.
% One possible reason why this happens is that neural network initialization with pretrained weights may lead to 
% might on newer information,  degradation of previously learned policy
 % Why is continual learning hard in general? —
One naive solution to mitigate performance degradation due to catastrophic forgetting involves periodically re-training models with the entire dataset collected, which generally allows models to avoid forgetting issues, since the process encompasses both old and new data. However, this method demands significant computational and memory resources. In contrast, training or fine-tuning only on new tasks or current data, without revisiting previous data, is less resource-intensive but incurs catastrophic forgetting due to the model's tendency to overwrite previously learned information. This forgetting can be attributed to task interference between old and new data, concept drifts as data distributions evolve over time, and limitations in model expressivity based on their size. 

Additionally, with the increasing capacities of models, continuously re-training them on expanding data corpora becomes less feasible. Recent works in vision and language continual learning \cite{kasai2022realtime, Mehta2022DSIUT, mehta2023empirical, smith2023construct} have proposed various solutions, yet achieving effective continual learning, that can be applied to robotics, still remains a challenging objective.
%\zkn{This sentence isn't correct if I understand it; training on ``all of the hitherto collected dataset'' doesn't have forgetting issues, but of course requires significant compute/memory. Training (fine-tuning) only on the current task doesn't have this but then incurs forgetting.} -- Solved
For continual learning, large pre-trained foundational models currently face the above challenges and more, primarily because their extensive size makes retraining more difficult.  % Why continual learning in robotics is needed? —
In Robotics applications, specifically, continual learning is imperative to the deployability of robot learning policies in diverse environments, yet it is still a largely-unexplored domain. Whereas some recent works have studied various sub-topics of continual learning \cite{Lesort_2020}---e.g., incremental learning \cite{pmlr-v78-maeda17a}, rapid motor adaptation \cite{kumar2021rma}, human-in-the-loop learning \cite{hejna2023inverse, li2023reinforcement}---these solutions are often designed for a single task/platform and do not yet consider foundation models. 

% open questions.
We need continual learning algorithms that are designed with  machine learning fundamentals in mind and practical real-time systems considerations. Some open research problems and viable approaches are:
 (1) mixing different proportions of the prior data distribution when fine-tuning on latest data to alleviate catastrophic forgetting \cite{kumar2022finetuning}, (2) developing efficient prototypes from prior distributions or curriculum to learn new tasks \cite{openai2019solving} for task inference, (3) improving training stability and sample-efficiency of online learning algorithms \cite{mehta2023sample, an2023direct}, and (4) identifying principled ways to seamlessly incorporate large-capacity models into control frameworks (perhaps with hierarchical learning \cite{ajay2023compositional, frans2017meta, Nair2020Hierarchical} / slow-fast control \cite{feichtenhofer2019slowfast}) for real-time inference.
% \zkn{It's not clear to me why \textit{these} are the major things to mention. If possible, would be nice to bolster with citations or reasoning/justification.  partially done by vidhi and daniel. for review}

% by Vidhi
\paragraph{Standardization and Reproducibility}
The robotics community needs to encourage standardized and reproducible research practices to ensure that published findings can be validated and compared by others. To enable reproducibility at scale, we need to bridge the gap between simulated environments and real-world hardware and improve the transferability of ML models. Homerobot \cite{homerobotovmmchallenge2023} is a promising step towards enabling both simulation and hardware platforms for open vocabulary pick-and-place tasks. However, we need to establish standardized task definitions and affordances to handle different robot morphologies, enabling more efficient model development. 

% added by yafei, insipied by Dietor Fox's talk
\paragraph{Simulation or Real-world Data Collection}
The path to robotic foundation models could be diverse. One method to achieve this is through large-scale real-world collected data ~\cite{embodimentcollaboration2023open, khazatsky2024droid}. Another approach, however, is though simulation data and reduce the sim-to-real gap ~\cite{dalal2023optimus}. Simulation has the advantage of providing theoretically unlimited amount of data, training robot learning policies without worrying about safety concerns, and potential efficient training via parallelism. The large-scale training on simulation has also shown good generalization abilities in robot policy learning \cite{2023_Cheng_extremeParkour} and perception models \citep{wen2024foundationposeunified6dpose}. However, building the twin version of the real world also faces various of challenges. We look forward to seeing the potential of both these two approaches to robotic foundation models.

% by dong-ki and shayegan
% add photos here
% Yafei: modified some text
\paragraph{Deployability to Industrial Settings}
Discussions in previous subsections have primarily focused on academic settings, but foundation models for robotics have significant potential to transform unstructured industrial environments such as construction, oil and gas, solar farms, mines, and agriculture (see \cref{fig:industrial_robotics}).
These environments are dynamic, large in scale, and can lack prior maps or external communication (e.g., can be GPS-denied). Additionally, hazardous materials, physical constraints like background noise, and limited visibility (e.g., masks and helmets) pose significant human-robot interaction challenges that bring to surface novel problems that must be addressed in such settings.

\begin{floatingfigure}[r]{0.49\textwidth}
    \centering
    \begin{subfigure}[t]{0.24\linewidth}
        \includegraphics[width=\linewidth]{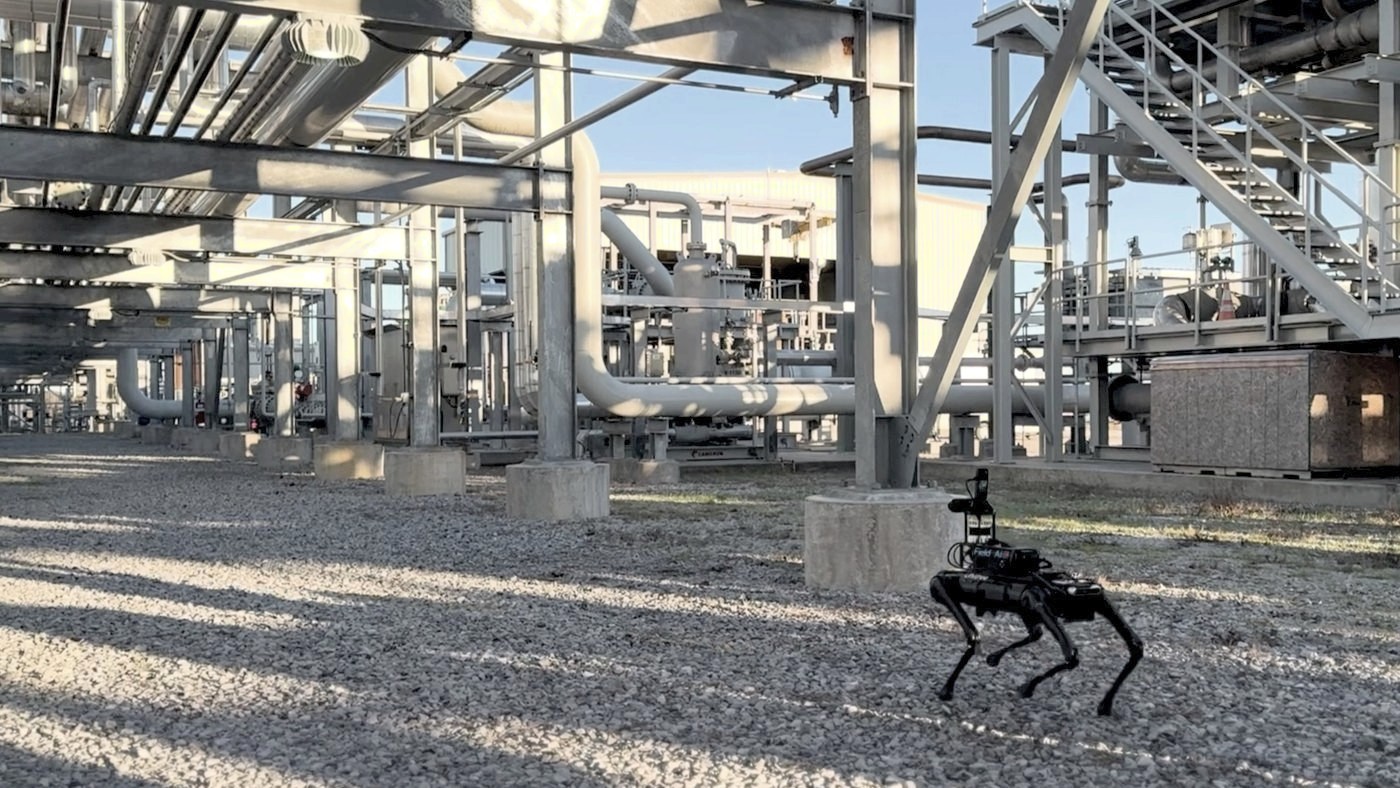}
    \end{subfigure}%
    \hfill%
    \begin{subfigure}[t]{0.24\linewidth}
        \includegraphics[width=\linewidth]{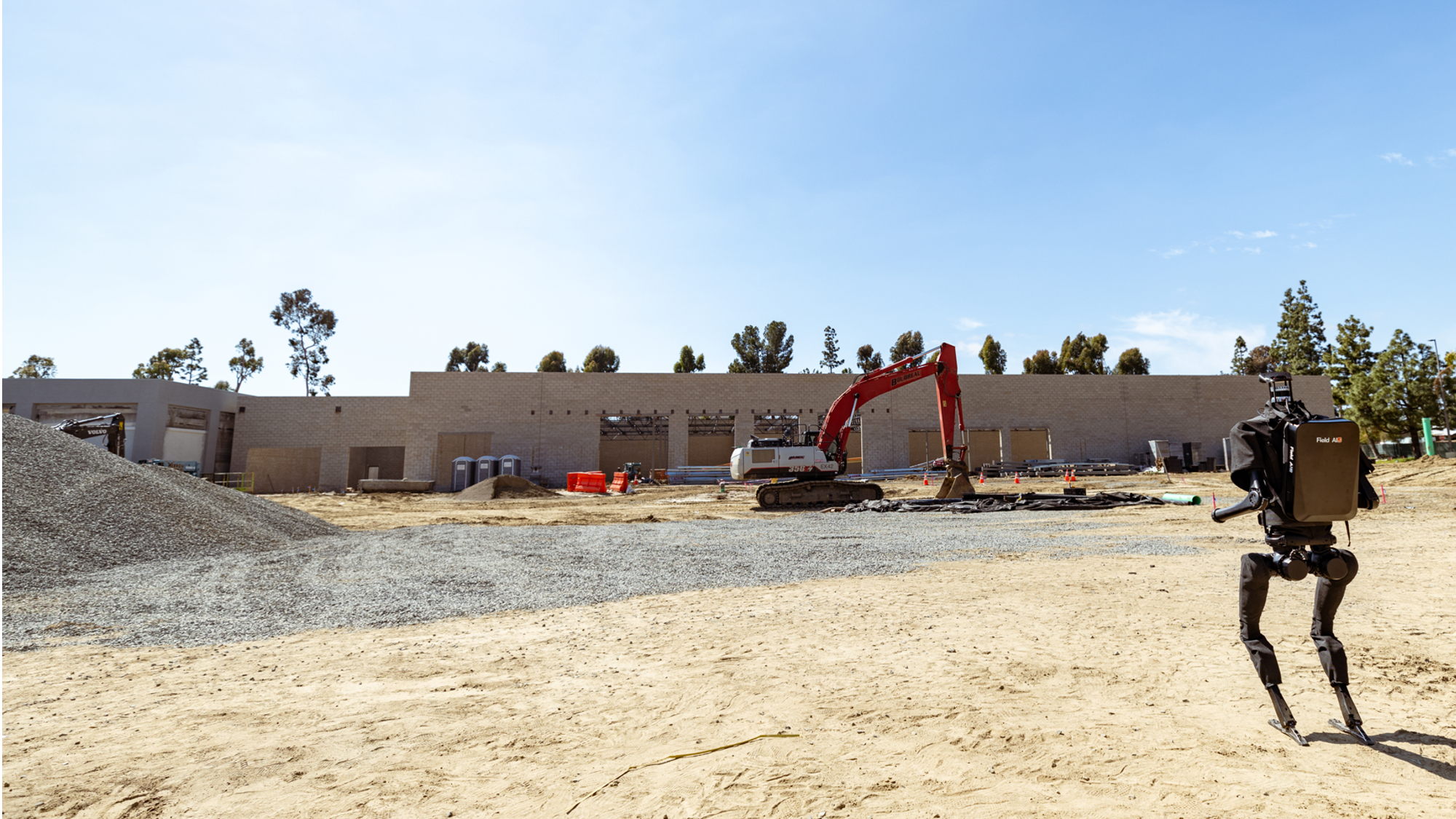}
    \end{subfigure}\\
    \vspace{5pt}%
    \begin{subfigure}[t]{0.24\linewidth}
        \includegraphics[width=\linewidth]{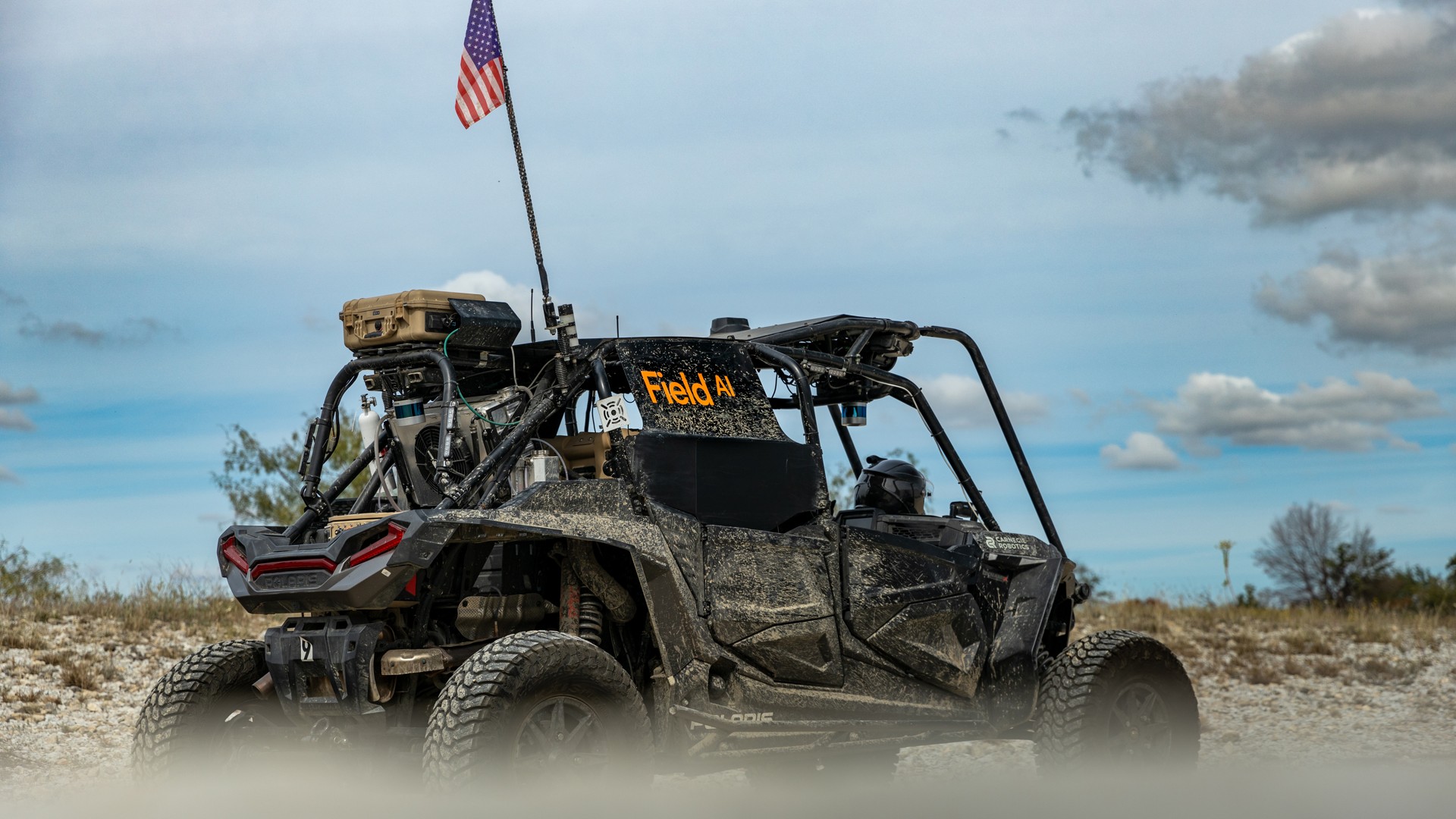}
    \end{subfigure}%
    \hfill%
    \begin{subfigure}[t]{0.24\linewidth}
        \includegraphics[width=\linewidth]{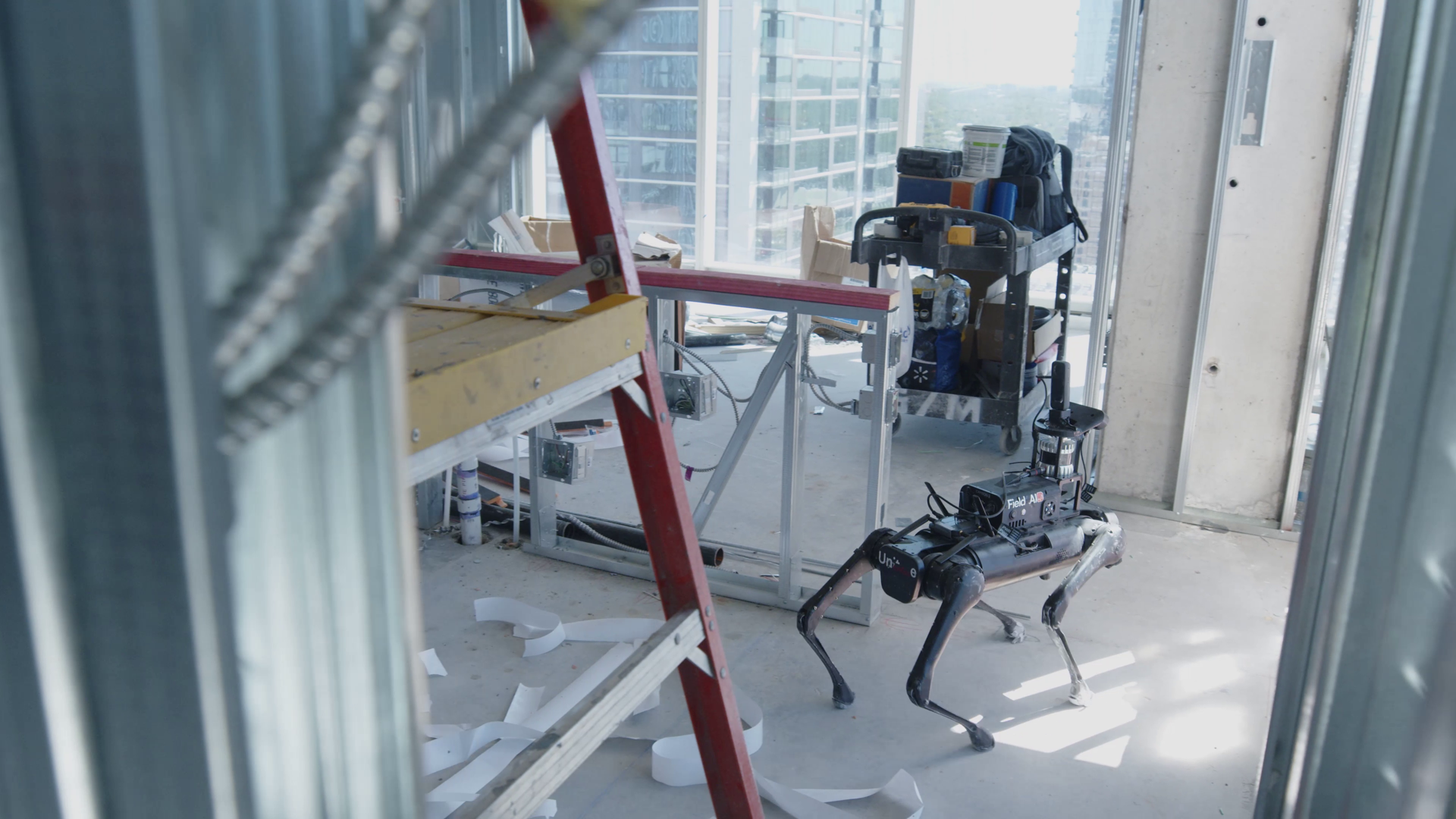}
    \end{subfigure}
    \caption{\justifying Industrial environments present unique challenges like dynamic obstacles, unstructured terrains, and safety risks. Foundation models in robotics offer the potential to navigate, inspect, and optimize operations in such demanding settings, improving safety and efficiency while enabling real-time decision-making and interaction with existing machinery.}
    \label{fig:industrial_robotics}
\end{floatingfigure}

Foundation models could revolutionize industrial operations by enabling robots navigate and inspect demanding outdoor settings, while supporting applications including automated data analysis~\cite{hollmann2024large}, natural language processing for actionable insights~\cite{liu2023reflect}, automated documentation, job site activity recognition~\cite{dang2020sensor}, field operations optimization~\cite{rizk2023case}, and safety violation detection~\cite{goldberg2022characterizing}.
For instance, ConceptFusion~\cite{jatavallabhula2023conceptfusion} and  Open-Fusion~\cite{yamazaki2024open} enable open-set multimodal 3D mapping capabilities, potentially useful for dynamically-changing industrial settings where predefined object categories are unavailable.
Such advancements can enable more flexible and adaptive robot performance in unstructured settings.

However, deploying these models in industrial contexts presents unique challenges.
Ensuring safety and risk awareness to meet stringent industrial safety standards is crucial.
Industrial robots often must adapt to dynamic obstacles and situations, varying surface conditions, uneven slopes, weather conditions, while maintaining awareness of potential hazards, including heavy machinery, all of which must be managed by the foundation model.
For example, in construction environments, robots must navigate ever-changing sites due to ongoing project development, all while making safety-critical decisions~\cite{robla2017working}.
In mining, such foundation models must handle adverse conditions including low visibility, variable terrains, high dust levels, and unstable surfaces~\cite{marshall2016robotics}.
The variability of these real-world environments means that foundation models deployed to such settings must operate in out-of-distribution settings, increasing the risk of unsafe decisions and erroneous outputs.
ConceptGraphs~\cite{gu2023conceptgraphs}, for example, can provide a framework to represent aspects of such environments using 3D scene graphs, while methods inspired by PlayFusion~\cite{chen2023playfusion}, which uses language annotation of unstructured / unguided robotics interaction data, bear potential to support robots in acquiring the necessary skills to interact with machinery.

Deployment also requires edge computing due to potential network limitations and latency while on-site.
These mobile robots, are also constrained by computational resources, operating and collecting data for real-time monitoring.
Techniques such as model quantization~\cite{hubara2018quantized}, distillation~\cite{hinton2015distilling}, and teacher-student based frameworks~\cite{omidshafiei2019teaching,wang2021knowledge} can help address these constraints, though balancing performance with model size remains an ongoing challenge. 

\subsection{Summary}
In this survey paper, we analyzed the current research works on foundation models for robotics based on two major categories: (1) works which apply foundation models to robotic tasks, and (2) works attempting to develop robotics foundation models for robotics tasks using robotics data. We went through the methodologies and experiments of these papers, and provided analysis and insights based on these research works. Furthermore, we specially covered how these foundation models help resolve the common challenges in robotics. Finally, we discussed remaining challenges in robotics that have not been solved by foundation models, as well as other promising research directions.

\section*{Meta Analysis Tables}
Below is a detailed analysis answering the questions raised in section 
\ref{analysis}. We list all the tables by classifying the topics as manipulation in Table \ref{table:manipulation}, dexterous manipulation \ref{table:dextrous}, mobile manipulation \ref{table:mobile_manipulation}, locomotion \ref{table:locomotion}, navigation \ref{table:navigation} and multi-task learning \ref{table:multi-tasks}.

\section*{Disclaimer}
% Due to the large volume of the papers in the areas of foundation models for robotics being published recently, we may not include all the published works. If the readers think we omit some important papers or have inaccuracies or mistakes in the paper, we encourage the readers to email us or send pull requests in the GitHub repo of the paper. 
Due to the rapidly changing nature of the field, we checkpointed this version of literature review on September 1st 2024, and might have missed some relevant work. In addition, due to the rich body of literature and the extensiveness of this survey, there may be inaccuracies or mistakes in the paper. We welcome readers to send pull requests to our GitHub repository (inside \href{\webpage}{\textbf{\color{blue}{\webpage}}}) so we may continue to update our references, correct the mistakes and inaccuracies, as well as updating the entries of the meta studies in the paper. Please refer to the contribution guide in the GitHub repository.

\section*{Acknowledgments}
We would like to thank Vincent Vanhoucke for feedbacks on a draft of this survey paper.
In addition, we would like to thank Yu Quan Chong and Kedi Xu for insightful discussions about the papers list.

\pagebreak

% tables of mata analysis
\begin{table*}[!htbp]  
    \vspace{-1em}  
    \centering
    \caption{Tabletop Manipulation}
    \label{table:manipulation}
    \fontsize{6pt}{8pt}\selectfont
    \adjustbox{max width=\textwidth}{
        % \FloatBarrier
% \begin{table*}
    % \caption{Tabletop Manipulation}
    % \fontsize{6pt}{8pt}\selectfont
    % \label{table:manipulation}
    % \begin{tabular}{@{}|l|l|l|r|l|l|l|l|@{}}
    \begin{tabular*}{\textwidth}{@{\extracolsep{\fill}} l|l|l|c|l|l|l|l}

    \toprule
    \multicolumn{1}{c|}{Title}                           & \multicolumn{1}{c|}{Datasets \& Simulation}                                                                                                                                      & \multicolumn{1}{c|}{Real Robot}                                                            & \begin{tabular}[c]{@{}c@{}}Number\\ of Tasks\end{tabular} & \multicolumn{1}{c|}{Base Model}                                          & \multicolumn{1}{c|}{SR}                                                               & \multicolumn{1}{c|}{SR Descriptions}                                                                                     & Frequency   \\ \hline
    RT-X                                                       & \begin{tabular}[c]{@{}l@{}}Kitchen \\ Manipulation \cite{dass2023clvr},\\ Cable Routing \cite{luo2023multistage},\\ NYU Door \\ Opening \cite{pari2021surprising},\\ AUTOLab UR5 \cite{chen2023berkeley}, \\ Robot Play \cite{rosete2022tacorl},\\ Bridge \cite{walke2023bridgedata},\\ RL Bench (RT-1) \cite{brohan2022rt}\end{tabular} & Google robot & \textcolor{green1}{160266}                                                                         & RT-2(PaLI)                                                                      & \begin{tabular}[c]{@{}l@{}}\textcolor{color30}{30\%}\\ \textcolor{color60}{63\%}\\ \textcolor{color90}{92\%}\end{tabular}        & \begin{tabular}[c]{@{}l@{}}Bridge,\\ Small lab datasets,\\ Google robot\end{tabular}                    &   3-10Hz   \\ \hline 
    RT-2 & \begin{tabular}[c]{@{}l@{}} RL Bench (RT-1) \cite{brohan2022rt} \\ Language-Table \cite{lynch2022interactive}\end{tabular}                                                                                                                                                                                                                                                                                                                                            & Google robot & \textcolor{green2}{480+}                   & PaLI, PaLM-E                                                                      & \begin{tabular}[c]{@{}l@{}}\textcolor{color90}{93\%}\\ \textcolor{color60}{62\%}\end{tabular}               & \begin{tabular}[c]{@{}l@{}}Seen tasks,\\ Unseen tasks\end{tabular}                                         &   1-5Hz    \\ \hline
    RT-1\cite{brohan2022rt}                   & RL Bench (RT-1) \cite{brohan2022rt}                                                                                                                                                                                                                                                                                                                                                                                                          & Google robot & \textcolor{green2}{744}               & RT-1                                                                            & \begin{tabular}[c]{@{}l@{}}\textcolor{color90}{97\%}\\ \textcolor{color80}{76\%}\\ \textcolor{color80}{83\%}\\ \textcolor{color60}{59\%}\end{tabular} & \begin{tabular}[c]{@{}l@{}}Seen Tasks,\\ Unseen Tasks,\\ With Distractors,\\ Novel background\end{tabular} &   3 Hz     \\ \hline
    GNFactor \cite{Ze2023GNFactor} & RL Bench (RT-1) \cite{brohan2022rt} & XArm7 & \textcolor{green3}{166} &  \begin{tabular}[c]{@{}l@{}}Stable Diffusion,\\ CLIP\end{tabular} & \begin{tabular}[c]{@{}l@{}}\textcolor{color30}{32\%}\\ \textcolor{color30}{28\%}\end{tabular} & \begin{tabular}[c]{@{}l@{}}Multi-task test,\\ Unseen tasks\end{tabular} & Unspec. \\ \hline

    MOO \cite{stone2023open}                   &  
    \begin{tabular}[c]{@{}l@{}} Self-created \\ robotic data  \end{tabular}                                                                                                                                                                                                                                                                                                                                                                                                                                           & Google robot & \textcolor{green3}{106}                                                                            & \begin{tabular}[c]{@{}l@{}}Owl-ViT,\\ RT-1\end{tabular}                          & \begin{tabular}[c]{@{}l@{}}\textcolor{color90}{92\%}\\ \textcolor{color80}{75\%}\end{tabular}               & \begin{tabular}[c]{@{}l@{}}Seen Tasks,\\ Unseen Tasks\end{tabular}                                         &  Unspec.  \\ \hline
    PhysObjects \cite{gao2023physically}       & PhysObjects \cite{gao2023physically}                                                                                                                                                                                                                                                                                                                                                                                              & Franka Panda    & \textcolor{green3}{51}                                                                             & \begin{tabular}[c]{@{}l@{}}FlanT5-XXL,\\ GPT-4\end{tabular}                     & \begin{tabular}[c]{@{}l@{}}\textcolor{color90}{88\%}\\ \textcolor{color90}{88\%}\end{tabular}               & \begin{tabular}[c]{@{}l@{}}PhysObjects Test Set,\\ Real scenes\end{tabular}                               &  open loop \\ \hline
    Matcha \cite{zhao2023chat}                & CoppeliaSim \cite{coppeliasim}                                                                                                                                                                                                                                                                                                                                                                                                     & NiCOL           & \textcolor{green3}{50}                                                                             & \begin{tabular}[c]{@{}l@{}}ViLD,\\ text-davinci-003\end{tabular}                 & \begin{tabular}[c]{@{}l@{}}\textcolor{color90}{91\%}\\ \textcolor{color60}{57\%}\end{tabular}               & \begin{tabular}[c]{@{}l@{}}Distinct descriptions,\\ Indistinct descriptions\end{tabular}                   &  open loop \\ \hline
        Scalingup \cite{ha2023scalingup}           & Scalingup benchmark \cite{ha2023scalingup}                                                                                                                                                                                                                                                                                                                                                                                         & UR5 arm         & \textcolor{yellow1}{18}                                                                             & GPT-3                                                                           & \multicolumn{1}{l|}{\textcolor{color80}{79\%}}                                         & Mean Success                                                                                               & 35 Hz    \\ \hline
    F3RM \cite{Shen2023f3rm} & \begin{tabular}[c]{@{}l@{}}PyBullet~\cite{CoumansPybullet} and \\
    Self-created \\ robotic data  \end{tabular}  & Franka Panda & \textcolor{yellow1}{18} & CLIP & \begin{tabular}[c]{@{}l@{}}\textcolor{color80}{78\%}\\ \textcolor{color60}{62\%}\end{tabular} & \begin{tabular}[c]{@{}l@{}}Grasp-Place,\\ Object Manip.\end{tabular} & Unspec.\\ \hline

    VIMA \cite{Jiang2022vima}                 & VIMABench \cite{Jiang2022vima}                                                                                                                                                                                                                                                                                                                                                                                                    &           None      & \textcolor{yellow1}{17}                                                                             & \begin{tabular}[c]{@{}l@{}}pre-trained T5\\ tokenizer\end{tabular}              & \begin{tabular}[c]{@{}l@{}}\textcolor{color80}{81\%}\\ \textcolor{color80}{81\%}\\ \textcolor{color80}{79\%}\\ \textcolor{color50}{49\%}\end{tabular} & \begin{tabular}[c]{@{}l@{}}L1,\\ L2,\\ L3,\\ L4\end{tabular}                                               &  Unspec.  \\ \hline
    Instruct2Act \cite{huang2023instruct2act} & VIMABench \cite{Jiang2022vima}                                                                                                                                                                                                                                                                                                                                                                                                    &       None          & \textcolor{yellow1}{17}                                                                             & \begin{tabular}[c]{@{}l@{}}CLIP,\\ SAM,\\ text-davinci-003,\\ LLaMA\end{tabular} & \textcolor{color80}{84\%}                                                              & Mean Success                                                                                               &  open loop \\ \hline
    VoxPoser \cite{huang2023voxposer}          &\begin{tabular}[c]{@{}l@{}} Saipen \cite{xiang2020sapien}  \\ Where2act \cite{mo2021where2act}  \end{tabular}                                                                                                                                                                                                                                                                                                                                                  & Franka Emika    & \textcolor{yellow1}{13}                                                                             & \begin{tabular}[c]{@{}l@{}}GPT-4,\\ SAM,\\ Owl-ViT\end{tabular}                 & \begin{tabular}[c]{@{}l@{}}\textcolor{color90}{88\%}\\ \textcolor{color70}{70\%}\\ \textcolor{color70}{67\%}\end{tabular}        & \begin{tabular}[c]{@{}l@{}}Static environments,\\ With disturbances,\\ Unseen semarios\end{tabular}        &   5HZ      \\ \hline
    Text2Motion \cite{lin2023text2motion}      & TableEnv \cite{lin2023text2motion}                                                                                                                                                                                                                                                                                                                                                                                                 & Franka Panda    & \textcolor{yellow2}{6}                                                                              & text-davinci-003                                                                 & \multicolumn{1}{l|}{\textcolor{color80}{82\%}}                                         & Mean Success                                                                                               &  open loop 
    \\ \hline 
    GenSim \cite{wang2023gensim}  & \begin{tabular}[c]{@{}l@{}}PyBullet~\cite{CoumansPybullet} and \\
    Self-created \\ robotic data  \end{tabular} & XArm7    & \textcolor{yellow2}{100+} & GPT-4   & \begin{tabular}[c]{@{}l@{}}\textcolor{color90}{53.3\%}\\ \textcolor{color60}{68.8\%}\end{tabular}   &\begin{tabular}[c]{@{}l@{}}50 Tasks \\ 70 Tasks \end{tabular}  &  Unspec. 
    \\ \bottomrule
    \end{tabular*}
%     \end{table*}
% \FloatBarrier
    }
\end{table*}

\begin{table*}[!htbp] 
    \centering
    \caption{Dexterous Manipulation}
    \label{table:dextrous}
    \fontsize{6pt}{8pt}\selectfont
    \adjustbox{max width=\textwidth}{
        % \FloatBarrier
% \begin{table*}
% \fontsize{6pt}{8pt}\selectfont

% \caption{Dexterous Manipulation}
% \label{table:dextrous}
\renewcommand{\arraystretch}{0.8}  % Reduce vertical space
% \begin{tabular}{@{}|l|l|l|l|l|l|l|l|@{}}
\begin{tabular*}{\textwidth}{@{\extracolsep{\fill}} p{3cm}|l|l|c|l|l|l|l}

\toprule
\multicolumn{1}{c|}{Title}                           & \multicolumn{1}{c|}{Datasets \& Simulation}                                                                                                                                      & \multicolumn{1}{c|}{Real Robot}                                                            & \begin{tabular}[c]{@{}c@{}}Number\\ of Tasks\end{tabular} & \multicolumn{1}{c|}{Base Model}                                          & \multicolumn{1}{c|}{SR}                                                               & \multicolumn{1}{c|}{SR Descriptions}                                                                                     & Frequency                                                \\ \hline
RoboCat \cite{bousmalis2023robocat} & \begin{tabular}[c]{@{}l@{}}RGB-Stacking \\ Benchmark \cite{lee2021pickandplace}\end{tabular} & \begin{tabular}[c]{@{}l@{}}Sawyer 7-DoF,\\ Panda 7-DoF, \\ KUKA 14-DoF\end{tabular} & \textcolor{green2}{253}                                  & RoboCat    & \begin{tabular}[c]{@{}l@{}}\textcolor{color80}{82\%},\\ \textcolor{color70}{74\%},\\ \textcolor{color90}{86\%}\end{tabular} & \begin{tabular}[c]{@{}l@{}}Sawyer in sim, \\ Panda in sim,\\ KUKA in real\end{tabular} & \begin{tabular}[c]{@{}l@{}}10 Hz\\ 20 Hz\end{tabular} 
\\ \bottomrule
\end{tabular*}
% \end{table*}
% \FloatBarrier
    }
\end{table*}

\begin{table*}[!htbp]  
    \centering
    \caption{Mobile Manipulation}
    \label{table:mobile_manipulation}
    \fontsize{6pt}{8pt}\selectfont
    \adjustbox{max width=\textwidth}{
        % \begin{table*}
%     \caption{Mobile Manipulation}
%     \label{table:mobile_manipulation}
%     \fontsize{6pt}{8pt}\selectfont
    % \begin{tabular}{@{}|l|l|l|c|l|l|l|l|@{}}
\renewcommand{\arraystretch}{0.9}  % Reduce vertical space
    \begin{tabular*}{\textwidth}{@{\extracolsep{\fill}} l|l|l|c|l|l|l|l}

    \toprule
    \multicolumn{1}{c|}{Title}                           & \multicolumn{1}{c|}{Datasets \& Simulation}                                                                                                                                      & \multicolumn{1}{c|}{Real Robot}                                                            & \begin{tabular}[c]{@{}c@{}}Number\\ of Tasks\end{tabular} & \multicolumn{1}{c|}{Base Model}                                          & \multicolumn{1}{c|}{SR}                                                               & \multicolumn{1}{c|}{SR Descriptions}                                                                                     & Frequency                                \\ \hline
                                                   
    LLaRP \cite{szot2023large}           & \begin{tabular}[c]{@{}l@{}}Language \\ Rearrangement \cite{szot2023large}\end{tabular}                                                       &                                              None                                        & \textcolor{green2}{1000}                                                      & LLaMA-7B                                                                 & \begin{tabular}[c]{@{}l@{}}\textcolor{color40}{42\%}\\ \textcolor{color10}{0\%}\\ \textcolor{color10}{8\%}\\ \textcolor{color40}{39\%}\end{tabular}                       & \begin{tabular}[c]{@{}l@{}}Total,\\ Multiple objects,\\ Spatial Reasoning,\\ Conditional instuct.\end{tabular}        &     open loop         \\ \hline
    Code-As-Policies \cite{codeaspolicy} & \begin{tabular}[c]{@{}l@{}}Customized\\ RoboCodeGen \cite{codeaspolicy},\\ HumanEval \cite{chen2021evaluating}\end{tabular} & \begin{tabular}[c]{@{}l@{}}Eveyday Robots,\\ UR5 Robot arm\end{tabular}              & \textcolor{green2}{214}                                                       & \begin{tabular}[c]{@{}l@{}}Codex,\\ GPT-3\end{tabular}                    & \begin{tabular}[c]{@{}l@{}}\textcolor{color100}{95\%}\\ \textcolor{color100}{96\%}\end{tabular}                                   & \begin{tabular}[c]{@{}l@{}}RoboCodeGen,\\ HumanEval\end{tabular}                                                         &     open loop         \\ \hline
    InnerMonlogue \cite{innermonologe}    & Ravens \cite{zeng2020transporter};                                                                                                            & \begin{tabular}[c]{@{}l@{}}UR5e robot,\\ Google robot\end{tabular}                & \textcolor{green3}{130}                                                       & \begin{tabular}[c]{@{}l@{}}ViLD\\ GPT-3.5\end{tabular}                   & \begin{tabular}[c]{@{}l@{}}\textcolor{color50}{51\%}\\ \textcolor{color50}{50\%}\\ \textcolor{color90}{90\%}\\ \textcolor{color60}{60\%}\end{tabular}                     & \begin{tabular}[c]{@{}l@{}}Seen Ravens tasks,\\ Unseen Ravens tasks,\\ Real robot arm,\\ Google robot\end{tabular}    &     Unspec. \\ \hline
    LIV \cite{ma2023liv}                 & \begin{tabular}[c]{@{}l@{}}MetaWorld \cite{yu2021metaworld}\\ FrankaKitchen \cite{nair2022r3m}\end{tabular}                  & Franka panda                                                                         & \textcolor{green3}{114}                                                       & LIV                                                                      & \begin{tabular}[c]{@{}l@{}}\textcolor{color30}{30\%}\\ \textcolor{color30}{30\%}\\ \textcolor{color50}{45\%}\end{tabular}                            & \begin{tabular}[c]{@{}l@{}}FrankaKitchen,\\ MetaWorld,\\ Real robots\end{tabular}                                        &     15hz             \\ \hline
    SayCan \cite{Ahn_2022_saycan}      &                        \begin{tabular}[c]{@{}l@{}}
    proprietary simulator \\ and self-created \\ robotic data  \end{tabular}                                                                                                                                        & Google robot                                                                      & \textcolor{green3}{101}                                                       & \begin{tabular}[c]{@{}l@{}}PaLM,\\ Flan\end{tabular}                     & \textcolor{color70}{74\%}                                                                                  & Mean Success                                                                                                             &     open loop         \\ \hline
    PaLM-E \cite{driess2023PaLMe}                                                & Lang-table \cite{lynch2022interactive}                                                                                                       & \begin{tabular}[c]{@{}l@{}}Google robot, \\xArm6 \end{tabular}                                                                   & \textcolor{green3}{100}                                                       & PaLM-E                                                                   & \begin{tabular}[c]{@{}l@{}}\textcolor{color80}{83\%}\\ \textcolor{color80}{76\%}\\ \textcolor{color50}{52\%}\\ \textcolor{color90}{91\%}\end{tabular}                 & \begin{tabular}[c]{@{}l@{}}Grasping,\\ Stacking,\\ Lang-table tasks,\\ Google robot\end{tabular}                      & 1-5Hz            \\ \hline
    TidyBot \cite{wu2023tidybot}         & TidyBot \cite{wu2023tidybot}                                                                                                                 & \begin{tabular}[c]{@{}l@{}}TidyBot\end{tabular}                & \textcolor{green3}{96}                                                        & \begin{tabular}[c]{@{}l@{}}ViLD,\\ CLIP ,\\ text-davinci003\end{tabular} & \begin{tabular}[c]{@{}l@{}}\textcolor{color90}{91\%}\\ \textcolor{color80}{83\%}\end{tabular}                               & \begin{tabular}[c]{@{}l@{}}Sim,\\ Real\end{tabular}                                                                      &     open loop         \\ \hline
    LLM-Grop \cite{ding2023task}         & Gazebo \cite{1389727}                                                                                                                        & \begin{tabular}[c]{@{}l@{}}Segway base\\ + UR5e arm\end{tabular}        & \textcolor{yellow2}{8}                                                         & GPT-3                                                                    & 4.08                                                                                  & \begin{tabular}[c]{@{}l@{}}Average human \\ rating(1-5)\end{tabular}                                                     &     open loop         \\ \hline
    LLM+P \cite{liu2023llmp}             & \begin{tabular}[c]{@{}l@{}}International Planning \\ Competition \cite{seipp2022pddl}\end{tabular}                                                 &                                                                  None                    & \textcolor{yellow2}{7}                                                         & GPT-4                                                                    & \begin{tabular}[c]{@{}l@{}}\textcolor{color20}{20\%}\\ \textcolor{color90}{90\%}\\ \textcolor{color10}{0\%}\\ \textcolor{color100}{95\%}\\ \textcolor{color90}{85\%}\\ \textcolor{color20}{20\%}\\ \textcolor{color10}{10\%}\end{tabular} & \begin{tabular}[c]{@{}l@{}}Barman,\\ BLOCKSWORLD,\\ FLOORTILE,\\ GRIPPERS,\\ STORAGE,\\ TERMES,\\ TYREWORLD\end{tabular} &     open loop         \\
    \hline
    HomeRobot \cite{homerobotovmmchallenge2023}         & Habitat \cite{yadav2023habitatmatterport}                                                                                                                        & \begin{tabular}[c]{@{}l@{}}Hello-Robot Stretch\end{tabular}        & \textcolor{color20}{8}                                                         & -                                                                    & 20\%                                                                                 & \begin{tabular}[c]{@{}l@{}}RL baseline\end{tabular}                                                     &     closed loop   \\
    % \hline
    % HomeRobot \cite{homerobotovmmchallenge2023}         & Habitat \cite{yadav2023habitatmatterport}                                                                                                                        & \begin{tabular}[c]{@{}l@{}}Hello-Robot\end{tabular}        & \textcolor{color20}{8}                                                         & -                                                                    & 20\%                                                                                 & \begin{tabular}[c]{@{}l@{}}RL baseline\end{tabular}                                                     &     open loop   
    \bottomrule
    \end{tabular*}
%     \end{table*}
    }
\end{table*}

\begin{table*}[!htbp]  
    \centering
    \caption{Locomotion}
    \fontsize{6pt}{8pt}\selectfont
    \label{table:locomotion}
    \adjustbox{max width=\textwidth}{
       % \FloatBarrier
% \begin{table*}
% \fontsize{8pt}{12pt}\selectfont
% \caption{Locomotion}
% \label{table:locomotion}

\begin{tabular*}{\textwidth}{@{\extracolsep{\fill}} p{2.6cm}|p{2.5cm}|p{1.9cm}|c|p{0.8cm}|p{1.1cm}|p{2.5cm}|p{1.2cm}}
\toprule
\multicolumn{1}{c|}{Title}                           & \multicolumn{1}{c|}{Datasets \& Simulation}                                                                                                                                      & \multicolumn{1}{c|}{Real Robot}                                                            & \begin{tabular}[c]{@{}c@{}}Number\\ of Tasks\end{tabular} & \multicolumn{1}{c|}{Base Model}                                          & \multicolumn{1}{c|}{SR}                                                               & \multicolumn{1}{c|}{SR Descriptions}                                                                                     & Frequency \\ \hline
SayTap \cite{tang2023saytap} & IsaacGym \cite{makoviychuk2021isaac} &   Unitree A1   & \multicolumn{1}{r|}{\textcolor{yellow1}{30}}                                    & GPT-4      & \multicolumn{1}{r|}{\textcolor{color100}{97\%}} & Mean Success                         &     openloop  \\ \hline
Prompt2Walk \cite{wang2023prompt} & Mujoco \cite{todorov2012mujoco} &   None   & \multicolumn{1}{r|}{\textcolor{yellow2}{1}}                                    & GPT-4      & \multicolumn{1}{r|}{\textcolor{color80}{80\%}} & Mean Success                         &     10 Hz  \\ \hline
TokenHumanoid2024 \cite{tokenhumanoid2024} & IsaacGym \cite{makoviychuk2021isaac}, Agility Robotics’ simulator & Agility Robotics Humanoid & \multicolumn{1}{r|}{\textcolor{yellow3}{15}} & GPT-4 & \multicolumn{1}{r|}{\textcolor{color90}{--}} & Tracking error \& Prediction error & openloop \\
\bottomrule
\end{tabular*}
% \end{table*}
% \FloatBarrier

    }
\end{table*}

\begin{table*}[!htbp]
    \centering
    \caption{Navigation}
    \fontsize{6pt}{8pt}\selectfont
    \label{table:navigation}
    \adjustbox{max width=\textwidth}{
       % \begin{table*}
%     \caption{Navigation}
%     \label{table:navigation}
%     \fontsize{8pt}{12pt}\selectfont
\begin{tabular*}{\textwidth}{@{\extracolsep{\fill}} p{2.6cm}|p{2.5cm}|p{1.9cm}|c|p{0.8cm}|p{1.1cm}|p{2.5cm}|p{1.2cm}}
    \toprule
    \multicolumn{1}{c|}{Title}                           & \multicolumn{1}{c|}{Datasets \& Simulation}                                                                                                                                      & \multicolumn{1}{c|}{Real Robot}                                                            & \begin{tabular}[c]{@{}c@{}}Number\\ of Tasks\end{tabular} & \multicolumn{1}{c|}{Base Model}                                          & \multicolumn{1}{c|}{SR}                                                               & \multicolumn{1}{c|}{SR Descriptions}                                                                                     & Frequency           \\ \hline
    LM-Nav \cite{Shah_2022_LMNav}      &                        Self-created datasets                                                                                                 & \begin{tabular}[c]{@{}l@{}}Clearpath \\ Jackal UGV\end{tabular}                   & \textcolor{yellow1}{20}                                                                             & GPT-3                        & \textcolor{color80}{80\%}                    & Mean Success                         &       open loop         \\ \hline
    CLIP-Fields \cite{Shafiullah_2022_CLIPFields} & \begin{tabular}[c]{@{}l@{}}Habitat-Matterport \\ 3D Semantic \\ \cite{yadav2023habitatmatterport}\end{tabular} & \begin{tabular}[c]{@{}l@{}}Stretch\\ Robot\end{tabular}                           & \textcolor{yellow1}{14}                                                                             & \begin{tabular}[c]{@{}l@{}} CLIP \\ Detic \end{tabular} & \textcolor{color80}{79\%}                 & Mean Success                         &       open loop         \\ \hline

    GNM \cite{Shah_2023_GNM}         & GNM datasets \cite{Shah_2023_GNM}                                                                                                                 & \begin{tabular}[c]{@{}l@{}}LoCoBot\\ Vizbot\\ DJI Tello\\ Jackal UGV\end{tabular} & \textcolor{yellow2}{3}                                                                              & GNM                          & \textcolor{color100}{96\%}                 & Mean Success                         &      Unspec. \\ \bottomrule
    \end{tabular*}
%     \end{table*}
    }
\end{table*}

\begin{table*}[!htbp]  
    \centering
    \caption{Multi-Tasks}
    \fontsize{6pt}{8pt}\selectfont
    \label{table:multi-tasks}
    \adjustbox{max width=\textwidth}{
        % \FloatBarrier
% \begin{table*}
%     \caption{Multi-Tasks}
%     \label{table:multi-tasks}
%     \fontsize{6pt}{8pt}\selectfont
%     % \begin{tabular}{@{}|l|l|l|c|l|l|l|l|@{}}
    \begin{tabular*}{\textwidth}{@{\extracolsep{\fill}} l|l|l|c|l|l|l|l}
    \toprule
    \multicolumn{1}{c|}{Title}                           & \multicolumn{1}{c|}{Datasets \& Simulation}                                                                                                                                      & \multicolumn{1}{c|}{Real Robot}                                                            & \begin{tabular}[c]{@{}c@{}}Number\\ of Tasks\end{tabular} & \multicolumn{1}{c|}{Base Model}                                          & \multicolumn{1}{c|}{SR}                                                               & \multicolumn{1}{c|}{SR Descriptions}                                                                                     & Frequency   \\ \hline
   
    Gato \cite{Reed_2022_Gato}                & \begin{tabular}[c]{@{}l@{}}Meta-World \cite{yu2021metaworld},\\ Sokoban \cite{weber2017imagination},\\ BabyAI \cite{chevalier2018babyai},\\ Procgen \\ Benchmark \cite{cobbe2020leveraging},\\ Arcade Learning \\ Environment \cite{bellemare2013arcade},\\ DM Control \\ Suite \cite{tassa2018deepmind}\end{tabular} & Sawyer arm      & \textcolor{green2}{604}                                                                            & GPT-4                        & \begin{tabular}[c]{@{}l@{}} \textcolor{color100}{97\%} \\ \textcolor{color70}{68\%} \\ \textcolor{color80}{80\%} \\ \textcolor{color70}{68\%}\\\textcolor{color100}{super-human} \\ \textcolor{color60}{64\%}\end{tabular}                                       &        \begin{tabular}[c]{@{}l@{}} Meta-World,\\ Sokoban, \\ BabyAI,\\ Procgen, \\ ALE Atari,\\ DM Control Suite\end{tabular}                     & 20 Hz    \\ \hline
    \begin{tabular}[c]{@{}l@{}}Grounded \\ Decoding \cite{huang2023grounded} \end{tabular}& \begin{tabular}[c]{@{}l@{}}2D Maze \cite{chevalierboisvert2018minimalistic},\\ Ravens \cite{zeng2020transporter}\end{tabular}                                                                                                                                                                                                                                                    &       None          & \textcolor{green2}{124}                                                                            & \begin{tabular}[c]{@{}l@{}}GPT-3.5 \\PaLM \end{tabular} & \begin{tabular}[c]{@{}l@{}}\textcolor{color70}{71\%} \\ \textcolor{color50}{46\%}  \\ \textcolor{color100}{95\%} \\ \textcolor{color50}{51\%}\end{tabular}                                                                                                        & \begin{tabular}[c]{@{}l@{}}Ravens Seen Tasks, \\ Ravens Seen Tasks, \\ 2D Maze, \\ Mobile Manipulation \end{tabular}                                                                                                                &     open loop \\ \hline
    Eureka \cite{ma2023eureka}                 & \begin{tabular}[c]{@{}l@{}}Issac Gym \cite{makoviychuk2021isaac}, \\       Bidexterous \\ Manipulation \cite{chen2022towards} \end{tabular}                                                                                                                                                                                                                                                                                                                                                          &          None       & \textcolor{yellow1}{29}                                                                             & GPT-4                        & \begin{tabular}[c]{@{}l@{}} \textcolor{color60}{55\%} \\ 3.7 \\ \textcolor{white}{1} \\ \end{tabular} & \begin{tabular}[c]{@{}l@{}}Bi-Dextrous, \\ Isaac Tasks -- \\ (Human Score)\end{tabular} &     open loop \\ \hline
     Lang2Reward \cite{yu2023language}           & MuJoCo MPC \cite{howell2022predictive}                                                                                                                                                                                                                                                                                                                                                                    & Google robot & \textcolor{yellow1}{17}                                                                             & GPT-4                        & \begin{tabular}[c]{@{}l@{}}\textcolor{color100}{95\%} \\ \textcolor{color80}{80\%} \end{tabular}                       & \begin{tabular}[c]{@{}l@{}}Quadruped,\\ Dextrous manipulator\end{tabular}  &     open loop \\ \hline
    VC-1 \cite{Majumdar2023VC1}                & CortexBench \cite{Majumdar2023VC1}                                                                                                                                                                                                                                                                                                                                                                         &        None         & \textcolor{yellow1}{17}                                                                             & VC-1                         & \begin{tabular}[c]{@{}l@{}}\textcolor{color60}{59\%}\\ \textcolor{color90}{89\%}\\ \textcolor{color70}{67\%}\\ \textcolor{color70}{72\%}\\ \textcolor{color60}{60\%}\\ \textcolor{color70}{70\%}\\ \textcolor{color60}{63\%}\end{tabular}                                                                      & \begin{tabular}[c]{@{}l@{}} Adroit, \\ Meta-World, \\ DMControl, \\ Trifinger, \\ ObjectNav, \\ ImageNav, \\ Mobile Pick\end{tabular}                                                                                                           &     open loop \\ \bottomrule
    \end{tabular*}
% \end{table*}
% \FloatBarrier
    }
\end{table*}

\clearpage
\FloatBarrier

{\small
 \bibliographystyle{unsrt} 
 \bibliography{main}

\begin{thebibliography}{100}

\bibitem{Geiger2013KITTI}
Andreas Geiger, Philip Lenz, Christoph Stiller, and Raquel Urtasun.
\newblock Vision meets robotics: The kitti dataset.
\newblock {\em International Journal of Robotics Research (IJRR)}, 2013.

\bibitem{maturana2018offroad}
Daniel Maturana, Po-Wei Chou, Masashi Uenoyama, and Sebastian Scherer.
\newblock Real-time semantic mapping for autonomous off-road navigation.
\newblock In {\em Field and Service Robotics}, pages 335--350. Springer, 2018.

\bibitem{Calli2017YCB}
Berk Calli, Arjun Singh, James Bruce, Aaron Walsman, Kurt Konolige, Siddhartha Srinivasa, Pieter Abbeel, and Aaron~M Dollar.
\newblock Yale-cmu-berkeley dataset for robotic manipulation research.
\newblock In {\em International Journal of Robotics Research}, page 261 – 268, 2017.

\bibitem{Donahue2014DeCAF}
Jeff Donahue, Yangqing Jia, Oriol Vinyals, Judy Hoffman, Ning Zhang, Eric Tzeng, and Trevor Darrell.
\newblock Decaf: A deep convolutional activation feature for generic visual recognition.
\newblock In {\em ICML}, 2014.

\bibitem{Tzeng2017ADDA}
Eric Tzeng, Judy Hoffman, Kate Saenko, and Trevor Darrell.
\newblock Adversarial discriminative domain adaptation.
\newblock In {\em CVPR}, 2017.

\bibitem{shen2020learning}
William Shen, Felipe Trevizan, and Sylvie Thi{\'e}baux.
\newblock Learning domain-independent planning heuristics with hypergraph networks.
\newblock In {\em Proceedings of the International Conference on Automated Planning and Scheduling}, volume~30, pages 574--584, 2020.

\bibitem{kim2020learning}
Beomjoon Kim and Luke Shimanuki.
\newblock Learning value functions with relational state representations for guiding task-and-motion planning.
\newblock In {\em Conference on Robot Learning}, pages 955--968. PMLR, 2020.

\bibitem{Williams2016MPPI}
Grady Williams, Paul Drews, Brian Goldfain, James~M. Rehg, and Evangelos~A. Theodorou.
\newblock Aggressive driving with model predictive path integral control.
\newblock In {\em ICRA}, 2016.

\bibitem{qureshi2019mpnet}
Ahmed~H Qureshi, Yinglong Miao, Anthony Simeonov, and Michael~C Yip.
\newblock Motion planning networks: Bridging the gap between learning-based and classical motion planners.
\newblock {\em IEEE Transactions on Robotics}, pages 1--9, 2020.

\bibitem{fishman2022mpinets}
Adam Fishman, Adithyavairavan Murali, Clemens Eppner, Bryan Peele, Byron Boots, and Dieter Fox.
\newblock Motion policy networks.
\newblock In {\em Proceedings of the 6th Conference on Robot Learning (CoRL)}, 2022.

\bibitem{2020_Peng_LearningAgileImitating}
Xue~Bin Peng, Erwin Coumans, Tingnan Zhang, Tsang-Wei Lee, Jie Tan, and Sergey Levine.
\newblock Learning agile robotic locomotion skills by imitating animals.
\newblock In {\em RSS}, 2020.

\bibitem{2016_Levine_end2endDeepVisuomotor}
Sergey Levine, Chelsea Finn, Trevor Darrell, and Pieter Abbeel.
\newblock End-to-end training of deep visuomotor policies.
\newblock In {\em Journal of Machine Learning Research}, 2016.

\bibitem{2019_Hwangbo_LearningAgile}
Jemin Hwangbo, Joonho Lee, Alexey Dosovitskiy, Dario Bellicoso, Vassilios Tsounis, Vladlen Koltun, and Marco Hutter.
\newblock Learning agile and dynamic motor skills for legged robots.
\newblock In {\em Science Robotics}, 30 Jan 2019.

\bibitem{2019_Nagabandi_DeepDynamicsModels}
Anusha Nagabandi, Kurt Konoglie, Sergey Levine, and Vikash Kumar.
\newblock Deep dynamics models for learning dexterous manipulation.
\newblock In {\em CoRL}, 2019.

\bibitem{Kalashnkov2021MTOPT}
Dmitry Kalashnkov, Jake Varley, Yevgen Chebotar, Ben Swanson, Rico Jonschkowski, Chelsea Finn, Sergey Levine, and Karol Hausman.
\newblock Mt-opt: Continuous multi-task robotic reinforcement learning at scale.
\newblock {\em arXiv:2104.08212}, 2021.

\bibitem{jang2021bc}
Eric Jang, Alex Irpan, Mohi Khansari, Daniel Kappler, Frederik Ebert, Corey Lynch, Sergey Levine, and Chelsea Finn.
\newblock {{BC}-Z: Zero-Shot Task Generalization with Robotic Imitation Learning}.
\newblock In {\em 5th Annual Conference on Robot Learning}, 2021.

\bibitem{brown2020language}
Tom~B. Brown, Benjamin Mann, Nick Ryder, Melanie Subbiah, Jared Kaplan, Prafulla Dhariwal, Arvind Neelakantan, Pranav Shyam, Girish Sastry, Amanda Askell, Sandhini Agarwal, Ariel Herbert-Voss, Gretchen Krueger, Tom Henighan, Rewon Child, Aditya Ramesh, Daniel~M. Ziegler, Jeffrey Wu, Clemens Winter, Christopher Hesse, Mark Chen, Eric Sigler, Mateusz Litwin, Scott Gray, Benjamin Chess, Jack Clark, Christopher Berner, Sam McCandlish, Alec Radford, Ilya Sutskever, and Dario Amodei.
\newblock Language models are few-shot learners, 2020.

\bibitem{Ramesh2022DALLE2}
Aditya Ramesh, Prafulla Dhariwal~Alex Nichol, Casey Chu, and Mark Chen.
\newblock Hierarchical text-conditional image generation with clip latents.
\newblock {\em arXiv preprint arXiv:2204.06125}, 2022.

\bibitem{Saharia2022ImaGen}
Chitwan Saharia, William Chan, Saurabh Saxena, Lala Li, Jay Whang, Emily Denton, Seyed Kamyar, Seyed Ghasemipour, Burcu~Karagol Ayan, S.~Sara Mahdavi, Rapha~Gontijo Lopes, Tim Salimans, Jonathan Ho, David~J Fleet, and Mohammad Norouzi.
\newblock Photorealistic text-to-image diffusion models with deep language understanding.
\newblock {\em arXiv preprint arXiv:2205.11487}, 2022.

\bibitem{2021_Caron_DINO}
Mathilde Caron, Hugo Touvron, Ishan Misra, Herv\'e J\'egou, Julien Mairal, Piotr Bojanowski, and Armand Joulin.
\newblock Emerging properties in self-supervised vision transformers.
\newblock In {\em Proceedings of the International Conference on Computer Vision (ICCV)}, 2021.

\bibitem{Oquab_2023_DINOv2}
Maxime Oquab, Timothée Darcet, Theo Moutakanni, Huy~V. Vo, Marc Szafraniec, Vasil Khalidov, Pierre Fernandez, Daniel Haziza, Francisco Massa, Alaaeldin El-Nouby, Russell Howes, Po-Yao Huang, Hu~Xu, Vasu Sharma, Shang-Wen Li, Wojciech Galuba, Mike Rabbat, Mido Assran, Nicolas Ballas, Gabriel Synnaeve, Ishan Misra, Herve Jegou, Julien Mairal, Patrick Labatut, Armand Joulin, and Piotr Bojanowski.
\newblock Dinov2: Learning robust visual features without supervision, 2023.

\bibitem{kirillov2023segany}
Alexander Kirillov, Eric Mintun, Nikhila Ravi, Hanzi Mao, Chloe Rolland, Laura Gustafson, Tete Xiao, Spencer Whitehead, Alexander~C. Berg, Wan-Yen Lo, Piotr Doll{\'a}r, and Ross Girshick.
\newblock Segment anything.
\newblock {\em arXiv:2304.02643}, 2023.

\bibitem{Radford_2021_CLIP}
Alec Radford, Jong~Wook Kim, Chris Hallacy, Aditya Ramesh, Gabriel Goh, Sandhini Agarwal, Girish Sastry, Amanda Askell, Pamela Mishkin, Jack Clark, Gretchen Krueger, and Ilya Sutskever.
\newblock Learning transferable visual models from natural language supervision.
\newblock In {\em ICML}, 2021.

\bibitem{Alayrac2022FlamingoAV}
Jean-Baptiste Alayrac, Jeff Donahue, Pauline Luc, Antoine Miech, Iain Barr, Yana Hasson, Karel Lenc, Arthur Mensch, Katie Millican, Malcolm Reynolds, Roman Ring, Eliza Rutherford, Serkan Cabi, Tengda Han, Zhitao Gong, Sina Samangooei, Marianne Monteiro, Jacob Menick, Sebastian Borgeaud, Andy Brock, Aida Nematzadeh, Sahand Sharifzadeh, Mikolaj Binkowski, Ricardo Barreira, Oriol Vinyals, Andrew Zisserman, and Karen Simonyan.
\newblock Flamingo: a visual language model for few-shot learning.
\newblock {\em ArXiv}, abs/2204.14198, 2022.

\bibitem{bommasani2021opportunities}
Rishi Bommasani, Drew~A Hudson, Ehsan Adeli, Russ Altman, Simran Arora, Sydney von Arx, Michael~S Bernstein, Jeannette Bohg, Antoine Bosselut, Emma Brunskill, et~al.
\newblock On the opportunities and risks of foundation models.
\newblock {\em arXiv preprint arXiv:2108.07258}, 2021.

\bibitem{Ahn_2022_saycan}
Michael Ahn, Anthony Brohan, Noah Brown, Yevgen Chebotar, Omar Cortes, Byron David, Chelsea Finn, Chuyuan Fu, Keerthana Gopalakrishnan, Karol Hausman, et~al.
\newblock Do as i can, not as i say: Grounding language in robotic affordances.
\newblock {\em arXiv preprint arXiv:2204.01691}, 2022.

\bibitem{Chen_2022_nlmapsaycan}
Boyuan Chen, Fei Xia, Brian Ichter, Kanishka Rao, Keerthana Gopalakrishnan, Michael~S. Ryoo, Austin Stone, and Daniel Kappler.
\newblock Open-vocabulary queryable scene representations for real world planning.
\newblock In {\em arXiv:2209.09874}, 2022.

\bibitem{tang2023saytap}
Yujin Tang, Wenhao Yu, Jie Tan, Heiga Zen, Aleksandra Faust, and Tatsuya Harada.
\newblock Saytap: Language to quadrupedal locomotion, 2023.

\bibitem{brohan2022rt}
Anthony Brohan, Noah Brown, Justice Carbajal, Yevgen Chebotar, Joseph Dabis, Chelsea Finn, Keerthana Gopalakrishnan, Karol Hausman, Alex Herzog, Jasmine Hsu, et~al.
\newblock Rt-1: Robotics transformer for real-world control at scale.
\newblock {\em arXiv preprint arXiv:2212.06817}, 2022.

\bibitem{2023rt2}
Anthony Brohan, Noah Brown, Justice Carbajal, Yevgen Chebotar, Xi~Chen, Krzysztof Choromanski, Tianli Ding, Danny Driess, Avinava Dubey, Chelsea Finn, Pete Florence, Chuyuan Fu, Montse~Gonzalez Arenas, Keerthana Gopalakrishnan, Kehang Han, Karol Hausman, Alex Herzog, Jasmine Hsu, Brian Ichter, Alex Irpan, Nikhil Joshi, Ryan Julian, Dmitry Kalashnikov, Yuheng Kuang, Isabel Leal, Lisa Lee, Tsang-Wei~Edward Lee, Sergey Levine, Yao Lu, Henryk Michalewski, Igor Mordatch, Karl Pertsch, Kanishka Rao, Krista Reymann, Michael Ryoo, Grecia Salazar, Pannag Sanketi, Pierre Sermanet, Jaspiar Singh, Anikait Singh, Radu Soricut, Huong Tran, Vincent Vanhoucke, Quan Vuong, Ayzaan Wahid, Stefan Welker, Paul Wohlhart, Jialin Wu, Fei Xia, Ted Xiao, Peng Xu, Sichun Xu, Tianhe Yu, and Brianna Zitkovich.
\newblock Rt-2: Vision-language-action models transfer web knowledge to robotic control.
\newblock In {\em arXiv preprint arXiv:2307.15818}, 2023.

\bibitem{Shah_2023_ViNT}
Dhruv Shah, Ajay Sridhar, Nitish Dashora, Kyle Stachowicz, Kevin Black, Noriaki Hirose, and Sergey Levine.
\newblock Vint: A foundation model for visual navigation.
\newblock In {\em arxiv preprint arXiv:2306.14846}, 2023.

\bibitem{Reed_2022_Gato}
Scott Reed, Konrad Zolna, Emilio Parisotto, Sergio~Gomez Colmenarejo, Alexander Novikov, Gabriel Barth-Maron, Mai Gimenez, Yury Sulsky, Jackie Kay, Jost~Tobias Springenberg, Tom Eccles, Jake Bruce, Ali Razavi, Ashley Edwards, Nicolas Heess, Yutian Chen, Raia Hadsell, Oriol Vinyals, Mahyar Bordbar, and Nando de~Freitas.
\newblock A generalist agent.
\newblock In {\em Transactions on Machine Learning Research (TMLR)}, 2022.

\bibitem{driess2023PaLMe}
Danny Driess, F.~Xia, Mehdi S.~M. Sajjadi, Corey Lynch, Aakanksha Chowdhery, Brian Ichter, Ayzaan Wahid, Jonathan Tompson, Quan~Ho Vuong, Tianhe Yu, Wenlong Huang, Yevgen Chebotar, Pierre Sermanet, Daniel Duckworth, Sergey Levine, Vincent Vanhoucke, Karol Hausman, Marc Toussaint, Klaus Greff, Andy Zeng, Igor Mordatch, and Peter~R. Florence.
\newblock Pa{L}{M}-{E}: An embodied multimodal language model.
\newblock {\em ArXiv}, abs/2303.03378, 2023.

\bibitem{kim2024openVLA}
Moo~Jin Kim, Karl Pertsch, Siddharth Karamcheti, Ted Xiao, Ashwin Balakrishna, Suraj Nair, Rafael Rafailov, Ethan Foster, Grace Lam, Pannag Sanketi, Quan Vuong, Thomas Kollar, Benjamin Burchfiel, Russ Tedrake, Dorsa Sadigh, Sergey Levine, Percy Liang, and Chelsea Finn.
\newblock Openvla: An open-source vision-language-action model, 2024.

\bibitem{Kaddour2023Challenges}
Jean Kaddour, Joshua Harris, Maximilian Mozes, Herbie Bradley, Roberta Raileanu, and Robert McHardy.
\newblock Challenges and applications of large language models.
\newblock {\em arXiv:2307.10169}, 2023.

\bibitem{Zhang2023Text}
Chenshuang Zhang, Chaoning Zhang, Mengchun Zhang, and In~So Kweon.
\newblock Text-to-image diffusion models in generative ai: A survey.
\newblock {\em arXiv:2303.07909}, 2023.

\bibitem{2023Harnessing}
Jingfeng Yang, Hongye Jin, Ruixiang Tang, Xiaotian Han, Qizhang Feng, Haoming Jiang, Bing Yin, and Xia Hu.
\newblock Harnessing the power of llms in practice: A survey on chatgpt and beyond.
\newblock {\em arXiv:2304.13712}, 2023.

\bibitem{Yang2023SurveyFMDecision}
Sherry Yang, Ofir Nachum, Yilun Du, Jason Wei, Pieter Abbeel, and Dale Schuurmans.
\newblock Foundation models for decision making: Problems, methods, and opportunities.
\newblock {\em arXiv:2303.04129}, 2023.

\bibitem{zhang2023survey}
Chaoning Zhang, Fachrina~Dewi Puspitasari, Sheng Zheng, Chenghao Li, Yu~Qiao, Taegoo Kang, Xinru Shan, Chenshuang Zhang, Caiyan Qin, Francois Rameau, Lik-Hang Lee, Sung-Ho Bae, and Choong~Seon Hong.
\newblock A survey on segment anything model (sam): Vision foundation model meets prompt engineering, 2023.

\bibitem{awais2023foundational}
Muhammad Awais, Muzammal Naseer, Salman Khan, Rao~Muhammad Anwer, Hisham Cholakkal, Mubarak Shah, Ming-Hsuan Yang, and Fahad~Shahbaz Khan.
\newblock Foundational models defining a new era in vision: A survey and outlook, 2023.

\bibitem{Du2022ASurvey}
Yifan Du, Zikang Liu, Junyi Li, and Wayne~Xin Zhao.
\newblock A survey of vision-language pre-trained models.
\newblock {\em IJCAI-2022 survey track}, 2022.

\bibitem{Gu2023ASystematic}
Jindong Gu, Zhen Han, Shuo Chen, Ahmad Beirami, Bailan He, Ruotong~Liao Gengyuan~Zhang, Yao Qin, Volker Tresp, and Philip Torr.
\newblock A systematic survey of prompt engineering on vision-language foundation models.
\newblock {\em arXiv:2307.12980}, 2023.

\bibitem{firoozi2023foundation}
Roya Firoozi, Johnathan Tucker, Stephen Tian, Anirudha Majumdar, Jiankai Sun, Weiyu Liu, Yuke Zhu, Shuran Song, Ashish Kapoor, Karol Hausman, Brian Ichter, Danny Driess, Jiajun Wu, Cewu Lu, and Mac Schwager.
\newblock Foundation models in robotics: Applications, challenges, and the future, 2023.

\bibitem{Wang2023SurveyLLMAutonomous}
Lei Wang, Chen Ma, Xueyang Feng, Zeyu Zhang, Hao Yang, Jingsen Zhang, Zhiyuan Chen, Jiakai Tang, Xu~Chen, Yankai Lin, Wayne~Xin Zhao, Zhewei Wei, and Ji-Rong Wen.
\newblock A survey on large language model based autonomous agents.
\newblock {\em arXiv:2308.11432}, 2023.

\bibitem{Lin2023LLMsEmbodiedNavigation}
Jinzhou Lin, Han Gao, Rongtao Xu, Changwei Wang, Man Zhang, Li~Guo, and Shibiao Xu.
\newblock The development of llms for embodied navigation.
\newblock In {\em IEEE/ASME TRANSACTIONS ON MECHATRONICS}, volume~1, Sept. 2023.

\bibitem{majumdar2023robotics}
Anirudha Majumdar.
\newblock Robotics: An idiosyncratic snapshot in the age of llms, 8 2023.

\bibitem{xiao2023robot}
Xuan Xiao, Jiahang Liu, Zhipeng Wang, Yanmin Zhou, Yong Qi, Qian Cheng, Bin He, and Shuo Jiang.
\newblock Robot learning in the era of foundation models: A survey, 2023.

\bibitem{arenas2023how}
Montserrat~Gonzalez Arenas, Ted Xiao, Sumeet Singh, Vidhi Jain, Allen~Z. Ren, Quan Vuong, Jake Varley, Alexander Herzog, Isabel Leal, Sean Kirmani, Dorsa Sadigh, Vikas Sindhwani, Kanishka Rao, Jacky Liang, and Andy Zeng.
\newblock How to prompt your robot: A promptbook for manipulation skills with code as policies.
\newblock In {\em 2nd Workshop on Language and Robot Learning: Language as Grounding}, 2023.

\bibitem{huang2023went}
Peide Huang, Xilun Zhang, Ziang Cao, Shiqi Liu, Mengdi Xu, Wenhao Ding, Jonathan Francis, Bingqing Chen, and Ding Zhao.
\newblock What went wrong? closing the sim-to-real gap via differentiable causal discovery.
\newblock In {\em 7th Annual Conference on Robot Learning}, 2023.

\bibitem{herman2021learn}
James Herman, Jonathan Francis, Siddha Ganju, Bingqing Chen, Anirudh Koul, Abhinav Gupta, Alexey Skabelkin, Ivan Zhukov, Max Kumskoy, and Eric Nyberg.
\newblock Learn-to-race: A multimodal control environment for autonomous racing.
\newblock In {\em Proceedings of the IEEE/CVF International Conference on Computer Vision}, pages 9793--9802, 2021.

\bibitem{francis2022learn}
Jonathan Francis, Bingqing Chen, Siddha Ganju, Sidharth Kathpal, Jyotish Poonganam, Ayush Shivani, Vrushank Vyas, Sahika Genc, Ivan Zhukov, Max Kumskoy, et~al.
\newblock Learn-to-race challenge 2022: Benchmarking safe learning and cross-domain generalisation in autonomous racing.
\newblock {\em ICML Workshop on Safe Learning for Autonomous Driving}, 2022.

\bibitem{francis2022core}
Jonathan Francis, Nariaki Kitamura, Felix Labelle, Xiaopeng Lu, Ingrid Navarro, and Jean Oh.
\newblock Core challenges in embodied vision-language planning.
\newblock {\em Journal of Artificial Intelligence Research}, 74:459--515, 2022.

\bibitem{francis2022knowledge}
Jonathan Francis.
\newblock {\em Knowledge-enhanced Representation Learning for Multiview Context Understanding}.
\newblock PhD thesis, Carnegie Mellon University, 2022.

\bibitem{yenamandra2024towards}
Sriram Yenamandra, Arun Ramachandran, Mukul Khanna, Karmesh Yadav, Jay Vakil, Andrew Melnik, Michael B{\"u}ttner, Leon Harz, Lyon Brown, Gora~Chand Nandi, et~al.
\newblock Towards open-world mobile manipulation in homes: Lessons from the neurips 2023 homerobot open vocabulary mobile manipulation challenge.
\newblock {\em arXiv preprint arXiv:2407.06939}, 2024.

\bibitem{tatiya2023transferring}
Gyan Tatiya, Jonathan Francis, and Jivko Sinapov.
\newblock Transferring implicit knowledge of non-visual object properties across heterogeneous robot morphologies.
\newblock In {\em 2023 IEEE International Conference on Robotics and Automation (ICRA)}, pages 11315--11321. IEEE, 2023.

\bibitem{tatiya2023crosstool}
Gyan Tatiya, Jonathan Francis, and Jivko Sinapov.
\newblock Cross-tool and cross-behavior perceptual knowledge transfer for grounded object recognition.
\newblock {\em arXiv preprint arXiv:2303.04023}, 2023.

\bibitem{Sun_2020_CVPR}
Pei Sun, Henrik Kretzschmar, Xerxes Dotiwalla, Aurelien Chouard, Vijaysai Patnaik, Paul Tsui, James Guo, Yin Zhou, Yuning Chai, Benjamin Caine, Vijay Vasudevan, Wei Han, Jiquan Ngiam, Hang Zhao, Aleksei Timofeev, Scott Ettinger, Maxim Krivokon, Amy Gao, Aditya Joshi, Yu~Zhang, Jonathon Shlens, Zhifeng Chen, and Dragomir Anguelov.
\newblock Scalability in perception for autonomous driving: Waymo open dataset.
\newblock In {\em Proceedings of the IEEE/CVF Conference on Computer Vision and Pattern Recognition (CVPR)}, June 2020.

\bibitem{2018_Kalashnikov_QT-Opt}
Dmitry Kalashnikov, Alex Irpan, Peter Pastor, Julian Ibarz, Alexander Herzog, Eric Jang, Deirdre Quillen, Ethan Holly, Mrinal Kalakrishnan, Vincent Vanhoucke, and Sergey Levine.
\newblock Qt-opt: Scalable deep reinforcement learning for vision-based robotic manipulation.
\newblock In {\em CoRL}, 2018.

\bibitem{2016_Levine_LearningHandEye}
Sergey Levine, Peter Pastor, Alex Krizhevsky, and Deirdre Quillen.
\newblock Learning hand-eye coordination for robotic grasping with deep learning and large-scale data collection.
\newblock In {\em arXiv:1603.02199}, 2016.

\bibitem{rlscale2023rss}
Alexander Herzog*, Kanishka Rao*, Karol Hausman*, Yao Lu*, Paul Wohlhart*, Mengyuan Yan, Jessica Lin, Montserrat~Gonzalez Arenas, Ted Xiao, Daniel Kappler, Daniel Ho, Jarek Rettinghouse, Yevgen Chebotar, Kuang-Huei Lee, Keerthana Gopalakrishnan, Ryan Julian, Adrian Li, Chuyuan~Kelly Fu, Bob Wei, Sangeetha Ramesh, Khem Holden, Kim Kleiven, David Rendleman, Sean Kirmani, Jeff Bingham, Jon Weisz, Ying Xu, Wenlong Lu, Matthew Bennice, Cody Fong, David Do, Jessica Lam, Yunfei Bai, Benjie Holson, Michael Quinlan, Noah Brown, Mrinal Kalakrishnan, Julian Ibarz, Peter Pastor, and Sergey Levine.
\newblock Deep rl at scale: Sorting waste in office buildings with a fleet of mobile manipulators.
\newblock In {\em Robotics: Science and Systems (RSS)}, 2023.

\bibitem{todorov2012mujoco}
Emanuel Todorov, Tom Erez, and Yuval Tassa.
\newblock Mujoco: A physics engine for model-based control.
\newblock In {\em 2012 IEEE/RSJ international conference on intelligent robots and systems}, pages 5026--5033. IEEE, 2012.

\bibitem{makoviychuk2021isaac}
Viktor Makoviychuk, Lukasz Wawrzyniak, Yunrong Guo, Michelle Lu, Kier Storey, Miles Macklin, David Hoeller, Nikita Rudin, Arthur Allshire, Ankur Handa, and Gavriel State.
\newblock Isaac gym: High performance gpu-based physics simulation for robot learning, 2021.

\bibitem{mittal2023orbit}
Mayank Mittal, Calvin Yu, Qinxi Yu, Jingzhou Liu, Nikita Rudin, David Hoeller, Jia~Lin Yuan, Pooria~Poorsarvi Tehrani, Ritvik Singh, Yunrong Guo, Hammad Mazhar, Ajay Mandlekar, Buck Babich, Gavriel State, Marco Hutter, and Animesh Garg.
\newblock Orbit: A unified simulation framework for interactive robot learning environments, 2023.

\bibitem{savva2019habitat}
Manolis Savva, Abhishek Kadian, Oleksandr Maksymets, Yili Zhao, Erik Wijmans, Bhavana Jain, Julian Straub, Jia Liu, Vladlen Koltun, Jitendra Malik, et~al.
\newblock Habitat: A platform for embodied ai research.
\newblock In {\em Proceedings of the IEEE/CVF international conference on computer vision}, pages 9339--9347, 2019.

\bibitem{szot2022habitat}
Andrew Szot, Alex Clegg, Eric Undersander, Erik Wijmans, Yili Zhao, John Turner, Noah Maestre, Mustafa Mukadam, Devendra Chaplot, Oleksandr Maksymets, Aaron Gokaslan, Vladimir Vondrus, Sameer Dharur, Franziska Meier, Wojciech Galuba, Angel Chang, Zsolt Kira, Vladlen Koltun, Jitendra Malik, Manolis Savva, and Dhruv Batra.
\newblock Habitat 2.0: Training home assistants to rearrange their habitat, 2022.

\bibitem{puig2023habitat}
Xavier Puig, Eric Undersander, Andrew Szot, Mikael~Dallaire Cote, Tsung-Yen Yang, Ruslan Partsey, Ruta Desai, Alexander~William Clegg, Michal Hlavac, So~Yeon Min, Vladimír Vondruš, Theophile Gervet, Vincent-Pierre Berges, John~M. Turner, Oleksandr Maksymets, Zsolt Kira, Mrinal Kalakrishnan, Jitendra Malik, Devendra~Singh Chaplot, Unnat Jain, Dhruv Batra, Akshara Rai, and Roozbeh Mottaghi.
\newblock Habitat 3.0: A co-habitat for humans, avatars and robots, 2023.

\bibitem{embodimentcollaboration2023open}
Embodiment Collaboration.
\newblock Open x-embodiment: Robotic learning datasets and rt-x models, 2023.

\bibitem{fang2024rh20t}
Hao-Shu Fang, Hongjie Fang, Zhenyu Tang, Jirong Liu, Chenxi Wang, Junbo Wang, Haoyi Zhu, and Cewu Lu.
\newblock {RH20T}: A comprehensive robotic dataset for learning diverse skills in one-shot.
\newblock In {\em IEEE international conference on robotics and automation (ICRA)}. IEEE, 2024.

\bibitem{shafiullah2023bringing}
Nur Muhammad~Mahi Shafiullah, Anant Rai, Haritheja Etukuru, Yiqian Liu, Ishan Misra, Soumith Chintala, and Lerrel Pinto.
\newblock On bringing robots home.
\newblock {\em arXiv preprint arXiv:2311.16098}, 2023.

\bibitem{arcari2023bayesian}
Elena Arcari, Maria~Vittoria Minniti, Anna Scampicchio, Andrea Carron, Farbod Farshidian, Marco Hutter, and Melanie~N. Zeilinger.
\newblock Bayesian multi-task learning mpc for robotic mobile manipulation, 2023.

\bibitem{Cao2021TARE}
Chao Cao, Hongbiao Zhu, Howie Choset, and Ji~Zhang.
\newblock {TARE: A Hierarchical Framework for Efficiently Exploring Complex 3D Environments}.
\newblock In {\em ICRA}, 2023.

\bibitem{islam2020provably}
Fahad Islam, Oren Salzman, Aditya Agarwal, and Maxim Likhachev.
\newblock Provably constant-time planning and replanning for real-time grasping objects off a conveyor belt.
\newblock In {\em RSS}, 2020.

\bibitem{Saxena_2021}
Dhruv~Mauria Saxena, Muhammad~Suhail Saleem, and Maxim Likhachev.
\newblock Manipulation planning among movable obstacles using physics-based adaptive motion primitives.
\newblock In {\em 2021 IEEE International Conference on Robotics and Automation (ICRA)}. IEEE, May 2021.

\bibitem{Garrett2010InteTAMP}
Caelan~Reed Garrett, Rohan Chitnis, Rachel Holladay, Beomjoon Kim, Tom Silver, Leslie~Pack Kaelbling, and Tomas Lozano-P´erez.
\newblock {Integrated Task and Motion Planning}.
\newblock In {\em arXiv:2010.01083}, 2020.

\bibitem{curtis2022long}
Aidan Curtis, Xiaolin Fang, Leslie~Pack Kaelbling, Tomas Lozano-Perez, and Caelan~Reed Garrett.
\newblock Long-horizon manipulation of unknown objects via task and motion planning with estimated affordances.
\newblock In {\em “IEEE International Conference on Robotics and Automation (ICRA)“}, 2022.

\bibitem{liu2024okrobot}
Peiqi Liu, Yaswanth Orru, Chris Paxton, Nur Muhammad~Mahi Shafiullah, and Lerrel Pinto.
\newblock Ok-robot: What really matters in integrating open-knowledge models for robotics.
\newblock {\em arXiv preprint arXiv:2401.12202}, 2024.

\bibitem{cui2022zest}
Yuchen Cui, Scott Niekum, Abhinav Gupta, Vikash Kumar, and Aravind Rajeswaran.
\newblock Can foundation models perform zero-shot task specification for robot manipulation?
\newblock In {\em Learning for Dynamics and Control Conference}, pages 893--905. PMLR, 2022.

\bibitem{ma2023eureka}
Yecheng~Jason Ma, William Liang, Guanzhi Wang, De-An Huang, Osbert Bastani, Dinesh Jayaraman, Yuke Zhu, Linxi Fan, and Anima Anandkumar.
\newblock Eureka: Human-level reward design via coding large language models.
\newblock In {\em 2nd Workshop on Language and Robot Learning: Language as Grounding}, 2023.

\bibitem{gawlikowski2023survey}
Jakob Gawlikowski, Cedrique Rovile~Njieutcheu Tassi, Mohsin Ali, Jongseok Lee, Matthias Humt, Jianxiang Feng, Anna Kruspe, Rudolph Triebel, Peter Jung, Ribana Roscher, et~al.
\newblock A survey of uncertainty in deep neural networks.
\newblock {\em Artificial Intelligence Review}, 56(Suppl 1):1513--1589, 2023.

\bibitem{li2022interpretable}
Xuhong Li, Haoyi Xiong, Xingjian Li, Xuanyu Wu, Xiao Zhang, Ji~Liu, Jiang Bian, and Dejing Dou.
\newblock Interpretable deep learning: Interpretation, interpretability, trustworthiness, and beyond.
\newblock {\em Knowledge and Information Systems}, 64(12):3197--3234, 2022.

\bibitem{chi2020just}
Ta-Chung Chi, Minmin Shen, Mihail Eric, Seokhwan Kim, and Dilek Hakkani-tur.
\newblock Just ask: An interactive learning framework for vision and language navigation.
\newblock In {\em Proceedings of the AAAI Conference on Artificial Intelligence}, volume~34, pages 2459--2466, 2020.

\bibitem{angelopoulos2021gentle}
Anastasios~N Angelopoulos and Stephen Bates.
\newblock A gentle introduction to conformal prediction and distribution-free uncertainty quantification.
\newblock {\em arXiv preprint arXiv:2107.07511}, 2021.

\bibitem{ren2023robots}
Allen~Z. Ren, Anushri Dixit, Alexandra Bodrova, Sumeet Singh, Stephen Tu, Noah Brown, Peng Xu, Leila Takayama, Fei Xia, Jake Varley, Zhenjia Xu, Dorsa Sadigh, Andy Zeng, and Anirudha Majumdar.
\newblock Robots that ask for help: Uncertainty alignment for large language model planners.
\newblock In {\em 7th Annual Conference on Robot Learning}, 2023.

\bibitem{wabersich2023data}
Kim~P Wabersich, Andrew~J Taylor, Jason~J Choi, Koushil Sreenath, Claire~J Tomlin, Aaron~D Ames, and Melanie~N Zeilinger.
\newblock Data-driven safety filters: Hamilton-jacobi reachability, control barrier functions, and predictive methods for uncertain systems.
\newblock {\em IEEE Control Systems Magazine}, 43(5):137--177, 2023.

\bibitem{hsu2023safety}
Kai-Chieh Hsu, Haimin Hu, and Jaime~Fern{\'a}ndez Fisac.
\newblock The safety filter: A unified view of safety-critical control in autonomous systems.
\newblock {\em arXiv preprint arXiv:2309.05837}, 2023.

\bibitem{ames2019control}
Aaron~D Ames, Samuel Coogan, Magnus Egerstedt, Gennaro Notomista, Koushil Sreenath, and Paulo Tabuada.
\newblock Control barrier functions: Theory and applications.
\newblock In {\em 2019 18th European control conference (ECC)}, pages 3420--3431. IEEE, 2019.

\bibitem{bansal2017hamilton}
Somil Bansal, Mo~Chen, Sylvia Herbert, and Claire~J Tomlin.
\newblock Hamilton-jacobi reachability: A brief overview and recent advances.
\newblock In {\em 2017 IEEE 56th Annual Conference on Decision and Control (CDC)}, pages 2242--2253. IEEE, 2017.

\bibitem{chen2021safe}
Bingqing Chen, Jonathan Francis, Jean Oh, Eric Nyberg, and Sylvia~L Herbert.
\newblock Safe autonomous racing via approximate reachability on ego-vision.
\newblock {\em arXiv preprint arXiv:2110.07699}, 2021.

\bibitem{leung2023backpropagation}
Karen Leung, Nikos Ar{\'e}chiga, and Marco Pavone.
\newblock Backpropagation through signal temporal logic specifications: Infusing logical structure into gradient-based methods.
\newblock {\em The International Journal of Robotics Research}, 42(6):356--370, 2023.

\bibitem{gu2022review}
Shangding Gu, Long Yang, Yali Du, Guang Chen, Florian Walter, Jun Wang, Yaodong Yang, and Alois Knoll.
\newblock A review of safe reinforcement learning: Methods, theory and applications.
\newblock {\em arXiv preprint arXiv:2205.10330}, 2022.

\bibitem{dawson2023safe}
Charles Dawson, Sicun Gao, and Chuchu Fan.
\newblock Safe control with learned certificates: A survey of neural lyapunov, barrier, and contraction methods for robotics and control.
\newblock {\em IEEE Transactions on Robotics}, 2023.

\bibitem{bousmalis2023robocat}
Konstantinos Bousmalis, Giulia Vezzani, Dushyant Rao, Coline Devin, Alex~X. Lee, Maria Bauza, Todor Davchev, Yuxiang Zhou, Agrim Gupta, Akhil Raju, Antoine Laurens, Claudio Fantacci, Valentin Dalibard, Martina Zambelli, Murilo Martins, Rugile Pevceviciute, Michiel Blokzijl, Misha Denil, Nathan Batchelor, Thomas Lampe, Emilio Parisotto, Konrad Żołna, Scott Reed, Sergio~Gómez Colmenarejo, Jon Scholz, Abbas Abdolmaleki, Oliver Groth, Jean-Baptiste Regli, Oleg Sushkov, Tom Rothörl, José~Enrique Chen, Yusuf Aytar, Dave Barker, Joy Ortiz, Martin Riedmiller, Jost~Tobias Springenberg, Raia Hadsell, Francesco Nori, and Nicolas Heess.
\newblock Robocat: A self-improving foundation agent for robotic manipulation, 2023.

\bibitem{huang23vlmaps}
Chenguang Huang, Oier Mees, Andy Zeng, and Wolfram Burgard.
\newblock Visual language maps for robot navigation.
\newblock In {\em Proceedings of the IEEE International Conference on Robotics and Automation (ICRA)}, London, UK, 2023.

\bibitem{Shafiullah_2022_CLIPFields}
Nur Muhammad~(Mahi) Shafiullah, Chris Paxton, Lerrel Pinto1~Soumith Chintala, and Arthur Szlam.
\newblock Clip-fields: Weakly supervised semantic fields for robotic memory.
\newblock In {\em RSS}, 2023.

\bibitem{Jatavallabhula_2023_ConceptFusion}
Krishna~Murthy Jatavallabhula, Alihusein Kuwajerwala, Qiao Gu, Mohd Omama, Tao Chen, Shuang Li, Ganesh Iyer, Soroush Saryazd, Nikhil Keetha, Ayush Tewari, Joshua~B. Tenenbaum, Celso~Miguel de~Melo, Madhava Krishna, Liam Paull, Florian Shkurti, and Antonio Torralba.
\newblock Conceptfusion: Open-set multimodal 3d mapping.
\newblock In {\em arXiv:2302.07241}, 2023.

\bibitem{Shah_2022_LMNav}
Dhruv Shah, Blazej Osinski, Brian Ichter, and Sergey Levine.
\newblock Lm-nav: Robotic navigation with large pre-trained models of language, vision, and action.
\newblock In {\em CoRL}, 2022.

\bibitem{Shridhar_2021_CLIPORT}
Mohit Shridhar, Lucas Manuelli, and Dieter Fox.
\newblock Cliport: What and where pathways for robotic manipulation.
\newblock In {\em CoRL}, 2021.

\bibitem{Shen2023f3rm}
William Shen, Ge~Yang, Alan Yu, Jansen Wong, Leslie~Pack Kaelbling, and Phillip Isola.
\newblock Distilled feature fields enable few-shot language-guided manipulation.
\newblock {\em CoRL}, 2023.

\bibitem{Ze2023GNFactor}
Yanjie Ze, Ge~Yan, Yueh-Hua Wu, Annabella Macaluso, Yuying Ge, Jianglong Ye, Nicklas Hansen, Li~Erran Li, and Xiaolong Wang.
\newblock Multi-task real robot learning with generalizable neural feature fields.
\newblock {\em CoRL}, 2023.

\bibitem{homerobotovmmchallenge2023}
Sriram Yenamandra, Arun Ramachandran, Mukul Khanna, Karmesh Yadav, Devendra~Singh Chaplot, Gunjan Chhablani, Alexander Clegg, Theophile Gervet, Vidhi Jain, Ruslan Partsey, Ram Ramrakhya, Andrew Szot, Tsung-Yen Yang, Aaron Edsinger, Charlie Kemp, Binit Shah, Zsolt Kira, Dhruv Batra, Roozbeh Mottaghi, Yonatan Bisk, and Chris Paxton.
\newblock The homerobot open vocab mobile manipulation challenge.
\newblock In {\em Thirty-seventh Conference on Neural Information Processing Systems: Competition Track}, 2023.

\bibitem{Li_2022_LSeg}
Boyi Li, Kilian~Q. Weinberger, Serge Belongie, Vladlen Koltun, and René Ranftl.
\newblock Language-driven semantic segmentation.
\newblock In {\em ICLR}, 2022.

\bibitem{qiuhu2024learning}
Ri-Zhao Qiu*, Yafei Hu*, Ge~Yang, Yuchen Song, Yang Fu, Jianglong Ye, Jiteng Mu, Ruihan Yang, Nikolay Atanasov, Sebastian Scherer, and Xiaolong Wang.
\newblock Learning generalizable feature fields for mobile manipulation, 2024.

\bibitem{gervet2023act3d}
Theophile Gervet, Zhou Xian, Nikolaos Gkanatsios, and Katerina Fragkiadaki.
\newblock Act3d: 3d feature field transformers for multi-task robotic manipulation, 2023.

\bibitem{kassab2023language}
Christina Kassab, Matias Mattamala, Lintong Zhang, and Maurice Fallon.
\newblock Language-extended indoor slam (lexis): A versatile system for real-time visual scene understanding.
\newblock {\em arXiv preprint arXiv:2309.15065}, 2023.

\bibitem{mirjalili2023fm}
Reihaneh Mirjalili, Michael Krawez, and Wolfram Burgard.
\newblock Fm-loc: Using foundation models for improved vision-based localization.
\newblock {\em arXiv:2304.07058}, 2023.

\bibitem{keetha2023anyloc}
Nikhil Keetha, Avneesh Mishra, Jay Karhade, Krishna~Murthy Jatavallabhula, Sebastian Scherer, Madhava Krishna, and Sourav Garg.
\newblock Anyloc: Towards universal visual place recognition.
\newblock {\em RA-L}, 2023.

\bibitem{he2023foundloc}
Yao He, Ivan Cisneros, Nikhil Keetha, Jay Patrikar, Zelin Ye, Ian Higgins, Yaoyu Hu, Parv Kapoor, and Sebastian Scherer.
\newblock Foundloc: Vision-based onboard aerial localization in the wild, 2023.

\bibitem{bohg2017interactive}
Jeannette Bohg, Karol Hausman, Bharath Sankaran, Oliver Brock, Danica Kragic, Stefan Schaal, and Gaurav~S. Sukhatme.
\newblock Interactive perception: Leveraging action in perception and perception in action.
\newblock {\em IEEE Transactions on Robotics (T-RO)}, 2017.

\bibitem{sinapov_grounding_2014}
Jivko Sinapov, Connor Schenck, Kerrick Staley, Vladimir Sukhoy, and Alexander Stoytchev.
\newblock Grounding semantic categories in behavioral interactions: Experiments with 100 objects.
\newblock {\em Robotics and Autonomous Systems}, 62(5):632--645, may 2014.

\bibitem{sinapov_learning_2014}
Jivko Sinapov, Connor Schenck, and Alexander Stoytchev.
\newblock Learning relational object categories using behavioral exploration and multimodal perception.
\newblock In {\em International Conference on Robotics and Automation (ICRA)}, pages 5691--5698, Hong Kong, China, may 2014. IEEE.

\bibitem{gemici_learning_2014}
Mevlana~C. Gemici and Ashutosh Saxena.
\newblock Learning haptic representation for manipulating deformable food objects.
\newblock In {\em Intelligent Robots and Systems (IROS)}, pages 638--645, Chicago, IL, USA, Sep 2014. IEEE.

\bibitem{tatiya2019sensorimotor}
Gyan Tatiya, Ramtin Hosseini, Michael C.~Hughes, and Jivko Sinapov.
\newblock Sensorimotor cross-behavior knowledge transfer for grounded category recognition.
\newblock In {\em International Conference on Development and Learning and Epigenetic Robotics (ICDL-EpiRob)}. IEEE, 2019.

\bibitem{tatiya2020framework}
Gyan Tatiya, Ramtin Hosseini, Michael Hughes, and Jivko Sinapov.
\newblock A framework for sensorimotor cross-perception and cross-behavior knowledge transfer for object categorization.
\newblock {\em Frontiers in Robotics and AI}, 7:137, 2020.

\bibitem{tatiya2020haptic}
Gyan Tatiya, Yash Shukla, Michael Edegware, and Jivko Sinapov.
\newblock Haptic knowledge transfer between heterogeneous robots using kernel manifold alignment.
\newblock In {\em IEEE/RSJ International Conference on Intelligent Robots and Systems (IROS)}. IEEE, 2020.

\bibitem{tatiya2024mosaic}
Gyan Tatiya, Jonathan Francis, Ho-Hsiang Wu, Yonatan Bisk, and Jivko Sinapov.
\newblock Mosaic: Learning unified multi-sensory object property representations for robot learning via interactive perception.
\newblock In {\em 2024 IEEE International Conference on Robotics and Automation (ICRA)}, pages 15381--15387. IEEE, 2024.

\bibitem{Fang2024Uncos}
Xiaolin Fang, Leslie~Pack Kaelbling, and Tomás Lozano-Pérez.
\newblock {Embodied Uncertainty-Aware Object Segmentation}.
\newblock In {\em IEEE/RSJ International Conference on Intelligent Robots and Systems (IROS)}, 2024.

\bibitem{anderson2018vision}
Peter Anderson, Qi~Wu, Damien Teney, Jake Bruce, Mark Johnson, Niko S{\"u}nderhauf, Ian Reid, Stephen Gould, and Anton Van Den~Hengel.
\newblock Vision-and-language navigation: Interpreting visually-grounded navigation instructions in real environments.
\newblock In {\em Proceedings of the IEEE conference on computer vision and pattern recognition}, pages 3674--3683, 2018.

\bibitem{du2023video}
Yilun Du, Mengjiao Yang, Pete Florence, Fei Xia, Ayzaan Wahid, Brian Ichter, Pierre Sermanet, Tianhe Yu, Pieter Abbeel, Joshua~B Tenenbaum, et~al.
\newblock Video language planning.
\newblock {\em arXiv preprint arXiv:2310.10625}, 2023.

\bibitem{ZSP}
Wenlong Huang, Pieter Abbeel, Deepak Pathak, and Igor Mordatch.
\newblock Language models as zero-shot planners: Extracting actionable knowledge for embodied agents, 2022.

\bibitem{lin2023text2motion}
Kevin Lin, Christopher Agia, Toki Migimatsu, Marco Pavone, and Jeannette Bohg.
\newblock Text2motion: From natural language instructions to feasible plans.
\newblock {\em arXiv preprint arXiv:2303.12153}, 2023.

\bibitem{singh2022progprompt}
Ishika Singh, Valts Blukis, Arsalan Mousavian, Ankit Goyal, Danfei Xu, Jonathan Tremblay, Dieter Fox, Jesse Thomason, and Animesh Garg.
\newblock {ProgPrompt}: Generating situated robot task plans using large language models, 2022.

\bibitem{codeaspolicy}
Jacky Liang, Wenlong Huang, Fei Xia, Peng Xu, Karol Hausman, Brian Ichter, Pete Florence, and Andy Zeng.
\newblock Code as policies: Language model programs for embodied control.
\newblock {\em ArXiv}, abs/2209.07753, 2023.

\bibitem{wang2023gensim}
Lirui Wang, Yiyang Ling, Zhecheng Yuan, Mohit Shridhar, Chen Bao, Yuzhe Qin, Bailin Wang, Huazhe Xu, and Xiaolong Wang.
\newblock Gensim: Generating robotic simulation tasks via large language models.
\newblock In {\em CoRL}, 2023.

\bibitem{huang2024rekep}
Wenlong Huang, Chen Wang, Yunzhu Li, Ruohan Zhang, and Li~Fei-Fei.
\newblock Rekep: Spatio-temporal reasoning of relational keypoint constraints for robotic manipulation.
\newblock {\em arXiv preprint arXiv:2409.01652}, 2024.

\bibitem{seipp2022pddl}
J.~Seipp, {\'A}.~Torralba, and J.~Hoffmann.
\newblock Pddl generators, 2022.

\bibitem{rana2023sayplan}
Krishan Rana, Jesse Haviland, Sourav Garg, Jad Abou-Chakra, Ian Reid, and Niko Suenderhauf.
\newblock Sayplan: Grounding large language models using 3d scene graphs for scalable task planning.
\newblock In {\em 7th Annual Conference on Robot Learning}, 2023.

\bibitem{xie2023reasoning}
Quanting Xie, Tianyi Zhang, Kedi Xu, Matthew Johnson-Roberson, and Yonatan Bisk.
\newblock Reasoning about the unseen for efficient outdoor object navigation, 2023.

\bibitem{chen2023not}
Junting Chen, Guohao Li, Suryansh Kumar, Bernard Ghanem, and Fisher Yu.
\newblock How to not train your dragon: Training-free embodied object goal navigation with semantic frontiers.
\newblock {\em arXiv preprint arXiv:2305.16925}, 2023.

\bibitem{huang2023voxposer}
Wenlong Huang, Chen Wang, Ruohan Zhang, Yunzhu Li, Jiajun Wu, and Li~Fei-Fei.
\newblock Voxposer: Composable 3d value maps for robotic manipulation with language models.
\newblock {\em arXiv preprint arXiv:2307.05973}, 2023.

\bibitem{wang2023prompt}
Yen-Jen Wang, Bike Zhang, Jianyu Chen, and Koushil Sreenath.
\newblock Prompt a robot to walk with large language models, 2023.

\bibitem{du2023guiding}
Yuqing Du, Olivia Watkins, Zihan Wang, Cédric Colas, Trevor Darrell, Pieter Abbeel, Abhishek Gupta, and Jacob Andreas.
\newblock Guiding pretraining in reinforcement learning with large language models, 2023.

\bibitem{kwon2023reward}
Minae Kwon, Sang~Michael Xie, Kalesha Bullard, and Dorsa Sadigh.
\newblock Reward design with language models, 2023.

\bibitem{xie2023text2reward}
Tianbao Xie, Siheng Zhao, Chen~Henry Wu, Yitao Liu, Qian Luo, Victor Zhong, Yanchao Yang, and Tao Yu.
\newblock Text2reward: Automated dense reward function generation for reinforcement learning, 2023.

\bibitem{yu2023language}
Wenhao Yu, Nimrod Gileadi, Chuyuan Fu, Sean Kirmani, Kuang-Huei Lee, Montse Gonzalez~Arenas, Hao-Tien Lewis~Chiang, Tom Erez, Leonard Hasenclever, Jan Humplik, Brian Ichter, Ted Xiao, Peng Xu, Andy Zeng, Tingnan Zhang, Nicolas Heess, Dorsa Sadigh, Jie Tan, Yuval Tassa, and Fei Xia.
\newblock Language to rewards for robotic skill synthesis.
\newblock {\em Arxiv preprint arXiv:2306.08647}, 2023.

\bibitem{Cangelosi2010IntegrationOA}
Angelo Cangelosi, Giorgio Metta, Gerhard Sagerer, Stefano Nolfi, Chrystopher Nehaniv, Kerstin Fischer, Jun Tani, Tony Belpaeme, Giulio Sandini, Francesco Nori, and et~al.
\newblock Integration of action and language knowledge: A roadmap for developmental robotics.
\newblock {\em IEEE Transactions on Autonomous Mental Development}, 2(3):167--195, 2010.

\bibitem{deshmukh2023pengi}
Soham Deshmukh, Benjamin Elizalde, Rita Singh, and Huaming Wang.
\newblock Pengi: An audio language model for audio tasks, 2023.

\bibitem{fang2023anygrasp}
Hao-Shu Fang, Chenxi Wang, Hongjie Fang, Minghao Gou, Jirong Liu, Hengxu Yan, Wenhai Liu, Yichen Xie, and Cewu Lu.
\newblock Anygrasp: Robust and efficient grasp perception in spatial and temporal domains.
\newblock {\em IEEE Transactions on Robotics}, 2023.

\bibitem{carta2023grounding}
Thomas Carta, Clément Romac, Thomas Wolf, Sylvain Lamprier, Olivier Sigaud, and Pierre-Yves Oudeyer.
\newblock Grounding large language models in interactive environments with online reinforcement learning, 2023.

\bibitem{gdm2024autort}
Michael Ahn, Debidatta Dwibedi, Chelsea Finn, Montse~Gonzalez Arenas, Keerthana Gopalakrishnan, Karol Hausman, Brian Ichter, Alex Irpan, Nikhil Joshi, Ryan Julian, Sean Kirmani, Isabel Leal, Edward Lee, Sergey Levine, Yao Lu, Isabel Leal, Sharath Maddineni, Kanishka Rao, Dorsa Sadigh, Pannag Sanketi, Pierre Sermanet, Quan Vuong, Stefan Welker, Fei Xia, Ted Xiao, Peng Xu, Steve Xu, and Zhuo Xu.
\newblock Autort: Embodied foundation models for large scale orchestration of robotic agents, 2024.

\bibitem{chen2023palix}
Xi~Chen, Josip Djolonga, Piotr Padlewski, Basil Mustafa, Soravit Changpinyo, Jialin Wu, Carlos~Riquelme Ruiz, Sebastian Goodman, Xiao Wang, Yi~Tay, Siamak Shakeri, Mostafa Dehghani, Daniel Salz, Mario Lucic, Michael Tschannen, Arsha Nagrani, Hexiang Hu, Mandar Joshi, Bo~Pang, Ceslee Montgomery, Paulina Pietrzyk, Marvin Ritter, AJ~Piergiovanni, Matthias Minderer, Filip Pavetic, Austin Waters, Gang Li, Ibrahim Alabdulmohsin, Lucas Beyer, Julien Amelot, Kenton Lee, Andreas~Peter Steiner, Yang Li, Daniel Keysers, Anurag Arnab, Yuanzhong Xu, Keran Rong, Alexander Kolesnikov, Mojtaba Seyedhosseini, Anelia Angelova, Xiaohua Zhai, Neil Houlsby, and Radu Soricut.
\newblock Pali-x: On scaling up a multilingual vision and language model, 2023.

\bibitem{dwibedi2024flexcap}
Debidatta Dwibedi, Vidhi Jain, Jonathan Tompson, Andrew Zisserman, and Yusuf Aytar.
\newblock Flexcap: Generating rich, localized, and flexible captions in images, 2024.

\bibitem{ha2023scalingup}
Huy Ha, Pete Florence, and Shuran Song.
\newblock Scaling up and distilling down: Language-guided robot skill acquisition.
\newblock In {\em Proceedings of the 2023 Conference on Robot Learning}, 2023.

\bibitem{wang2023robogen}
Yufei Wang, Zhou Xian, Feng Chen, Tsun-Hsuan Wang, Yian Wang, Katerina Fragkiadaki, Zackory Erickson, David Held, and Chuang Gan.
\newblock Robogen: Towards unleashing infinite data for automated robot learning via generative simulation, 2023.

\bibitem{Yu_2023_ROSIE}
Tianhe Yu, Ted Xiao, Austin Stone, Jonathan Tompson, Anthony Brohan, Su~Wang, Jaspiar Singh, Clayton Tan, Jodilyn~Peralta Dee~M, Brian Ichter, Karol Hausman, and Fei Xia.
\newblock Scaling robot learning with semantically imagined experience.
\newblock In {\em arXiv:2302.11550}, 2023.

\bibitem{gu2023rttrajectory}
Jiayuan Gu, Sean Kirmani, Paul Wohlhart, Yao Lu, Montserrat~Gonzalez Arenas, Kanishka Rao, Wenhao Yu, Chuyuan Fu, Keerthana Gopalakrishnan, Zhuo Xu, Priya Sundaresan, Peng Xu, Hao Su, Karol Hausman, Chelsea Finn, Quan Vuong, and Ted Xiao.
\newblock Rt-trajectory: Robotic task generalization via hindsight trajectory sketches, 2023.

\bibitem{black2023zeroshot}
Kevin Black, Mitsuhiko Nakamoto, Pranav Atreya, Homer Walke, Chelsea Finn, Aviral Kumar, and Sergey Levine.
\newblock Zero-shot robotic manipulation with pretrained image-editing diffusion models, 2023.

\bibitem{du2023learning}
Yilun Du, Mengjiao Yang, Bo~Dai, Hanjun Dai, Ofir Nachum, Joshua~B. Tenenbaum, Dale Schuurmans, and Pieter Abbeel.
\newblock Learning universal policies via text-guided video generation.
\newblock In {\em NeurIPS}, 2023.

\bibitem{wei2023chainofthought}
Jason Wei, Xuezhi Wang, Dale Schuurmans, Maarten Bosma, Brian Ichter, Fei Xia, Ed~Chi, Quoc Le, and Denny Zhou.
\newblock Chain-of-thought prompting elicits reasoning in large language models, 2023.

\bibitem{dziri2023faith}
Nouha Dziri, Ximing Lu, Melanie Sclar, Xiang~Lorraine Li, Liwei Jiang, Bill~Yuchen Lin, Peter West, Chandra Bhagavatula, Ronan~Le Bras, Jena~D. Hwang, Soumya Sanyal, Sean Welleck, Xiang Ren, Allyson Ettinger, Zaid Harchaoui, and Yejin Choi.
\newblock Faith and fate: Limits of transformers on compositionality, 2023.

\bibitem{besta2023graph}
Maciej Besta, Nils Blach, Ales Kubicek, Robert Gerstenberger, Lukas Gianinazzi, Joanna Gajda, Tomasz Lehmann, Michal Podstawski, Hubert Niewiadomski, Piotr Nyczyk, and Torsten Hoefler.
\newblock Graph of thoughts: Solving elaborate problems with large language models, 2023.

\bibitem{yao2023tree}
Shunyu Yao, Dian Yu, Jeffrey Zhao, Izhak Shafran, Thomas~L. Griffiths, Yuan Cao, and Karthik Narasimhan.
\newblock Tree of thoughts: Deliberate problem solving with large language models, 2023.

\bibitem{zhang2023planning}
Shun Zhang, Zhenfang Chen, Yikang Shen, Mingyu Ding, Joshua~B. Tenenbaum, and Chuang Gan.
\newblock Planning with large language models for code generation.
\newblock In {\em The Eleventh International Conference on Learning Representations}, 2023.

\bibitem{liu2023llmp}
Bo~Liu, Yuqian Jiang, Xiaohan Zhang, Qiang Liu, Shiqi Zhang, Joydeep Biswas, and Peter Stone.
\newblock Llm+p: Empowering large language models with optimal planning proficiency, 2023.

\bibitem{innermonologe}
Wenlong Huang, Fei Xia, Ted Xiao, Harris Chan, Jacky Liang, Pete Florence, Andy Zeng, Jonathan Tompson, Igor Mordatch, Yevgen Chebotar, Pierre Sermanet, Noah Brown, Tomas Jackson, Linda Luu, Sergey Levine, Karol Hausman, and Brian Ichter.
\newblock Inner monologue: Embodied reasoning through planning with language models, 2022.

\bibitem{shafiullah2022behaviorTransformers}
Nur Muhammad~Mahi Shafiullah, Zichen~Jeff Cui, Ariuntuya Altanzaya, and Lerrel Pinto.
\newblock Behavior transformers: Cloning $k$ modes with one stone, 2022.

\bibitem{jain2024vid2robot}
Vidhi Jain, Maria Attarian, Nikhil~J Joshi, Ayzaan Wahid, Danny Driess, Quan Vuong, Pannag~R Sanketi, Pierre Sermanet, Stefan Welker, Christine Chan, Igor Gilitschenski, Yonatan Bisk, and Debidatta Dwibedi.
\newblock Vid2robot: End-to-end video-conditioned policy learning with cross-attention transformers, 2024.

\bibitem{stone2023open}
Austin Stone, Ted Xiao, Yao Lu, Keerthana Gopalakrishnan, Kuang-Huei Lee, Quan Vuong, Paul Wohlhart, Brianna Zitkovich, Fei Xia, Chelsea Finn, et~al.
\newblock Open-world object manipulation using pre-trained vision-language models.
\newblock {\em arXiv preprint arXiv:2303.00905}, 2023.

\bibitem{duan2017one}
Yan Duan, Marcin Andrychowicz, Bradly Stadie, OpenAI Jonathan~Ho, Jonas Schneider, Ilya Sutskever, Pieter Abbeel, and Wojciech Zaremba.
\newblock One-shot imitation learning.
\newblock {\em Advances in neural information processing systems}, 30, 2017.

\bibitem{finn2017one}
Chelsea Finn, Tianhe Yu, Tianhao Zhang, Pieter Abbeel, and Sergey Levine.
\newblock One-shot visual imitation learning via meta-learning.
\newblock In {\em Conference on robot learning}, pages 357--368. PMLR, 2017.

\bibitem{dasari2021transformers}
Sudeep Dasari and Abhinav Gupta.
\newblock Transformers for one-shot visual imitation.
\newblock In {\em Conference on Robot Learning}, pages 2071--2084. PMLR, 2021.

\bibitem{mandi2022towards}
Zhao Mandi, Fangchen Liu, Kimin Lee, and Pieter Abbeel.
\newblock Towards more generalizable one-shot visual imitation learning.
\newblock In {\em 2022 International Conference on Robotics and Automation (ICRA)}, pages 2434--2444. IEEE, 2022.

\bibitem{lynch2020learning}
Corey Lynch, Mohi Khansari, Ted Xiao, Vikash Kumar, Jonathan Tompson, Sergey Levine, and Pierre Sermanet.
\newblock Learning latent plans from play.
\newblock In {\em Conference on robot learning}, pages 1113--1132. PMLR, 2020.

\bibitem{james2018task}
Stephen James, Michael Bloesch, and Andrew~J Davison.
\newblock Task-embedded control networks for few-shot imitation learning.
\newblock In {\em Conference on robot learning}, pages 783--795. PMLR, 2018.

\bibitem{zhou2019watch}
Allan Zhou, Eric Jang, Daniel Kappler, Alex Herzog, Mohi Khansari, Paul Wohlhart, Yunfei Bai, Mrinal Kalakrishnan, Sergey Levine, and Chelsea Finn.
\newblock Watch, try, learn: Meta-learning from demonstrations and reward.
\newblock {\em arXiv preprint arXiv:1906.03352}, 2019.

\bibitem{wang2023mimicplay}
Chen Wang, Linxi Fan, Jiankai Sun, Ruohan Zhang, Li~Fei-Fei, Danfei Xu, Yuke Zhu, and Anima Anandkumar.
\newblock Mimicplay: Long-horizon imitation learning by watching human play.
\newblock {\em arXiv preprint arXiv:2302.12422}, 2023.

\bibitem{yu2018one}
Tianhe Yu, Chelsea Finn, Annie Xie, Sudeep Dasari, Tianhao Zhang, Pieter Abbeel, and Sergey Levine.
\newblock One-shot imitation from observing humans via domain-adaptive meta-learning.
\newblock {\em arXiv preprint arXiv:1802.01557}, 2018.

\bibitem{bonardi2020learning}
Alessandro Bonardi, Stephen James, and Andrew~J Davison.
\newblock Learning one-shot imitation from humans without humans.
\newblock {\em IEEE Robotics and Automation Letters}, 5(2):3533--3539, 2020.

\bibitem{lynch2020language}
Corey Lynch and Pierre Sermanet.
\newblock Language conditioned imitation learning over unstructured data.
\newblock {\em arXiv preprint arXiv:2005.07648}, 2020.

\bibitem{stepputtis2020language}
Simon Stepputtis, Joseph Campbell, Mariano Phielipp, Stefan Lee, Chitta Baral, and Heni Ben~Amor.
\newblock Language-conditioned imitation learning for robot manipulation tasks.
\newblock {\em Advances in Neural Information Processing Systems}, 33:13139--13150, 2020.

\bibitem{rahmatizadeh2018vision}
Rouhollah Rahmatizadeh, Pooya Abolghasemi, Ladislau B{\"o}l{\"o}ni, and Sergey Levine.
\newblock Vision-based multi-task manipulation for inexpensive robots using end-to-end learning from demonstration.
\newblock In {\em 2018 IEEE international conference on robotics and automation (ICRA)}, pages 3758--3765. IEEE, 2018.

\bibitem{Jiang2022vima}
Yunfan Jiang, Agrim Gupta, Zichen Zhang, Guanzhi Wang, Yongqiang Dou, Yanjun Chen, Li~Fei{-}Fei, Anima Anandkumar, Yuke Zhu, and Linxi Fan.
\newblock {VIMA:} general robot manipulation with multimodal prompts.
\newblock {\em ArXiv}, abs/2210.03094, 2022.

\bibitem{lynch2022interactive}
Corey Lynch, Ayzaan Wahid, Jonathan Tompson, Tianli Ding, James Betker, Robert Baruch, Travis Armstrong, and Pete Florence.
\newblock Interactive language: Talking to robots in real time.
\newblock {\em arXiv preprint arXiv:2210.06407}, 2022.

\bibitem{lee2022pi}
Kuang-Huei Lee, Ted Xiao, Adrian Li, Paul Wohlhart, Ian Fischer, and Yao Lu.
\newblock Pi-qt-opt: Predictive information improves multi-task robotic reinforcement learning at scale.
\newblock {\em CoRL}, 2022.

\bibitem{herzog2023deep}
Alexander Herzog, Kanishka Rao, Karol Hausman, Yao Lu, Paul Wohlhart, Mengyuan Yan, Jessica Lin, Montserrat~Gonzalez Arenas, Ted Xiao, Daniel Kappler, Daniel Ho, Jarek Rettinghouse, Yevgen Chebotar, Kuang-Huei Lee, Keerthana Gopalakrishnan, Ryan Julian, Adrian Li, Chuyuan~Kelly Fu, Bob Wei, Sangeetha Ramesh, Khem Holden, Kim Kleiven, David Rendleman, Sean Kirmani, Jeff Bingham, Jon Weisz, Ying Xu, Wenlong Lu, Matthew Bennice, Cody Fong, David Do, Jessica Lam, Yunfei Bai, Benjie Holson, Michael Quinlan, Noah Brown, Mrinal Kalakrishnan, Julian Ibarz, Peter Pastor, and Sergey Levine.
\newblock Deep rl at scale: Sorting waste in office buildings with a fleet of mobile manipulators.
\newblock In {\em RSS}, 2023.

\bibitem{chebotar2023qtransformer}
Yevgen Chebotar, Quan Vuong, Alex Irpan, Karol Hausman, Fei Xia, Yao Lu, Aviral Kumar, Tianhe Yu, Alexander Herzog, Karl Pertsch, Keerthana Gopalakrishnan, Julian Ibarz, Ofir Nachum, Sumedh Sontakke, Grecia Salazar, Huong~T Tran, Jodilyn Peralta, Clayton Tan, Deeksha Manjunath, Jaspiar Singht, Brianna Zitkovich, Tomas Jackson, Kanishka Rao, Chelsea Finn, and Sergey Levine.
\newblock Q-transformer: Scalable offline reinforcement learning via autoregressive q-functions.
\newblock In {\em CoRL}, 2023.

\bibitem{kumar2023pretraining}
Aviral Kumar, Anikait Singh, Frederik Ebert, Mitsuhiko Nakamoto, Yanlai Yang, Chelsea Finn, and Sergey Levine.
\newblock Pre-training for robots: Offline rl enables learning new tasks from a handful of trials.
\newblock In {\em RSS}, 2023.

\bibitem{kumar2020conservative}
Aviral Kumar, Aurick Zhou, George Tucker, and Sergey Levine.
\newblock Conservative q-learning for offline reinforcement learning.
\newblock In {\em NeurIPS}, 2020.

\bibitem{huang2021generalization}
Wenlong Huang, Igor Mordatch, Pieter Abbeel, and Deepak Pathak.
\newblock Generalization in dexterous manipulation via geometry-aware multi-task learning.
\newblock {\em arXiv preprint arXiv:2111.03062}, 2021.

\bibitem{chen2022system}
Tao Chen, Jie Xu, and Pulkit Agrawal.
\newblock A system for general in-hand object re-orientation.
\newblock In {\em Conference on Robot Learning}, pages 297--307. PMLR, 2022.

\bibitem{chen2023visual}
Tao Chen, Megha Tippur, Siyang Wu, Vikash Kumar, Edward Adelson, and Pulkit Agrawal.
\newblock Visual dexterity: In-hand reorientation of novel and complex object shapes.
\newblock {\em Science Robotics}, 8(84):eadc9244, 2023.

\bibitem{parisi2022unsurprising}
Simone Parisi, Aravind Rajeswaran, Senthil Purushwalkam, and Abhinav Gupta.
\newblock The unsurprising effectiveness of pre-trained vision models for control, 2022.

\bibitem{li2022pre}
Shuang Li, Xavier Puig, Chris Paxton, Yilun Du, Clinton Wang, Linxi Fan, Tao Chen, De-An Huang, Ekin Aky{\"u}rek, Anima Anandkumar, et~al.
\newblock Pre-trained language models for interactive decision-making.
\newblock {\em Advances in Neural Information Processing Systems}, 35:31199--31212, 2022.

\bibitem{nair2022r3m}
Suraj Nair, Aravind Rajeswaran, Vikash Kumar, Chelsea Finn, and Abhinav Gupta.
\newblock R3m: A universal visual representation for robot manipulation, 2022.

\bibitem{radosavovic2023real}
Ilija Radosavovic, Tete Xiao, Stephen James, Pieter Abbeel, Jitendra Malik, and Trevor Darrell.
\newblock Real-world robot learning with masked visual pre-training.
\newblock In {\em Conference on Robot Learning}, pages 416--426. PMLR, 2023.

\bibitem{radosavovic2023robot}
Ilija Radosavovic, Baifeng Shi, Letian Fu, Ken Goldberg, Trevor Darrell, and Jitendra Malik.
\newblock Robot learning with sensorimotor pre-training, 2023.

\bibitem{hansen2023pretraining}
Nicklas Hansen, Zhecheng Yuan, Yanjie Ze, Tongzhou Mu, Aravind Rajeswaran, Hao Su, Huazhe Xu, and Xiaolong Wang.
\newblock On pre-training for visuo-motor control: Revisiting a learning-from-scratch baseline.
\newblock In {\em ICML}, 2023.

\bibitem{Majumdar2023VC1}
Arjun Majumdar, Karmesh Yadav, Sergio Arnaud, Yecheng~Jason Ma, Claire Chen, Sneha Silwal, Aryan Jain, Vincent-Pierre Berges, Pieter Abbeel, Jitendra Malik, Dhruv Batra, Yixin Lin, Oleksandr Maksymets, Aravind Rajeswaran, and Franziska Meier.
\newblock Where are we in the search for an artificial visual cortex for embodied intelligence?, 2023.

\bibitem{bahl2023affordances}
Shikhar Bahl, Russell Mendonca, Lili Chen, Unnat Jain, and Deepak Pathak.
\newblock Affordances from human videos as a versatile representation for robotics.
\newblock {\em CVPR}, 2023.

\bibitem{thankaraj2022sounds}
Abitha Thankaraj and Lerrel Pinto.
\newblock That sounds right: Auditory self-supervision for dynamic robot manipulation, 2022.

\bibitem{guzey2023dexterity}
Irmak Guzey, Ben Evans, Soumith Chintala, and Lerrel Pinto.
\newblock Dexterity from touch: Self-supervised pre-training of tactile representations with robotic play, 2023.

\bibitem{Shah_2023_GNM}
Dhruv Shah, Ajay Sridhar, Arjun Bhorkar, Noriaki Hirose, and Sergey Levine.
\newblock Gnm: A general navigation model to drive any robot.
\newblock In {\em ICRA}, 2023.

\bibitem{Truong_2023_IndoorSimtoOutdoorReal}
Joanne Truong, April Zitkovich, Sonia Chernova, Dhruv Batra, Tingnan Zhang, Jie Tan, and Wenhao Yu.
\newblock Indoorsim-to-outdoorreal: Learning to navigate outdoors without any outdoor experience.
\newblock In {\em arXiv:2305.01098}, 2023.

\bibitem{Bonatti_2022_PACT}
Rogerio Bonatti, Sai Vemprala, Shuang Ma, Felipe Frujeri, Shuhang Chen, and Ashish Kapoor.
\newblock Pact: Perception-action causal transformer for autoregressive robotics pre-training.
\newblock In {\em arXiv:2209.11133}, 2022.

\bibitem{gu2023conceptgraphs}
Qiao Gu, Alihusein Kuwajerwala, Sacha Morin, Krishna~Murthy Jatavallabhula, Bipasha Sen, Aditya Agarwal, Corban Rivera, William Paul, Kirsty Ellis, Rama Chellappa, et~al.
\newblock Conceptgraphs: Open-vocabulary 3d scene graphs for perception and planning.
\newblock {\em arXiv preprint arXiv:2309.16650}, 2023.

\bibitem{hao2023reasoning}
Shibo Hao, Yi~Gu, Haodi Ma, Joshua~Jiahua Hong, Zhen Wang, Daisy~Zhe Wang, and Zhiting Hu.
\newblock Reasoning with language model is planning with world model.
\newblock {\em arXiv preprint arXiv:2305.14992}, 2023.

\bibitem{yang2023learning}
Mengjiao Yang, Yilun Du, Kamyar Ghasemipour, Jonathan Tompson, Dale Schuurmans, and Pieter Abbeel.
\newblock Learning interactive real-world simulators.
\newblock {\em arXiv preprint arXiv:2310.06114}, 2023.

\bibitem{karamcheti2022lila}
Siddharth Karamcheti, Megha Srivastava, Percy Liang, and Dorsa Sadigh.
\newblock Lila: Language-informed latent actions.
\newblock In {\em Conference on Robot Learning}, pages 1379--1390. PMLR, 2022.

\bibitem{Doshi24-crossformer}
Ria Doshi, Homer Walke, Oier Mees, Sudeep Dasari, and Sergey Levine.
\newblock Scaling cross-embodied learning: One policy for manipulation, navigation, locomotion and aviation.
\newblock {\em arXiv preprint arXiv:2408.11812}, 2024.

\bibitem{bahl2022human}
Shikhar Bahl, Abhinav Gupta, and Deepak Pathak.
\newblock Human-to-robot imitation in the wild.
\newblock In {\em RSS}, 2022.

\bibitem{jain2022transformers}
Vidhi Jain, Yixin Lin, Eric Undersander, Yonatan Bisk, and Akshara Rai.
\newblock Transformers are adaptable task planners.
\newblock In {\em 6th Annual Conference on Robot Learning}, 2022.

\bibitem{andreas2017modular}
Jacob Andreas, Dan Klein, and Sergey Levine.
\newblock Modular multitask reinforcement learning with policy sketches.
\newblock In {\em International conference on machine learning}, pages 166--175. PMLR, 2017.

\bibitem{skubic2007using}
Marjorie Skubic, Derek Anderson, Samuel Blisard, Dennis Perzanowski, and Alan Schultz.
\newblock Using a hand-drawn sketch to control a team of robots.
\newblock {\em Autonomous Robots}, 22:399--410, 2007.

\bibitem{wei2022chain}
Jason Wei, Xuezhi Wang, Dale Schuurmans, Maarten Bosma, Fei Xia, Ed~Chi, Quoc~V Le, Denny Zhou, et~al.
\newblock Chain-of-thought prompting elicits reasoning in large language models.
\newblock {\em Advances in Neural Information Processing Systems}, 35:24824--24837, 2022.

\bibitem{wu2023towards}
Jianzong Wu, Xiangtai Li, Shilin Xu~Haobo Yuan, Henghui Ding, Yibo Yang, Xia Li, Jiangning Zhang, Yunhai Tong, Xudong Jiang, Bernard Ghanem, et~al.
\newblock Towards open vocabulary learning: A survey.
\newblock {\em arXiv preprint arXiv:2306.15880}, 2023.

\bibitem{cui2023holistic}
Chenhang Cui, Yiyang Zhou, Xinyu Yang, Shirley Wu, Linjun Zhang, James Zou, and Huaxiu Yao.
\newblock Holistic analysis of hallucination in gpt-4v (ision): Bias and interference challenges.
\newblock {\em arXiv preprint arXiv:2311.03287}, 2023.

\bibitem{gao2023physically}
Jensen Gao, Bidipta Sarkar, Fei Xia, Ted Xiao, Jiajun Wu, Brian Ichter, Anirudha Majumdar, and Dorsa Sadigh.
\newblock Physically grounded vision-language models for robotic manipulation.
\newblock In {\em arxiv}, 2023.

\bibitem{dasari2020robonet}
Sudeep Dasari, Frederik Ebert, Stephen Tian, Suraj Nair, Bernadette Bucher, Karl Schmeckpeper, Siddharth Singh, Sergey Levine, and Chelsea Finn.
\newblock Robonet: Large-scale multi-robot learning, 2020.

\bibitem{ebert2021bridge}
Frederik Ebert, Yanlai Yang, Karl Schmeckpeper, Bernadette Bucher, Georgios Georgakis, Kostas Daniilidis, Chelsea Finn, and Sergey Levine.
\newblock Bridge data: Boosting generalization of robotic skills with cross-domain datasets, 2021.

\bibitem{walke2023bridgedata}
Homer Walke, Kevin Black, Abraham Lee, Moo~Jin Kim, Max Du, Chongyi Zheng, Tony Zhao, Philippe Hansen-Estruch, Quan Vuong, Andre He, et~al.
\newblock Bridgedata v2: A dataset for robot learning at scale.
\newblock {\em arXiv preprint arXiv:2308.12952}, 2023.

\bibitem{shaw2023leap}
Kenneth Shaw, Ananye Agarwal, and Deepak Pathak.
\newblock Leap hand: Low-cost, efficient, and anthropomorphic hand for robot learning.
\newblock {\em arXiv preprint arXiv:2309.06440}, 2023.

\bibitem{chang2017matterport3d}
Angel Chang, Angela Dai, Thomas Funkhouser, Maciej Halber, Matthias Niessner, Manolis Savva, Shuran Song, Andy Zeng, and Yinda Zhang.
\newblock Matterport3d: Learning from rgb-d data in indoor environments.
\newblock {\em arXiv preprint arXiv:1709.06158}, 2017.

\bibitem{xia2018gibson}
Fei Xia, Amir~R Zamir, Zhiyang He, Alexander Sax, Jitendra Malik, and Silvio Savarese.
\newblock Gibson env: Real-world perception for embodied agents.
\newblock In {\em Proceedings of the IEEE conference on computer vision and pattern recognition}, pages 9068--9079, 2018.

\bibitem{19iccvhabitat}
{Manolis Savva*}, {Abhishek Kadian*}, {Oleksandr Maksymets*}, Yili Zhao, Erik Wijmans, Bhavana Jain, Julian Straub, Jia Liu, Vladlen Koltun, Jitendra Malik, Devi Parikh, and Dhruv Batra.
\newblock Habitat: {A} {P}latform for {E}mbodied {AI} {R}esearch.
\newblock In {\em Proceedings of the IEEE/CVF International Conference on Computer Vision (ICCV)}, 2019.

\bibitem{kolve2017ai2}
Eric Kolve, Roozbeh Mottaghi, Winson Han, Eli VanderBilt, Luca Weihs, Alvaro Herrasti, Matt Deitke, Kiana Ehsani, Daniel Gordon, Yuke Zhu, et~al.
\newblock Ai2-thor: An interactive 3d environment for visual ai.
\newblock {\em arXiv preprint arXiv:1712.05474}, 2017.

\bibitem{min2022film}
So~Yeon Min, Devendra~Singh Chaplot, Pradeep Ravikumar, Yonatan Bisk, and Ruslan Salakhutdinov.
\newblock Film: Following instructions in language with modular methods, 2022.

\bibitem{shah2017airsim}
Shital Shah, Debadeepta Dey, Chris Lovett, and Ashish Kapoor.
\newblock Airsim: High-fidelity visual and physical simulation for autonomous vehicles, 2017.

\bibitem{mujoco3}
{Google DeepMind}.
\newblock {Mujoco} 3.0.
\newblock \url{https://github.com/google-deepmind/mujoco/releases/tag/3.0.0}, 2023.
\newblock Accessed: [Insert date of access].

\bibitem{chignoli2021humanoid}
Matthew Chignoli, Donghyun Kim, Elijah Stanger-Jones, and Sangbae Kim.
\newblock The mit humanoid robot: Design, motion planning, and control for acrobatic behaviors.
\newblock In {\em 2020 IEEE-RAS 20th International Conference on Humanoid Robots (Humanoids)}, pages 1--8. IEEE, 2021.

\bibitem{zhao2021modelfree}
Weiye Zhao, Tairan He, and Changliu Liu.
\newblock Model-free safe control for zero-violation reinforcement learning.
\newblock In {\em 5th Annual Conference on Robot Learning}, 2021.

\bibitem{herbert2021scalable}
Sylvia Herbert, Jason~J. Choi, Suvansh Sanjeev, Marsalis Gibson, Koushil Sreenath, and Claire~J. Tomlin.
\newblock Scalable learning of safety guarantees for autonomous systems using hamilton-jacobi reachability, 2021.

\bibitem{tian2022safety}
Ran Tian, Liting Sun, Andrea Bajcsy, Masayoshi Tomizuka, and Anca~D Dragan.
\newblock Safety assurances for human-robot interaction via confidence-aware game-theoretic human models.
\newblock In {\em 2022 International Conference on Robotics and Automation (ICRA)}, pages 11229--11235. IEEE, 2022.

\bibitem{safetyasaffordance8525627}
S.H. Cheong, J.H. Lee, and C.H. Kim.
\newblock A new concept of safety affordance map for robots object manipulation.
\newblock In {\em 2018 27th IEEE International Symposium on Robot and Human Interactive Communication (RO-MAN)}, pages 565--570, 2018.

\bibitem{yang2023plug}
Ziyi Yang, Shreyas~Sundara Raman, Ankit Shah, and Stefanie Tellex.
\newblock Plug in the safety chip: Enforcing constraints for {LLM}-driven robot agents.
\newblock In {\em 2nd Workshop on Language and Robot Learning: Language as Grounding}, 2023.

\bibitem{sporns2016modular}
Olaf Sporns and Richard~F Betzel.
\newblock Modular brain networks.
\newblock {\em Annual review of psychology}, 67:613--640, 2016.

\bibitem{meunier2010modular}
David Meunier, Renaud Lambiotte, and Edward~T Bullmore.
\newblock Modular and hierarchically modular organization of brain networks.
\newblock {\em Frontiers in neuroscience}, 4:200, 2010.

\bibitem{vanhoucke2018endtoend}
Vincent Vanhoucke.
\newblock The end-to-end false dichotomy: Roboticists arguing lego vs. playmo.
\newblock {\em Medium}, October 28 2018.

\bibitem{lecun2022path}
Yann LeCun.
\newblock A path towards autonomous machine intelligence version 0.9. 2, 2022-06-27.
\newblock {\em Open Review}, 62, 2022.

\bibitem{videoworldsimulators2024}
Tim Brooks, Bill Peebles, Connor Holmes, Will DePue, Yufei Guo, Li~Jing, David Schnurr, Joe Taylor, Troy Luhman, Eric Luhman, Clarence Ng, Ricky Wang, and Aditya Ramesh.
\newblock Video generation models as world simulators.
\newblock 2024.

\bibitem{bruce2024genie}
Jake Bruce, Michael Dennis, Ashley Edwards, Jack Parker-Holder, Yuge Shi, Edward Hughes, Matthew Lai, Aditi Mavalankar, Richie Steigerwald, Chris Apps, Yusuf Aytar, Sarah Bechtle, Feryal Behbahani, Stephanie Chan, Nicolas Heess, Lucy Gonzalez, Simon Osindero, Sherjil Ozair, Scott Reed, Jingwei Zhang, Konrad Zolna, Jeff Clune, Nando de~Freitas, Satinder Singh, and Tim Rocktäschel.
\newblock Genie: Generative interactive environments, 2024.

\bibitem{shadowrohand}
{Shadow Robot Company}.
\newblock Dexterous hand series.
\newblock \url{https://www.shadowrobot.com/dexterous-hand-series/}, 2023.
\newblock Accessed: 2023-12-10.

\bibitem{Dong_2017}
Siyuan Dong, Wenzhen Yuan, and Edward~H. Adelson.
\newblock Improved gelsight tactile sensor for measuring geometry and slip.
\newblock In {\em 2017 IEEE/RSJ International Conference on Intelligent Robots and Systems (IROS)}. IEEE, September 2017.

\bibitem{si2023robotsweater}
Zilin Si, Tianhong~Catherine Yu, Katrene Morozov, James McCann, and Wenzhen Yuan.
\newblock Robotsweater: Scalable, generalizable, and customizable machine-knitted tactile skins for robots.
\newblock {\em arXiv preprint arXiv:2303.02858}, 2023.

\bibitem{romero2024eyesight}
Branden Romero, Hao-Shu Fang, Pulkit Agrawal, and Edward Adelson.
\newblock Eyesight hand: Design of a fully-actuated dexterous robot hand with integrated vision-based tactile sensors and compliant actuation.
\newblock {\em arXiv preprint arXiv:2408.06265}, 2024.

\bibitem{zhao2023learning}
Tony~Z. Zhao, Vikash Kumar, Sergey Levine, and Chelsea Finn.
\newblock Learning fine-grained bimanual manipulation with low-cost hardware, 2023.

\bibitem{fang2024airexo}
Hongjie Fang, Hao-Shu Fang, Yiming Wang, Jieji Ren, Jingjing Chen, Ruo Zhang, Weiming Wang, and Cewu Lu.
\newblock Airexo: Low-cost exoskeletons for learning whole-arm manipulation in the wild.
\newblock In {\em 2024 IEEE International Conference on Robotics and Automation (ICRA)}, pages 15031--15038. IEEE, 2024.

\bibitem{kasai2022realtime}
Jungo Kasai, Keisuke Sakaguchi, Yoichi Takahashi, Ronan~Le Bras, Akari Asai, Xinyan Yu, Dragomir Radev, Noah~A. Smith, Yejin Choi, and Kentaro Inui.
\newblock Realtime qa: What's the answer right now?, 2022.

\bibitem{Mehta2022DSIUT}
Sanket~Vaibhav Mehta, Jai Gupta, Yi~Tay, Mostafa Dehghani, Vinh~Q. Tran, Jinfeng Rao, Marc Najork, Emma Strubell, and Donald Metzler.
\newblock Dsi++: Updating transformer memory with new documents.
\newblock {\em ArXiv}, abs/2212.09744, 2022.

\bibitem{mehta2023empirical}
Sanket~Vaibhav Mehta, Darshan Patil, Sarath Chandar, and Emma Strubell.
\newblock An empirical investigation of the role of pre-training in lifelong learning.
\newblock {\em Journal of Machine Learning Research}, 24(214):1--50, 2023.

\bibitem{smith2023construct}
James~Seale Smith, Paola Cascante-Bonilla, Assaf Arbelle, Donghyun Kim, Rameswar Panda, David Cox, Diyi Yang, Zsolt Kira, Rogerio Feris, and Leonid Karlinsky.
\newblock Construct-vl: Data-free continual structured vl concepts learning.
\newblock In {\em Proceedings of the IEEE/CVF Conference on Computer Vision and Pattern Recognition}, pages 14994--15004, 2023.

\bibitem{Lesort_2020}
Timothée Lesort, Vincenzo Lomonaco, Andrei Stoian, Davide Maltoni, David Filliat, and Natalia Díaz-Rodríguez.
\newblock Continual learning for robotics: Definition, framework, learning strategies, opportunities and challenges.
\newblock {\em Information Fusion}, 58:52–68, June 2020.

\bibitem{pmlr-v78-maeda17a}
Guilherme Maeda, Marco Ewerton, Takayuki Osa, Baptiste Busch, and Jan Peters.
\newblock Active incremental learning of robot movement primitives.
\newblock In Sergey Levine, Vincent Vanhoucke, and Ken Goldberg, editors, {\em Proceedings of the 1st Annual Conference on Robot Learning}, volume~78 of {\em Proceedings of Machine Learning Research}, pages 37--46. PMLR, 13--15 Nov 2017.

\bibitem{kumar2021rma}
Ashish Kumar, Zipeng Fu, Deepak Pathak, and Jitendra Malik.
\newblock Rma: Rapid motor adaptation for legged robots.
\newblock In {\em Robotics: Science and Systems}, 2021.

\bibitem{hejna2023inverse}
Joey Hejna and Dorsa Sadigh.
\newblock Inverse preference learning: Preference-based rl without a reward function, 2023.

\bibitem{li2023reinforcement}
Zihao Li, Zhuoran Yang, and Mengdi Wang.
\newblock Reinforcement learning with human feedback: Learning dynamic choices via pessimism, 2023.

\bibitem{kumar2022finetuning}
Ananya Kumar, Aditi Raghunathan, Robbie~Matthew Jones, Tengyu Ma, and Percy Liang.
\newblock Fine-tuning can distort pretrained features and underperform out-of-distribution.
\newblock In {\em International Conference on Learning Representations}, 2022.

\bibitem{openai2019solving}
OpenAI, Ilge Akkaya, Marcin Andrychowicz, Maciek Chociej, Mateusz Litwin, Bob McGrew, Arthur Petron, Alex Paino, Matthias Plappert, Glenn Powell, Raphael Ribas, Jonas Schneider, Nikolas Tezak, Jerry Tworek, Peter Welinder, Lilian Weng, Qiming Yuan, Wojciech Zaremba, and Lei Zhang.
\newblock Solving rubik's cube with a robot hand, 2019.

\bibitem{mehta2023sample}
Viraj Mehta, Vikramjeet Das, Ojash Neopane, Yijia Dai, Ilija Bogunovic, Jeff Schneider, and Willie Neiswanger.
\newblock Sample efficient reinforcement learning from human feedback via active exploration, 2023.

\bibitem{an2023direct}
Gaon An, Junhyeok Lee, Xingdong Zuo, Norio Kosaka, Kyung-Min Kim, and Hyun~Oh Song.
\newblock Direct preference-based policy optimization without reward modeling, 2023.

\bibitem{ajay2023compositional}
Anurag Ajay, Seungwook Han, Yilun Du, Shuang Li, Abhi Gupta, Tommi Jaakkola, Josh Tenenbaum, Leslie Kaelbling, Akash Srivastava, and Pulkit Agrawal.
\newblock Compositional foundation models for hierarchical planning, 2023.

\bibitem{frans2017meta}
Kevin Frans, Jonathan Ho, Xi~Chen, Pieter Abbeel, and John Schulman.
\newblock Meta learning shared hierarchies, 2017.

\bibitem{Nair2020Hierarchical}
Suraj Nair and Chelsea Finn.
\newblock Hierarchical foresight: Self-supervised learning of long-horizon tasks via visual subgoal generation.
\newblock In {\em International Conference on Learning Representations}, 2020.

\bibitem{feichtenhofer2019slowfast}
Christoph Feichtenhofer, Haoqi Fan, Jitendra Malik, and Kaiming He.
\newblock Slowfast networks for video recognition, 2019.

\bibitem{khazatsky2024droid}
Alexander Khazatsky, Karl Pertsch, Suraj Nair, Ashwin Balakrishna, Sudeep Dasari, Siddharth Karamcheti, Soroush Nasiriany, Mohan~Kumar Srirama, Lawrence~Yunliang Chen, Kirsty Ellis, Peter~David Fagan, Joey Hejna, Masha Itkina, Marion Lepert, Yecheng~Jason Ma, Patrick~Tree Miller, Jimmy Wu, Suneel Belkhale, Shivin Dass, Huy Ha, Arhan Jain, Abraham Lee, Youngwoon Lee, Marius Memmel, Sungjae Park, Ilija Radosavovic, Kaiyuan Wang, Albert Zhan, Kevin Black, Cheng Chi, Kyle~Beltran Hatch, Shan Lin, Jingpei Lu, Jean Mercat, Abdul Rehman, Pannag~R Sanketi, Archit Sharma, Cody Simpson, Quan Vuong, Homer~Rich Walke, Blake Wulfe, Ted Xiao, Jonathan~Heewon Yang, Arefeh Yavary, Tony~Z. Zhao, Christopher Agia, Rohan Baijal, Mateo~Guaman Castro, Daphne Chen, Qiuyu Chen, Trinity Chung, Jaimyn Drake, Ethan~Paul Foster, Jensen Gao, David~Antonio Herrera, Minho Heo, Kyle Hsu, Jiaheng Hu, Donovon Jackson, Charlotte Le, Yunshuang Li, Kevin Lin, Roy Lin, Zehan Ma, Abhiram Maddukuri, Suvir Mirchandani, Daniel Morton, Tony Nguyen,
  Abigail O'Neill, Rosario Scalise, Derick Seale, Victor Son, Stephen Tian, Emi Tran, Andrew~E. Wang, Yilin Wu, Annie Xie, Jingyun Yang, Patrick Yin, Yunchu Zhang, Osbert Bastani, Glen Berseth, Jeannette Bohg, Ken Goldberg, Abhinav Gupta, Abhishek Gupta, Dinesh Jayaraman, Joseph~J Lim, Jitendra Malik, Roberto Martín-Martín, Subramanian Ramamoorthy, Dorsa Sadigh, Shuran Song, Jiajun Wu, Michael~C. Yip, Yuke Zhu, Thomas Kollar, Sergey Levine, and Chelsea Finn.
\newblock Droid: A large-scale in-the-wild robot manipulation dataset.
\newblock 2024.

\bibitem{dalal2023optimus}
Murtaza Dalal, Ajay Mandlekar, Caelan Garrett, Ankur Handa, Ruslan Salakhutdinov, and Dieter Fox.
\newblock Imitating task and motion planning with visuomotor transformers.
\newblock 2023.

\bibitem{2023_Cheng_extremeParkour}
Xuxin Cheng, Kexin Shi, Ananye Agarwal, and Deepak Pathak.
\newblock Extreme parkour with legged robots.
\newblock In {\em arXiv:2309.14341}, 2023.

\bibitem{wen2024foundationposeunified6dpose}
Bowen Wen, Wei Yang, Jan Kautz, and Stan Birchfield.
\newblock Foundationpose: Unified 6d pose estimation and tracking of novel objects, 2024.

\bibitem{hollmann2024large}
Noah Hollmann, Samuel M{\"u}ller, and Frank Hutter.
\newblock Large language models for automated data science: Introducing caafe for context-aware automated feature engineering.
\newblock {\em Advances in Neural Information Processing Systems}, 36, 2024.

\bibitem{liu2023reflect}
Zeyi Liu, Arpit Bahety, and Shuran Song.
\newblock Reflect: Summarizing robot experiences for failure explanation and correction.
\newblock In {\em Conference on Robot Learning}, pages 3468--3484. PMLR, 2023.

\bibitem{dang2020sensor}
L~Minh Dang, Kyungbok Min, Hanxiang Wang, Md~Jalil Piran, Cheol~Hee Lee, and Hyeonjoon Moon.
\newblock Sensor-based and vision-based human activity recognition: A comprehensive survey.
\newblock {\em Pattern Recognition}, 108:107561, 2020.

\bibitem{rizk2023case}
Yara Rizk, Praveen Venkateswaran, Vatche Isahagian, Austin Narcomey, and Vinod Muthusamy.
\newblock A case for business process-specific foundation models.
\newblock In {\em International Conference on Business Process Management}, pages 44--56. Springer, 2023.

\bibitem{goldberg2022characterizing}
David~M Goldberg.
\newblock Characterizing accident narratives with word embeddings: Improving accuracy, richness, and generalizability.
\newblock {\em Journal of safety research}, 80:441--455, 2022.

\bibitem{jatavallabhula2023conceptfusion}
Krishna~Murthy Jatavallabhula, Alihusein Kuwajerwala, Qiao Gu, et~al.
\newblock Conceptfusion: Open-set multimodal 3d mapping.
\newblock {\em RSS}, 2023.

\bibitem{yamazaki2024open}
Kashu Yamazaki, Taisei Hanyu, Khoa Vo, Thang Pham, Minh Tran, Gianfranco Doretto, Anh Nguyen, and Ngan Le.
\newblock Open-fusion: Real-time open-vocabulary 3d mapping and queryable scene representation.
\newblock In {\em 2024 IEEE International Conference on Robotics and Automation (ICRA)}, pages 9411--9417. IEEE, 2024.

\bibitem{robla2017working}
Sandra Robla-G{\'o}mez, Victor~M Becerra, Jos{\'e}~Ram{\'o}n Llata, Esther Gonzalez-Sarabia, Carlos Torre-Ferrero, and Juan Perez-Oria.
\newblock Working together: A review on safe human-robot collaboration in industrial environments.
\newblock {\em Ieee Access}, 5:26754--26773, 2017.

\bibitem{marshall2016robotics}
Joshua~A Marshall, Adrian Bonchis, Eduardo Nebot, and Steven Scheding.
\newblock Robotics in mining.
\newblock {\em Springer handbook of robotics}, pages 1549--1576, 2016.

\bibitem{chen2023playfusion}
Lili Chen, Shikhar Bahl, and Deepak Pathak.
\newblock Playfusion: Skill acquisition via diffusion from language-annotated play.
\newblock In {\em Conference on Robot Learning}, pages 2012--2029. PMLR, 2023.

\bibitem{hubara2018quantized}
Itay Hubara, Matthieu Courbariaux, Daniel Soudry, Ran El-Yaniv, and Yoshua Bengio.
\newblock Quantized neural networks: Training neural networks with low precision weights and activations.
\newblock {\em Journal of Machine Learning Research}, 18(187):1--30, 2018.

\bibitem{hinton2015distilling}
Geoffrey Hinton, Oriol Vinyals, and Jeff Dean.
\newblock Distilling the knowledge in a neural network.
\newblock {\em arXiv preprint arXiv:1503.02531}, 2015.

\bibitem{omidshafiei2019teaching}
Shayegan Omidshafiei, Dong-Ki Kim, Miao Liu, Gerald Tesauro, Matthew Riemer, Christopher Amato, Murray Campbell, and Jonathan~P. How.
\newblock Learning to teach in cooperative multiagent reinforcement learning.
\newblock {\em Proceedings of the AAAI Conference on Artificial Intelligence}, 33(01):6128--6136, Jul. 2019.

\bibitem{wang2021knowledge}
Lin Wang and Kuk-Jin Yoon.
\newblock Knowledge distillation and student-teacher learning for visual intelligence: A review and new outlooks.
\newblock {\em IEEE transactions on pattern analysis and machine intelligence}, 44(6):3048--3068, 2021.

\bibitem{dass2023clvr}
S.~Dass, J.~Yapeter, J.~Zhang, J.~Zhang, K.~Pertsch, S.~Nikolaidis, and J.~J. Lim.
\newblock Clvr jaco play dataset.
\newblock \url{https://github.com/clvrai/clvr-jaco-play-dataset}, 2023.

\bibitem{luo2023multistage}
Jianlan Luo, Charles Xu, Xinyang Geng, Gilbert Feng, Kuan Fang, Liam Tan, Stefan Schaal, and Sergey Levine.
\newblock Multi-stage cable routing through hierarchical imitation learning, 2023.

\bibitem{pari2021surprising}
Jyothish Pari, Nur~Muhammad Shafiullah, Sridhar~Pandian Arunachalam, and Lerrel Pinto.
\newblock The surprising effectiveness of representation learning for visual imitation.
\newblock {\em arXiv preprint arXiv:2112.01511}, 2021.

\bibitem{chen2023berkeley}
L.~Y. Chen, S.~Adebola, and K.~Goldberg.
\newblock Berkeley ur5 demonstration dataset.
\newblock \url{https://sites.google.com/view/berkeley-ur5/home}.
\newblock Accessed: [Insert Date Here].

\bibitem{rosete2022tacorl}
Erick Rosete-Beas, Oier Mees, Gabriel Kalweit, Joschka Boedecker, and Wolfram Burgard.
\newblock Latent plans for task agnostic offline reinforcement learning.
\newblock In {\em Proceedings of the 6th Conference on Robot Learning (CoRL)}, 2022.

\bibitem{zhao2023chat}
Xufeng Zhao, Mengdi Li, Cornelius Weber, Muhammad~Burhan Hafez, and Stefan Wermter.
\newblock Chat with the environment: Interactive multimodal perception using large language models, 2023.

\bibitem{coppeliasim}
{Coppelia Robotics}.
\newblock Coppeliasim.
\newblock \url{https://www.coppeliarobotics.com/}.
\newblock Accessed: [Insert Date Here].

\bibitem{CoumansPybullet}
E.~Coumans and Y.~Bai.
\newblock Pybullet, a python module for physics simulation for games, robotics and machine learning.

\bibitem{huang2023instruct2act}
Siyuan Huang, Zhengkai Jiang, Hao Dong, Yu~Qiao, Peng Gao, and Hongsheng Li.
\newblock Instruct2act: Mapping multi-modality instructions to robotic actions with large language model, 2023.

\bibitem{xiang2020sapien}
Fanbo Xiang, Yuzhe Qin, Kaichun Mo, Yikuan Xia, Hao Zhu, Fangchen Liu, Minghua Liu, Hanxiao Jiang, Yifu Yuan, He~Wang, et~al.
\newblock Sapien: A simulated part-based interactive environment.
\newblock In {\em Proceedings of the IEEE/CVF Conference on Computer Vision and Pattern Recognition}, pages 11097--11107, 2020.

\bibitem{mo2021where2act}
Kaichun Mo, Leonidas Guibas, Mustafa Mukadam, Abhinav Gupta, and Shubham Tulsiani.
\newblock Where2act: From pixels to actions for articulated 3d objects, 2021.

\bibitem{lee2021pickandplace}
Alex~X. Lee, Coline Devin, Yuxiang Zhou, Thomas Lampe, Konstantinos Bousmalis, Jost~Tobias Springenberg, Arunkumar Byravan, Abbas Abdolmaleki, Nimrod Gileadi, David Khosid, Claudio Fantacci, Jose~Enrique Chen, Akhil Raju, Rae Jeong, Michael Neunert, Antoine Laurens, Stefano Saliceti, Federico Casarini, Martin Riedmiller, Raia Hadsell, and Francesco Nori.
\newblock Beyond pick-and-place: Tackling robotic stacking of diverse shapes, 2021.

\bibitem{szot2023large}
Andrew Szot, Max Schwarzer, Harsh Agrawal, Bogdan Mazoure, Walter Talbott, Katherine Metcalf, Natalie Mackraz, Devon Hjelm, and Alexander Toshev.
\newblock Large language models as generalizable policies for embodied tasks, 2023.

\bibitem{chen2021evaluating}
Mark Chen, Jerry Tworek, Heewoo Jun, Qiming Yuan, Henrique Ponde de~Oliveira Pinto, Jared Kaplan, Harri Edwards, Yuri Burda, Nicholas Joseph, Greg Brockman, et~al.
\newblock Evaluating large language models trained on code.
\newblock {\em arXiv preprint arXiv:2107.03374}, 2021.

\bibitem{zeng2020transporter}
Andy Zeng, Pete Florence, Jonathan Tompson, Stefan Welker, Jonathan Chien, Maria Attarian, Travis Armstrong, Ivan Krasin, Dan Duong, Vikas Sindhwani, and Johnny Lee.
\newblock Transporter networks: Rearranging the visual world for robotic manipulation.
\newblock {\em Conference on Robot Learning (CoRL)}, 2020.

\bibitem{ma2023liv}
Yecheng~Jason Ma, William Liang, Vaidehi Som, Vikash Kumar, Amy Zhang, Osbert Bastani, and Dinesh Jayaraman.
\newblock Liv: Language-image representations and rewards for robotic control, 2023.

\bibitem{yu2021metaworld}
Tianhe Yu, Deirdre Quillen, Zhanpeng He, Ryan Julian, Avnish Narayan, Hayden Shively, Adithya Bellathur, Karol Hausman, Chelsea Finn, and Sergey Levine.
\newblock Meta-world: A benchmark and evaluation for multi-task and meta reinforcement learning, 2021.

\bibitem{wu2023tidybot}
Jimmy Wu, Rika Antonova, Adam Kan, Marion Lepert, Andy Zeng, Shuran Song, Jeannette Bohg, Szymon Rusinkiewicz, and Thomas Funkhouser.
\newblock Tidybot: Personalized robot assistance with large language models.
\newblock {\em arXiv preprint arXiv:2305.05658}, 2023.

\bibitem{ding2023task}
Yan Ding, Xiaohan Zhang, Chris Paxton, and Shiqi Zhang.
\newblock Task and motion planning with large language models for object rearrangement.
\newblock {\em arXiv preprint arXiv:2303.06247}, 2023.

\bibitem{1389727}
N.~Koenig and A.~Howard.
\newblock Design and use paradigms for gazebo, an open-source multi-robot simulator.
\newblock In {\em 2004 IEEE/RSJ International Conference on Intelligent Robots and Systems (IROS) (IEEE Cat. No.04CH37566)}, volume~3, pages 2149--2154 vol.3, 2004.

\bibitem{yadav2023habitatmatterport}
Karmesh Yadav, Ram Ramrakhya, Santhosh~Kumar Ramakrishnan, Theo Gervet, John Turner, Aaron Gokaslan, Noah Maestre, Angel~Xuan Chang, Dhruv Batra, Manolis Savva, Alexander~William Clegg, and Devendra~Singh Chaplot.
\newblock Habitat-matterport 3d semantics dataset, 2023.

\bibitem{tokenhumanoid2024}
Ilija Radosavovic, Bike Zhang, Baifeng Shi, Jathushan Rajasegaran, Sarthak Kamat, Trevor Darrell, Koushil Sreenath, and Jitendra Malik.
\newblock Humanoid locomotion as next token prediction.
\newblock {\em arXiv:2402.19469}, 2024.

\bibitem{weber2017imagination}
Th{\'e}ophane Weber, S{\'e}bastien Racaniere, David~P Reichert, Lars Buesing, Arthur Guez, Danilo~Jimenez Rezende, Adria~Puigdomenech Badia, Oriol Vinyals, Nicolas Heess, Yujia Li, et~al.
\newblock Imagination-augmented agents for deep reinforcement learning.
\newblock {\em arXiv preprint arXiv:1707.06203}, 2017.

\bibitem{chevalier2018babyai}
Maxime Chevalier-Boisvert, Dzmitry Bahdanau, Salem Lahlou, Lucas Willems, Chitwan Saharia, Thien~Huu Nguyen, and Yoshua Bengio.
\newblock Babyai: A platform to study the sample efficiency of grounded language learning.
\newblock {\em arXiv preprint arXiv:1810.08272}, 2018.

\bibitem{cobbe2020leveraging}
Karl Cobbe, Chris Hesse, Jacob Hilton, and John Schulman.
\newblock Leveraging procedural generation to benchmark reinforcement learning.
\newblock In {\em International conference on machine learning}, pages 2048--2056. PMLR, 2020.

\bibitem{bellemare2013arcade}
Marc~G Bellemare, Yavar Naddaf, Joel Veness, and Michael Bowling.
\newblock The arcade learning environment: An evaluation platform for general agents.
\newblock {\em Journal of Artificial Intelligence Research}, 47:253--279, 2013.

\bibitem{tassa2018deepmind}
Yuval Tassa, Yotam Doron, Alistair Muldal, Tom Erez, Yazhe Li, Diego de~Las Casas, David Budden, Abbas Abdolmaleki, Josh Merel, Andrew Lefrancq, et~al.
\newblock Deepmind control suite.
\newblock {\em arXiv preprint arXiv:1801.00690}, 2018.

\bibitem{huang2023grounded}
Wenlong Huang, Fei Xia, Dhruv Shah, Danny Driess, Andy Zeng, Yao Lu, Pete Florence, Igor Mordatch, Sergey Levine, Karol Hausman, and Brian Ichter.
\newblock Grounded decoding: Guiding text generation with grounded models for robot control.
\newblock {\em ArXiv}, abs/2303.00855, 2023.

\bibitem{chevalierboisvert2018minimalistic}
M.~Chevalier-Boisvert, L.~Willems, and S.~Pal.
\newblock Minimalistic gridworld environment for gymnasium.
\newblock \url{https://github.com/pierg/environments-rl}, 2018.

\bibitem{chen2022towards}
Yuanpei Chen, Tianhao Wu, Shengjie Wang, Xidong Feng, Jiechuan Jiang, Zongqing Lu, Stephen McAleer, Hao Dong, Song-Chun Zhu, and Yaodong Yang.
\newblock Towards human-level bimanual dexterous manipulation with reinforcement learning.
\newblock {\em Advances in Neural Information Processing Systems}, 35:5150--5163, 2022.

\bibitem{howell2022predictive}
Taylor Howell, Nimrod Gileadi, Saran Tunyasuvunakool, Kevin Zakka, Tom Erez, and Yuval Tassa.
\newblock Predictive sampling: Real-time behaviour synthesis with mujoco, 2022.

\end{thebibliography}
}

\end{document}